\documentclass[aoas]{imsart}

\RequirePackage{amsthm,amsmath,amsfonts,amssymb}
\RequirePackage[authoryear]{natbib}
\RequirePackage[colorlinks,citecolor=blue,urlcolor=blue]{hyperref}
\RequirePackage{graphicx}
\RequirePackage{lineno,hyperref}
\RequirePackage{forest}
\RequirePackage[linguistics]{adjustbox}
\RequirePackage{bm}
\RequirePackage{dsfont}
\usepackage{anyfontsize}
\DeclareFontShape{T1}{ptm}{m}{scit}{<-> ssub * ptm/m/sc}{}
\RequirePackage{float}
\RequirePackage{support-caption}
\RequirePackage{subcaption}
\RequirePackage{algorithm}
\RequirePackage{algpseudocode}
\RequirePackage{mwe}
\RequirePackage{tabularx}
\RequirePackage{array}
\RequirePackage{booktabs}
\RequirePackage{rotating}
\RequirePackage{multirow}
\RequirePackage{lscape}
\RequirePackage{siunitx}
\RequirePackage{ragged2e}
\RequirePackage{enumitem}
\RequirePackage{blkarray}% http://ctan.org/pkg/blkarray
\RequirePackage{array,makecell}
\newcommand{\matindex}[1]{\mbox{\scriptsize#1}}% Matrix index
 \newcolumntype{P}{>{\RaggedRight\hangindent1em\hangafter1\relax}X}
 \newcommand\given[1][]{\:#1\vert\:}
\startlocaldefs
\theoremstyle{plain}

\theoremstyle{remark}

\begin{document}

\begin{frontmatter}
\title{Accounting for shared covariates in semi-parametric Bayesian additive regression trees}
%\title{A sample article title with some additional note\thanksref{t1}}
\runtitle{Accounting for shared covariates in semi-parametric BART}
%\thankstext{T1}{A sample additional note to the title.}

\begin{aug}
%%%%%%%%%%%%%%%%%%%%%%%%%%%%%%%%%%%%%%%%%%%%%%%
%% Only one address is permitted per author. %%
%% Only division, organization and e-mail is %%
%% included in the address.                  %%
%% Additional information can be included in %%
\author[A]{\fnms{Estev\~ao}~\snm{B. Prado}},
\author[B, C, D]{\fnms{Andrew}~\snm{C. Parnell}},
\author[B, C]{\fnms{Keefe}~\snm{Murphy}},
\author[B, C]{\fnms{Nathan}~\snm{McJames}},
\author[C]{\fnms{Ann}~\snm{O’Shea}}
\and
\author[B,C]{\fnms{Rafael}~\snm{A. Moral}}

%%%%%%%%%%%%%%%%%%%%%%%%%%%%%%%%%%%%%%%%%%%%%%
%% Addresses                                %%
%%%%%%%%%%%%%%%%%%%%%%%%%%%%%%%%%%%%%%%%%%%%%%
\address[A]{School of Mathematical Sciences, Lancaster University, United Kingdom.}
\vspace{-\smallskipamount}
\address[B]{Hamilton Institute, Maynooth University, Co. Kildare, Ireland.}
\vspace{-\smallskipamount}
\address[C]{Department of Mathematics \& Statistics, Maynooth University, Co. Kildare, Ireland.}
\vspace{-\smallskipamount}
\address[D]{Insight Centre for Data Analytics, Maynooth University, Co. Kildare, Ireland.}
\end{aug}

\begin{abstract}
We propose some extensions to semi-parametric models based on Bayesian additive regression trees (BART). In the semi-parametric BART paradigm, the response variable is approximated by a linear predictor and a BART model, where the linear component is responsible for estimating the main effects and BART accounts for non-specified interactions and non-linearities. Previous semi-parametric models based on BART have assumed that the set of covariates in the linear predictor and the BART model are mutually exclusive in an attempt to avoid poor coverage properties and reduce bias in the estimates of the parameters in the linear predictor. The main novelty in our approach lies in the way we change the tree-generation moves in BART to deal with this bias and resolve non-identifiability issues between the parametric and non-parametric components, even when they have covariates in common. This allows us to model complex interactions involving the covariates of primary interest, both among themselves and with those in the BART component. Our novel method is developed with a view to analysing data from an international education assessment, where certain predictors of students' achievements in mathematics are of particular interpretational interest. Through additional simulation studies and another application to a well-known benchmark dataset, we also show competitive performance when compared to regression models, alternative formulations of semi-parametric BART, and other tree-based methods. The implementation of the proposed method is available at \url{https://github.com/ebprado/CSP-BART}.
\end{abstract}


\end{frontmatter}
%%%%%%%%%%%%%%%%%%%%%%%%%%%%%%%%%%%%%%%%%%%%%%
%% Please use \tableofcontents for articles %%
%% with 50 pages and more                   %%
%%%%%%%%%%%%%%%%%%%%%%%%%%%%%%%%%%%%%%%%%%%%%%
%\tableofcontents

\section{Introduction}

Generalised linear models \citep[GLMs;][]{nelder1972generalized,mccullagh1989generalized} are frequently used in many different applications to predict a univariate response due to the ease of interpretation of the parameter estimates as well as the wide availability of statistical software that facilitates simple analyses. Besides the estimation of main effects, a common use of regression models is to measure the effects that combinations of covariates may have on the response. However, standard GLM settings require pre-specification of interaction terms, which is a complicated task with high-dimensional data. Furthermore, a key assumption in GLMs is that the specified covariates in the linear predictor, including potential interactions and higher-order terms, have a linear relationship with the expected value of the response variable through a defined link function. Extensions such as generalised additive models \citep[GAMs;][]{hastie1990generalized, wood2017generalized} require specification of main and interaction effects via a sum of (potentially non-linear) predictors. In GAMs, the non-linear relationships are usually captured via basis expansions of the covariates and constrained by a smoothing parameter. However, in problems where the numbers of covariates and/or observations are large, it may not be simple to specify the covariates and interactions that impact most on the response. Semi-parametric models \citep{harezlak2018semiparametric} have been proposed for situations where a mixture of linear and non-linear trends, as well as interactions, are required for accurately fitting the data at hand.

Semi-parametric Bayesian additive regression tree (BART) models \citep{zeldow2019semiparametric, tan2019bayesian, dorie2022stan} are black-box type algorithms which aim to tackle some of the key limitations often encountered when using GLMs to analyse datasets with a large number of covariates. Most commonly, they are used when it is of interest to quantify the relationships between covariates and the response. It is well-known that tree-based algorithms such as BART \citep{chipman2010bart} and random forests \citep{breiman2001random} are flexible and can produce more accurate predictions, as they remove the often restrictive assumption of linearity between the covariates and the response. However, prediction is not the most important aspect in many situations \citep[e.g.,][]{hill2011bayesian,zeldow2019semiparametric, hahn2020bayesian}. Instead, knowing how covariates impact the response is crucial; but this quantification is not easily interpretable with the standard BART model or random forests. Thus, the main appeal of semi-parametric BART models is that they allow us to look inside the black-box and provide interpretations for how some key inputs of primary interest are converted into outputs. Unlike GLMs and GAMs, however, they account for non-specified interactions automatically. This is key to their appeal for analysing datasets with a large number of covariates where interaction effects are difficult to specify.

Motivated by data collected in 2019 under the seventh cycle of the quadrennial Trends in International Mathematics and Science Study \citep[TIMSS;][]{mullis2020timss,timss2019}, we extend the semi-parametric BART model introduced by \citet{zeldow2019semiparametric}, which we henceforth refer to as separated semi-parametric BART (SSP-BART) for clarity. TIMSS is an international assessment which evaluates students' performance in mathematics and science at different grade levels across several countries. A large number of features pertaining to students, teachers, and schools are recorded. We aim to quantify the impact of a small number covariates of primary interpretational interest (i.e., parents' education level, minutes spent on homework, and school discipline problems) on students' performance in mathematics, in the presence of other covariates of non-primary interest.

In the previously proposed SSP-BART, the design matrix is split into two disjoint subsets $\mathbf{X}_{1}$ and $\mathbf{X}_{2}$, which contain covariates of primary and non-primary interest, respectively. The specification of these matrices should be guided by the application at hand. The covariates in $\mathbf{X}_1$ are of interest in terms of being interpretable, but their impact on the response is also relevant. The covariates of non-primary interest in $\mathbf{X}_{2}$ may still be strongly related to the response, but are not considered important in terms of interpretation. The primary covariates in $\mathbf{X}_{1}$ are specified in a linear predictor and the others are exclusively used by BART; i.e., covariates in $\mathbf{X}_{2}$ are the only ones allowed to form interactions. SSP-BART applied to the TIMSS 2019 data would thus prohibit interactions between (or involving) the aforementioned primary covariates. This omission of important interactions represents a major limitation of SSP-BART, given that handling interactions automatically is supposedly part of its appeal.

Our work differs from SSP-BART in that i) we do not assume that $\mathbf{X}_{1}$ and $\mathbf{X}_{2}$ are disjoint; i.e., we allow $\mathbf{X}_{1} \cap \mathbf{X}_{2} \ne \emptyset$, or even $\mathbf{X}_{1} \subset \mathbf{X}_{2}$. This is important because primary and~non-primary covariates may also interact in complex ways and further impact the response. Unlike SSP-BART, our model's BART component accounts for this and yields better trees and notably improved predictive performance on the TIMSS 2019 data. Moreover, ii) we change the way the trees in BART are learned by introducing `double-grow' and `double-prune' moves, along with stricter checks on tree-structure validity, to resolve non-identifiability issues between the parametric (linear) and non-parametric (BART) components. Finally, iii) while \citet{zeldow2019semiparametric} assume that all parameters in the linear predictor have the same (diffuse) variance \emph{a priori}, we instead place a hyperprior on the full hyper-covariance matrix of the main effects, so that we are better able to model the correlations among them.

Thus, within the semi-parametric BART paradigm, we make a distinction between SSP-BART and our combined semi-parametric BART, which we call CSP-BART. In CSP-BART, we have made fundamental structural changes to the way that the trees are learned due to the fact that $\mathbf{X}_{1}$ and $\mathbf{X}_{2}$ can have covariates in common. Specifically, we prohibit the BART component from estimating marginal effects for variables in $\mathbf{X}_1$ in order to ensure that the parameter estimates in the linear component are identifiable. We also allow the specification of both fixed and random effects in the linear predictor, as in a linear mixed model, in which the parameter estimates can vary by a grouping factor. In contrast, interactions and non-linearities are handled by the BART component.

Beyond our proposed extensions to SSP-BART, another related work in this area is the varying coefficient BART \citep[VCBART;][]{deshpande2020vcbart}, which combines the idea of varying-coefficient models \citep{hastie1993varying} with BART and extends the work of \citet{hahn2020bayesian} to a framework with multiple covariates. In VCBART, the response is modelled via a linear predictor where the effect of each covariate is approximated by a BART model based on a set of modifiers (i.e., covariates that are not of primary interest). The only similarity between VCBART and CSP-BART is the use of a linear predictor along with BART. However, our work is structurally different as we do not estimate the parameters in the linear predictor via BART. Instead, they are obtained in the same fashion as a Bayesian linear mixed model approach, so as to yield interpretable and unbiased coefficient estimates. 

Another extension of SSP-BART is provided by the model of \citet{dorie2022stan}, which accommodates not only fixed effects but also random effects in its parametric component.\linebreak{}Unlike SSP-BART, their approach integrates BART with Stan \citep{stan} in the associated \textsf{R} package \texttt{stan4bart}, so that the fixed and random effects are updated via the No-U-Turn sampler rather than Gibbs steps. Without random effects, this is the only aspect of their approach which differs from SSP-BART. Crucially, \citet{dorie2022stan} also operate under the assumption of mutual-exclusivity between $\mathbf{X}_1$ and $\mathbf{X}_2$. As we do not make use of random effects in either our simulation experiments or our applications, we do not compare the results of CSP-BART with those of \texttt{stan4bart}, although we do describe a version of CSP-BART which also allows random effects in Section \ref{random_effect_sec}. Instead, we show using standard performance metrics that SSP-BART and VCBART compare unfavourably to CSP-BART in two simulation studies (where we demonstrate CSP-BART's ability to recover the true main effects in either the presence or absence of interactions), our primary analysis of the TIMSS 2019 data, and an additional application to the well-known Pima Indians Diabetes dataset \citep{uci_repo} (presented in Appendix \ref{appendix_pima} to demonstrate the practical use of CSP-BART in classification rather than regression settings).

The remainder of this paper is organised as follows. Section \ref{sec_TIMSS} provides an overview of the TIMSS 2019 data and discusses why our model is particularly well-suited for it. In Section \ref{sec_BART}, we summarise the BART model and introduce relevant notation. In Section \ref{semi_BART_sec}, we revise the separated semi-parametric BART model and describe our proposed CSP-BART extension in detail. In Section \ref{simulation_sec}, we compare the performance of CSP-BART with other relevant algorithms on synthetic data. We analyse the TIMSS 2019 dataset in Section \ref{real_data_sec}. To conclude, we present a discussion in Section \ref{conclusion_sec}. Finally, additional details on the theoretical underpinnings of CSP-BART and additional results on the TIMSS 2019 data are provided in the Appendices.

\section{Trends in International Mathematics and Science Study (TIMSS)}
\label{sec_TIMSS}
The Trends in International Mathematics and Science Study (TIMSS) is a large international evaluation programme which monitors the trends in students' achievement in science and mathematics. Since 1995, the assessment takes place every 4 years in more than 60 countries and serves as a valuable source of data to compare students' performance at the fourth and eighth grade levels of secondary schools \citep{mullis2020timss}. For the eighth grade students, the TIMSS 2019 data consisted of four components: i) mathematics and science assessments for students, either paper- or computed-based, ii) a background questionnaire for students, iii) a questionnaire for teachers, and iv) a school questionnaire for principals. The first component contains mathematics- and science-related questions, and are exclusively taken by students. In 2019, for the first time, half of the countries opted for an onscreen assessment, and more countries are planning to offer it digitally over the next cycles. The students were also asked to fill in a brief background questionnaire on subjects such as their family, school, classroom environment, and self-confidence in learning mathematics and science. The teachers and principals also responded to a set of questions on their teaching experience, practices, educational background, and learning resources available at the schools, such as computers, access to internet, library, etc. The main goal of TIMSS is to use all these sources of information to improve learning and teaching practices \citep{timss2019}.

Due to its international reach and the availability of public use versions of the databases, TIMSS data have been studied since its first cycle in 1995. The first works date back to~\citet{beaton1996mathematics}, who analysed students' performance in mathematics by different content areas, such as fractions and number sense, algebra, and probability. In addition, \citet{stigler1997understanding} compared mathematics teaching practices in Germany, Japan, and the U.S., while \citet{stigler2000using} studied some of the challenges involved in analysing cross-cultural data. In recent years, approaches based on linear mixed-effects models have often been used to investigate important topics \citep{grilli2016exploiting,tang2022impact, chen2022effects}, such as the impact of teachers' professional development on students' performance and the association between individual- and class-level achievement and students' self-confidence in mathematics. Though these models offer more flexibility than their counterparts with no mixed-effects, they still require the specification of main effects and interactions, which are assumed to have a linear relationship with the response. In practice, it can be challenging to fully specify the interaction effects without deep knowledge of the problem at hand. Moreover, this is naturally even more difficult in high-dimensional settings. Our approach, however, addresses the interaction effects through a non-parametric form which does not require their specification nor impose/assume any linear relation with the response.

\subsection{Data description}

We consider TIMSS 2019 data from Ireland pertaining to students at the eighth grade level. In total, the data provide details on the achievement of $4{,}118$ participating students, as well as information on their personal backgrounds, their teachers, and the schools they attend. As the focus of our work is on evaluating students’ performance in mathematics, we dropped variables related to science achievement. After merging the individual student data with that of their teachers and school, the resulting dataset provides us with a total of $270$ predictors. However, we initially do some data manipulation to remove predictors with a high level of missing information in order to ensure a fair comparison with as much data as possible between CSP-BART and its competitors in Section \ref{section_timss2019}, since the implementations of BCF, SSP-BART, and VCBART cannot deal with missing data. We adopt a BART-based variable-screening step which yields a dataset with $20$ predictors and $3{,}224$ observations (greater than the $1{,}448$ complete cases). The exact nature of this data manipulation is explained in detail in Section \ref{real_data_sec} and additional rationale for this procedure is provided in Appendix \ref{appendix_timss_screening}. Furthermore, we present additional results (for CSP-BART only) in Appendix \ref{appendix_csp_missing}, considering scenarios which make use of more predictors, both by analysing only the complete cases and by adapting CSP-BART to accommodate missing values in its trees.

In our analyses, we aim to quantify how a small group of covariates of primary interpretational interest impact students' performance in mathematics while automatically accounting for non-linearities and interactions involving a larger group of covariates, which are of non-primary interpretational interest. We chose the covariates of primary interest based on previous works which suggest that family background, time spent on academic activities, and school-level factors affect students' mathematics scores \citep{martin2000effective, mohammadpour2015multilevel, grilli2016exploiting}. Notably, we have \emph{primary} interpretational interest in quantifying the effect of `parents’ education level', `minutes spent on homework', and `school discipline problems'. As covariates of \emph{non-primary} interest, we consider predictors such as `gender of the student', `how often student is absent from school', `how often student feels hungry', `does the student have a tablet or a computer at home?', among others; see Table \ref{TIMSS_covs} in Appendix \ref{appendix_timss_dataset} for the list of the $20$ variables identified by the BART-based variable-screening step and Table \ref{tab_descriptive_stats_TIMSS} for descriptive statistics of the covariates of primary interest and some specifically student-related covariates of non-primary interest. We point out that the latter covariates are not of \emph{interpretational} interest, though they are important to predict students' performance in mathematics as they might form complex interactions, either among themselves or with those of primary interest, or have marginal linear/non-linear relationships with the response.\vspace{-\smallskipamount}
\begin{table}[H]%
\caption{Descriptive statistics of seven of the $20$ pre-selected covariates from the TIMSS 2019 data for Irish students at the eighth-grade level, based on the dataset with $3{,}224$ observations obtained after the BART-based variable-screening step. Predictors of primary interpretational interest are denoted with an asterisk ($\star$). A complete list of the $20$ covariates identified by the BART-based variable-screening step is provided in Table \ref{TIMSS_covs}.\label{tab_descriptive_stats_TIMSS}}\vskip\smallskipamount
\centering
\setlength{\tabcolsep}{4.125pt}
\extrarowheight 2.25pt
\begin{tabular*}{400pt}{lll}
\toprule
\textbf{Covariate} & \textbf{Category} & $\mathbf{n}$ \textbf{(percent)}  \\
\midrule

\multirow{6}{*}{Parents' education level ($\star$)} & University or higher & $1,078~(0.33)$  \\ 
& Post-secondary but not university & $666~(0.21)$ \\ 
& Upper secondary & $434~(0.13)$ \\ 
& Lower secondary & $112~(0.03)$ \\ 
& Primary, secondary, or no school & $59~(0.02)$ \\ 
& Do not know & $875~(0.27)$ \\
\hline
\multirow{6}{*}{Minutes spent on homework ($\star$)} & No homework & $27~(0.01)$ \\ 
& $1$ to $15$ minutes  & $1,065~(0.33)$ \\
& $16$ to $30$ minutes  & $1,469~(0.46)$ \\
& $31$ to $60$ minutes  & $541~(0.17)$ \\
& $61$ to $90$ minutes  & $79~(0.02)$ \\
& More than $90$ minutes  & $43~(0.01)$ \\
\hline

\multirow{3}{*}{School discipline problems ($\star$)} & Hardly any problems & $2,011~(0.62)$ \\
& Minor problems & $1{,}133~(0.35)$ \\
& Moderate to severe problems & $80~(0.02)$ \\
\hline
\multirow{2}{*}{Gender of the student} & Female & $1{,}636~(0.51)$ \\
& Male & $1{,}588~(0.49)$ \\
\hline
\multirow{6}{*}{How often student is absent from school} & Once a week & $89~(0.03)$ \\
& Once every two weeks & $283~(0.09)$ \\
& Once a month & $552~(0.17)$ \\
& Once every two month & $781~(0.24)$ \\
& Never or almost never & $1{,}519~(0.47)$ \\
\hline
\multirow{5}{*}{How often student feels hungry} & Every day & $293 (0.09)$ \\
& Almost every day & $412 (0.13)$ \\
& Sometimes & $1{,}383 (0.43)$ \\
& Never & $1{,}136 (0.35)$ \\
\hline
\multirow{3}{*}{Tablet or a computer at home} & Yes & $3{,}161 (0.98)$ \\
& No & $63 (0.02)$ \\
\bottomrule
\end{tabular*}
\end{table}
In general, data from TIMSS questionnaires tend to be analysed using parametric models, such as linear mixed-effects models \citep{lme4}. For instance, \citet{grilli2016exploiting} analysed the Italian data from TIMSS and the Progress in International Reading Literacy Study (PIRLS). They performed a multivariate analysis considering reading, mathematics, and science scores via a multivariate linear mixed-effects model. In addition, \citet{mohammadpour2015multilevel} analysed the science scores of the TIMSS 2007 data for eighth-grade students for 29 countries also using mixed-effects models. However, both works explored marginal effects of some student- and school-level predictors only, thus missing relevant complex interactions and non-linear effects. More recently, \citet{mcjames2023bayesian} modelled the mathematics and science scores of the TIMSS 2019 data by proposing a multivariate extension to Bayesian causal forests \citep[BCF;][]{hahn2020bayesian}. Though methods based on Bayesian forests offer great flexibility since they can deal with interaction effects without requiring pre-specification, their method, as per the vanilla BCF, is able to model the effect of one binary covariate of primary interpretational interest only. That is, if there is interest in analysing more than one covariate of interest, a continuous covariate, or a categorical covariate with more than 2 levels, their multivariate BCF cannot be used.

Unlike linear mixed-effects and BCF models, our approach is particularly well-suited to the TIMSS 2019 data, since here we are concerned not only with prediction, but also interpretability. First, similar in spirit to linear mixed-effects models, it allows the estimation of effects of multiple covariates of primary interest of any kind. Second, complex interactions and non-linearities involving covariates of non-primary interpretational interest require no pre-specification. In practice, predictors of non-primary interest tend to outnumber those of primary interest in high-dimensional settings and CSP-BART deals with this non-parametrically through its BART component. Compared to the work of \citet{zeldow2019semiparametric}, we allow for more interactions by \emph{not} assuming that the set of primary and non-primary covariates are mutually exclusive; in practice, there is no reason as to why such interactions should be prevented. In this sense, CSP-BART can be seen as the best of both worlds as it has a linear predictor where covariates of primary interest are specified and accounts for complex interactions and non-linear effects through a Bayesian non-parametric model.

\section{BART}\label{sec_BART}

BART \citep{chipman2010bart} is a Bayesian statistical model based on an ensemble of trees that was first proposed in the context of regression and classification problems. Through an iterative Bayesian backfitting MCMC algorithm, BART sequentially generates a set of trees that, when summed together, return predicted values. A branching process prior is placed on the tree structure to control the depth of the trees. In addition, the covariates and split-points used to define the tree structure (i.e., splitting rules) are randomly selected without the optimisation of a loss function, such as in random forests \citep{breiman2001random} and gradient boosting \citep{friedman2001greedy}. Compared to regression models, BART is more flexible in the sense that it does not assume linearity between the covariates and the response and does not require specification of a linear predictor. In particular, BART automatically determines non-linear marginal effects and multi-way interaction effects.

BART has been used and extended to different applications, and its theoretical properties have also gained attention more recently. For instance, BART has been applied to credit risk modelling \citep{zhang2010bayesian}, survival/competing analysis \citep{sparapani2016nonparametric, sparapani2020nonparametric, linero2021bayesian}, biomarker discovery \citep{hernandez2015bayesian}, plant-based genetics \citep{sarti2023bayesian}, and causal inference \citep{hill2011bayesian, green2012modeling,hahn2020bayesian}. Furthermore, it has also been extended to high-dimensional data \citep{linero2018bayesian, hernandez2018bayesian}, polychotomous responses \citep{kindo2016multinomial, murray2021log}, zero-inflated and semi-continuous data \citep{linero2020semipar, murray2021log}, heteroscedastic data \citep{pratola2017heteroscedastic}, and to estimate linear, smooth, and monotone functions \citep{starling2020bart, prado2021bayesian, chipman2022mbart}. Regarding theoretical developments, we highlight the works of \citet{lineroAnDyang2018}, \citet{rockova2019theory}, and \citet{rockova2020posterior}, who provide results related to the convergence of the posterior distribution generated by the BART model. Finally, we note that BART has also been previously employed in education assessment settings \citep{suk2021hybridizing}, similar to the TIMSS application we analyse herein.

The standard BART model approximates a univariate response $\{y_{i}\}_{i=1}^n$ by a sum of trees,
\[y_{i}\given\mathbf{x}_{i}, \bm{\mathcal{M}}, \bm{\mathcal{T}}, \sigma^{2}   \sim \mbox{N}\left(\sum_{t = 1}^{T} g\left(\mathbf{x}_{i}, \bm{\mathcal{M}}_{t}, \mathcal{T}_{t}\right), \sigma^{2} \right),\]
where $\mbox{N}(\cdot)$ denotes the Normal distribution, $\sigma^{2}$ is the error variance, $g(\cdot) = \mu_{t\ell}$ is a function which assigns predicted values $\mu_{t\ell}$ to all observations falling into terminal node $\ell$ of tree $t$, $\mathbf{x}_{i}$ denotes the $i$-th row of the design matrix $\mathbf{X}$, $\mathcal{T}_{t}$ represents the topology of tree $t$, and $\bm{\mathcal{M}}_{t} = \left(\mu_{t1}, \ldots, \mu_{t b_{t}} \right)$ is a vector comprising the predicted values from the $b_{t}$ terminal nodes of tree $t$. For notational convenience, we let $\bm{\mathcal{T}} = (\mathcal{T}_{1}, \ldots, \mathcal{T}_{T})$ and $\bm{\mathcal{M}} = (\bm{\mathcal{M}}_{1}, \ldots, \bm{\mathcal{M}}_{T})$ denote the sets of all trees and all predicted values, respectively. Regarding the number of trees $T$, \citet{chipman2010bart} recommend $T = 200$ as a default, though they suggest that $T$ can also be selected by cross-validation, depending on the application.

Unlike other tree-based algorithms where a loss function is minimised to define the splitting rules in the growing process, in BART the splitting rules are uniformly defined (i.e., the covariates and their split-points are selected at random based on a uniform distribution). In addition, the BART model learns the structure of the trees by greedy modifications consisting of four moves: grow, prune, change, and swap (see Figure \ref{BCART}). For instance, in the grow move, a terminal node is randomly selected and two new terminal nodes are created below it. During a prune move, a parent of two terminal nodes is picked at random and its children are removed. In the change move, an internal node is randomly selected and its splitting rule is changed. Finally, in the swap move, the splitting rules associated with a pair of parent-child internal nodes are exchanged, with the pair being selected at random. See \citet{JSSv070i04} for further details on these tree proposal moves.

As a Bayesian model, BART places priors on the parameters of interest, assuming that $\sigma^{2} \sim \mbox{IG}(\nu/2, \nu\lambda/2)$ and $\mu_{t\ell} \sim \mbox{N} (0, \sigma_{\mu}^{2})$, where $\mbox{IG}(\cdot)$ represents the inverse gamma distribution and $\sigma_{\mu} = 0.5/(k\sqrt{T})$, with $k \in \lbrack1,3\rbrack$ such that each terminal node in each tree contributes only a small amount to the overall fit. In addition, a branching process prior is considered to control the depth of the trees. With this prior, each internal node $\ell^\prime$ is observed at depth $d_{t{\ell^\prime}}$ with probability $\eta(1+d_{t{\ell^\prime}})^{-\zeta}$, where $\eta \in (0,1)$ and $\zeta \ge 0$. \citet{chipman2010bart} recommend $\eta=2$ and $\zeta = 0.95$, which tends to favour shallow trees. See Appendix \ref{appendix_BART} for further details.

Fitting and inference for BART models is accomplished via MCMC. It is common to begin with all trees set as stumps. While they are stumps, the only possible move is grow. Thereafter, each tree is updated in turn by proposing a potential grow, prune, change, or swap move, whereby the type of move is chosen with probabilities 0.25, 0.25, 0.4, and 0.1, respectively. Each modified tree is compared to its previous version considering the partial residuals $\mathbf{R}_{t} = \mathbf{y} - \sum_{j \ne t}^{T} g\left(\mathbf{X}, \bm{\mathcal{M}}_{j}, \mathcal{T}_{j}\right)$ and the structure of both trees via a marginal likelihood calculation. This comparison is carried out via a Metropolis-Hastings (MH) step, and it is needed to select only splitting rules that improve the final prediction, since they are chosen based on a uniform distribution. Hence, all node-level parameters ($\mu_{t \ell}$) are generated. After doing this for all $T$ trees, the error variance ($\sigma^{2}$) is updated from its full conditional distribution. This entire scheme is then iteratively repeated; see Appendix \ref{appendix_BART} for further details on the BART implementation. The BART algorithm is practically implemented in the \textsf{R} packages \texttt{bartMachine} \citep{JSSv070i04}, \texttt{dbarts} \citep{dbarts}, and \texttt{BART} \citep{BART_Rpkg}.
\begin{figure}[H]
  \captionsetup[subfigure]{font=normalsize}
        \centering
        \begin{subfigure}[t]{195pt}% approx 0.475 * \textwidth
            \centering
            \caption[]%
            {$\mathcal{T}^{(1)}_{1}$}
            \begin{forest}
for tree={
    grow=south, draw, minimum size=3ex, 
    inner sep=3pt, % control the overall tree size
    s sep=7mm,
    l sep=6mm
    }
[$\:x_{2} \le 1\:$,
    [$\:x_{1} \le 0.5\:$, edge label={node[midway, font=\footnotesize, left]{$\mbox{ TRUE }$}}
        [$\mu_{11}$, circle,  edge label={node[midway,left, font=\footnotesize]{$\mbox{ TRUE }$}},]
        [$\mu_{12}$, circle,  edge label={node[midway,right, font=\footnotesize]{$\mbox{ FALSE }$}},]
    ]
    [$\mu_{13}$, circle,  edge label={node[midway,right, font=\footnotesize]{$\mbox{ FALSE }$}},]
]
\end{forest}
            \label{fig:mean and std of net14}
        \end{subfigure}
        \hfill
        \begin{subfigure}[t]{195pt}  
            \centering 
            \caption[]%
            {$\mathcal{T}^{(2)}_{1}$}
\begin{forest}
for tree={
    grow=south, draw, minimum size=3ex, 
    inner sep=3pt, % control the overall tree size
    s sep=7mm,
    l sep=6mm
        }
[$\:x_{2} \le 1\:$,
    [$\:x_{1} \le 0.5\:$,
        [$\mu_{11}$, circle]
        [$\mu_{12}$, circle]
    ]
        [$\:x_{3} \le 2\:$,
        [$\mu_{13}$, circle]
        [$\mu_{14}$, circle]
    ]
]
\end{forest}
            \label{fig:mean and std of net24}
        \end{subfigure}
        \vskip\medskipamount
        \begin{subfigure}[t]{195pt}   
            \centering 
                        \caption[]%
            {$\mathcal{T}^{(3)}_{1}$}    
\begin{forest}
for tree={
    grow=south, draw, minimum size=3ex, inner sep=3pt, s sep=7mm, l sep=6mm
        }
[$\:x_{2} \le 1\:$,
    [$\:x_{1} \le 0.5\:$,
        [$\mu_{11}$, circle]
        [$\mu_{12}$, circle]
    ]
    [$\:x_{4} \le 0.75\:$,
        [$\mu_{13}$, circle]
        [$\mu_{14}$, circle]
    ]
]
\end{forest}
            \label{fig:mean and std of net34}
        \end{subfigure}
        \hfill
        \begin{subfigure}[t]{195pt}   
            \centering 
            \caption[]%
{$\mathcal{T}^{(4)}_{1}$}  
            \begin{forest}
for tree={
    grow=south, draw, minimum size=3ex, inner sep=3pt, s sep=7mm, l sep=6mm
        }
[$\:x_{1} \le 0.5\:$,
    [$\:x_{2} \le 1\:$,
        [$\mu_{11}$, circle]
        [$\mu_{12}$, circle]
    ]
    [$\:x_{4} \le 0.75\:$,
        [$\mu_{13}$, circle]
        [$\mu_{14}$, circle]
    ]
]
\end{forest}
\end{subfigure}
\vskip\smallskipamount
\caption{An example of a tree generated from BART in $4$ different instances. In principle, BART does not generate only one tree but rather a set of trees which, summed together, are responsible for the final prediction. As indicated in panel (a), observations are pushed to the left child node when the splitting criterion is satisfied. The tree is represented as $\mathcal{T}^{(r)}_{1}$, where $r = \{1,2,3,4\}$ denotes the number of the iteration in which the tree is updated. The splitting rules (covariates and their split-points) are presented inside the rectangles which represent the internal nodes. The predicted values $\mu_{t \ell}$ are shown inside the circles which correspond to the terminal nodes. $\mathcal{T}^{(1)}_{1}$ illustrates the tree at iteration one with two internal nodes and three terminal nodes. From $\mathcal{T}^{(1)}_{1}$ to $\mathcal{T}^{(2)}_{1}$, the grow move is illustrated, as $\mu_{13}$ in $\mathcal{T}^{(1)}_{1}$ is split into $\mu_{13}$ and $\mu_{14}$ in $\mathcal{T}^{(2)}_{1}$ by using $x_{3} \le 2$. In addition, the prune move can be seen when $\mathcal{T}^{(2)}_{1}$ reverts to $\mathcal{T}^{(1)}_{1}$. The change move is shown when comparing $\mathcal{T}^{(2)}_{1}$ and $\mathcal{T}^{(3)}_{1}$, as the splitting rule that defines $\mu_{13}$ and $\mu_{14}$ is changed from $x_{3} \le 2$ to $x_{4} \le 0.75$. Finally, the swap move is illustrated in the comparison of $\mathcal{T}^{(3)}_{1}$ and $\mathcal{T}^{(4)}_{1}$.\label{BCART}}
\end{figure}
% ------------------------------------------------------------------------------------------------------------------------------------------------------------------------------------------------------------------------------------------

\section{Semi-parametric BART}\label{semi_BART_sec}

The BART model above does not provide an easy way to quantify the effects of covariates on the response as in regression models, which is often the main goal in many applications. The semi-parametric BART framework aims to overcome this by adding a parametric linear component to the additive ensemble of non-parametric trees. We note that linear predictors and BART have also been previously combined by \citet{prado2021bayesian}, albeit in a different way. There, linear predictors are used at the terminal node level of each tree, with a focus more on prediction accuracy than interpretability. In this Section, we first revise briefly the SSP-BART of \citet{zeldow2019semiparametric} in Section \ref{SSP_BART_sec}, then outline in detail our proposed extensions in the form of CSP-BART in Section \ref{CSP_BART_sec}, and finally further extend CSP-BART in Section \ref{random_effect_sec} to also allow random effects in the parametric component as per \citet{dorie2022stan}.

\subsection{Separated semi-parametric BART}\label{SSP_BART_sec}

In the separated semi-parametric BART proposed by \citet{zeldow2019semiparametric}, the design matrix $\mathbf{X}$ is split into two subsets, $\mathbf{X}_1$ and $\mathbf{X}_2$, with $p_1$ and $p_2$ columns, respectively. The matrix $\mathbf{X}_{1}$ contains covariates that should be included in a linear component to quantify the main effects and the $\mathbf{X}_{2}$ matrix contains covariates that might contribute to predicting the response but are not of primary interest. The linear predictor inside the BART framework is written as follows:
\begin{equation}
\label{semi_BART_model}
  y_{i}\given\mathbf{x}_{1i}, \mathbf{x}_{2i}, \boldsymbol\beta, \bm{\mathcal{M}}, \bm{\mathcal{T}}, \sigma^{2}  \sim \mbox{N}\left( \mathbf{x}_{1i} \boldsymbol\beta + \sum_{t = 1}^{T} g\left(\mathbf{x}_{2i}, \bm{\mathcal{M}}_{t}, \mathcal{T}_{t}\right), \sigma^{2} \right).  
\end{equation}
Furthermore, $\mathbf{X}_{1}$ and $\mathbf{X}_{2}$ are assumed to be mutually exclusive, such that $\mathbf{X}_{1} \cap \mathbf{X}_{2} = \emptyset$. Such a model with $\mathbf{X}_1=\emptyset$ is equivalent to the standard non-parametric BART, while such a model with $\mathbf{X}_2=\emptyset$ implies a fully parametric linear regression. In this semi-parametric setting, it is assumed that $p_2$ is large enough to ensure that BART is a reasonable model and that there are relatively few columns in $\mathbf{X}_1$; i.e., $p_1 \ll p_2$, typically. As above, the ensemble of trees used by the BART component is learned by the standard grow, prune, change, and swap moves.

The priors $\boldsymbol\beta \sim \mbox{MVN}(\mathbf{0}_{p_1}, \sigma^{2}_{b}\mathbf{I}_{p_1})$ and $\sigma^{2} \sim \mbox{IG}(\nu/2, \nu\lambda/2)$ are assumed for the linear reg\-ression coefficients and error variance, respectively, where $\mbox{MVN}(\cdot)$ represents the multivariate normal distribution, $\mathbf{0}_{p_1}$ and $\mathbf{I}_{p_1}$ respectively denote a $p_1$-dimensional vector of zeros and identity matrix, and $\nu$, $\lambda$, and $\sigma^2_b$ are user-specified hyperparameters. Typically, $\sigma^2_b$ is set large enough so that the prior on $\boldsymbol\beta$ is diffuse. Notably, the isotropic covariance structure $\sigma^{2}_{b}\mathbf{I}_{p_1}$ assumed by \citet{zeldow2019semiparametric} implies that all covariates in $\mathbf{X}_{1}$~have the same magnitude, which can easily be accomplished by appropriate transformations, and that covariate effects in $\boldsymbol\beta$ are \emph{a priori} uncorrelated, which may be unrealistic for many applications.

\subsection{Combined semi-parametric BART}\label{CSP_BART_sec}

In CSP-BART, we similarly allow for modelling covariates of primary and non-primary interest. Unlike SSP-BART, however, we consider that $\mathbf{X}_{1}$ and $\mathbf{X}_{2}$ may have covariates in common. This change is crucial as it allows primary covariates to interact both among themselves and with those in $\mathbf{X}_{2}$. Moreover, we change the tree-generation process in BART by introducing `double-grow' and `double-prune' moves to account for non-identifiability issues that may arise between the estimates from the linear and BART components. In CSP-BART, a univariate response $y_{i}$ is modelled in accordance with Equation \eqref{semi_BART_model}, along with the following prior distributions:
\begin{align*}
\boldsymbol\beta &\sim \mbox{MVN}\left(\mathbf{b}, \boldsymbol\Omega_{\beta}\right),\\
\boldsymbol\Omega_{\beta} &\sim \mbox{IW}\left(\mathbf{V}, v\right),\\
\sigma^{2} &\sim \mbox{IG}\left(\nu/2, \nu\lambda/2\right),
\end{align*}
where $\mbox{IW}(\cdot)$ represents the inverse Wishart distribution. We specify $\mathbf{V}=\mathbf{I}_{p_1}$ and $v=p_1$, while $\nu=3$ and $\lambda$ are set following \citet{chipman2010bart}. While the~prior on $\boldsymbol\beta$ used by SSP-BART assumes that the linear regression coefficients are uncorrelated and equivariant, this assumption is sensible only when the covariates in $\mathbf{X}_1$ have been standardised appropriately. Conversely, our hierarchical prior on $\boldsymbol\beta$ allows us to explicitly model correlation among the covariate effects in $\boldsymbol\beta$ (see Section \ref{random_effect_sec}). As an aside, numeric covariates in $\mathbf{X}_2$ need not be standardised under either CSP-BART or SSP-BART, as the splitting rules in BART are invariant under monotone transformations. Following \citet{chipman2010bart, linero2018bayesian}, we recommend transforming only the response to lie between $-0.5$ and $0.5$ to facilitate specification of the prior on $\mu_{t \ell}$ and improve numerical stability.

To allow for $\mathbf{X}_{1}$ and $\mathbf{X}_{2}$ sharing covariates, we propose to change the moves of the BART model in order to resolve non-identifiability issues between the linear component and BART. Thus, if $\mathbf{X}_{1} \cap \mathbf{X}_{2} \ne \emptyset$, we propose a `double-grow' move only when $x \in \{\mathbf{X}_{1} \cap \mathbf{X}_{2}\}$ is chosen to define a splitting rule for a stump. We therefore stress that the double-grow move applies \emph{only} when growing a stump where the splitting rule is based on $x \in \{\mathbf{X}_{1} \cap \mathbf{X}_{2}\}$. In cases where the tree is not a stump or $x \notin \mathbf{X}_1$, only the `single-grow' move applies. Furthermore, we point out that the double-grow move consists of two operations which must be performed simultaneously: i) proposing a second splitting rule using any variable, except the one used at the root node, and ii) shrinking the $\mu_{t\ell}$ parameter of the terminal node on the opposing branch of the initial split by modifying its prior. Below, we first illustrate part i) of the double-grow move and then state why part ii) is needed to prevent non-identifiability issues.

For example, if $\mathcal{T}_{t}$ is a stump and $x_{1} \in \{\mathbf{X}_{1} \cap \mathbf{X}_{2}\}$ is randomly chosen to define a splitting rule, then another covariate, e.g., $x_{2}$ --- which cannot be the same variable used to split at the root but otherwise can belong either to $\mathbf{X}_{1}$, $\mathbf{X}_{2}$, or $\mathbf{X}_{1} \cap \mathbf{X}_{2}$ --- will also be (randomly) chosen and the proposed tree will contain both $x_{1}$ and $x_2$. When double-growing a tree, the branch on which the second split is proposed is sampled with equal probability. If $\mathcal{T}_{t}$ is a stump and $x_{1} \notin \{\mathbf{X}_{1} \cap \mathbf{X}_{2}\}$ is chosen to define a splitting rule, a standard `single-grow' move is employed. The rationale behind double-growing is thus to induce interactions between covariates in $\mathbf{X}_1$ and others in either $\mathbf{X}_1$ or $\mathbf{X}_2$, and let only the linear component capture main effects associated with covariates in $\mathbf{X}_1$. With a single grow move, both components would eventually try to estimate the effects of covariates in $\mathbf{X}_1$ whenever $\mathbf{X}_{1}$ and $\mathbf{X}_{2}$ share at least one common covariate, which would lead to non-identifiability issues. However, the double-grow move ensures that the linear component estimates only main effects and forces the BART component to work specifically on interactions and non-linearities.

The `double-prune' move is proposed to prevent trees from containing only one covariate which belongs to $\mathbf{X}_{1} \cap \mathbf{X}_{2}$. To illustrate this move, we recall Figure \ref{BCART}. In panel (a), the tree has $3$ terminal nodes (circles) and $2$ internal nodes (rectangles). If the parent of the terminal nodes with parameters $\mu_{11}$ and $\mu_{12}$ is `single' pruned, the new tree structure will contain only $x_{2}$. If $x_{2} \notin \{\mathbf{X}_{1} \cap \mathbf{X}_{2}\}$, which implies that $x_{2} \in \mathbf{X}_{2}$, there will be no identifiability issues between the components in CSP-BART. However, if $x_{2} \in \{\mathbf{X}_{1} \cap \mathbf{X}_{2}\}$, the effect of $x_{2}$ will be estimated by both the linear predictor and BART. To avoid this issue, we prune the tree again. Thus, the result of a double-prune move will always be a stump. Like the double-grow move, it is vital to emphasise that the double-prune move is only accepted or rejected~via a MH step in its entirety; it is not possible to accept the first grow/prune and reject the second.

Despite these double moves, non-identifiability issues may still arise in two cases: a) when an intercept is specified in $\mathbf{X}_1$ and b) when any terminal node belongs to a branch whose splitting rules at each depth all involve only one covariate belonging to $\mathbf{X}_{1} \cap \mathbf{X}_{2}$. To avoid the first issue, we stress that $\mathbf{X}_{1}$ should not be equipped with a leading column of ones corresponding to an intercept. Doing so would conflate the linear component's constant with the constant node-level $\mu_{t\ell}$ parameters in the BART component. Accordingly, our removal of the intercept circumvents the need to impose the constraint $\mathbb{E}\left(\sum_{t=1}^Tg\left(\mathbf{x}_{2i},\bm{\mathcal{M}}_t,\mathcal{T}_t\right)\right) = 0$. The second issue is easily remedied by automatically rejecting proposed trees containing branches defined only by repeated splits on the same variable in $\mathbf{X}_1$; see Appendix \ref{appendix_semiBART} for details on CSP-BART's implementation. Given that this further prevents the BART component from estimating marginal effects associated with categorical variables of primary interest, it is especially pertinent for the TIMSS application where the covariates in $\mathbf{X}_1$ are all categorical.

% Despite these double moves, non-identifiability issues may still arise in three cases: i) when categorical variables with more than two levels in $\mathbf{X}_1$ are used to define splitting rules in the BART component, ii) when an intercept is specified in $\mathbf{X}_1$, and iii) when any terminal node is associated with splitting rules, at any depth, which all involve only one covariate belonging to $\mathbf{X}_{1} \cap \mathbf{X}_{2}$. The first issue is easily remedied by automatically rejecting proposed trees containing branches defined only by repeated splits on the same categorical variable in $\mathbf{X}_1$; see Appendix \ref{appendix_semiBART} for details on how CSP-BART's implementation handles categorical variables. Given that this further prevents the BART component from estimating marginal effects associated with categorical variables of primary interest, it is especially pertinent for the TIMSS application where the covariates in $\mathbf{X}_1$ are all categorical. To remedy the second issue, we stress that $\mathbf{X}_{1}$ should not be equipped with a leading column of ones corresponding to an intercept. Doing so would conflate the linear component's constant with the constant node-level $\mu_{t\ell}$ parameters in the BART component. Accordingly, our removal of the intercept circumvents the need to impose the constraint $\mathbb{E}\left(\sum_{t=1}^Tg\left(\mathbf{x}_{2i},\bm{\mathcal{M}}_t,\mathcal{T}_t\right)\right) = 0$.

To provide further details on the second issue, we recall Figure \ref{BCART} and assume that~$x_{2} \in \{\mathbf{X}_{1} \cap \mathbf{X}_{2}\}$. In panel (a), $\mathcal{T}_{1}^{(1)}$ represents a tree containing two predictors ($x_{2}$ and $x_{1}$), where $x_{1}$ can belong to either $\mathbf{X}_{1}$, $\mathbf{X}_{2}$, or $\mathbf{X}_{1} \cap \mathbf{X}_{2}$. For simplicity, imagine that $\mathcal{T}_{1}^{(1)}$ was generated by naively applying a single grow move twice\footnote{Notice that applying a single grow move twice to a stump is equivalent to part i) of the double-grow move only, though the `single-grow' move is not restricted to stumps.} to a stump, where $x_{2}$ and $x_{1}$ were randomly selected to create the splitting rules. The two left-most terminal nodes have $\mu_{11}$ and $\mu_{12}$ as predicted values, with splitting rules defined by $x_{1}$ and $x_{2}$. However, the right-most terminal node with predicted value $\mu_{13}$ has $x_{2}$ as its only ancestor along that branch, which causes~non-identifiability issues between the linear and BART components, since $x_{2} \in \{\mathbf{X}_{1} \cap \mathbf{X}_{2}\}$; see Appendix \ref{appendix_semiBART} for an example in the context of categorical predictors and binary indicators thereof. We avoid such issues by performing part ii) of the double-grow move, which modifies the prior on the relevant predicted value to $\mu_{t\ell} \sim \mbox{N}(0, \sigma^{2}_{\mu} \approx 0)$ and in turn shrinks the posterior predicted value towards zero. We therefore stress that the double-grow move cannot be thought of as simply applying a single grow move twice to a stump. Rather, it is a two-step procedure which induces interactions and guarantees that the trees in the BART component do not model effects already specified in the linear predictor. Finally, we note that the prior on the other terminal nodes ($\mu_{11}$ and $\mu_{12}$ in the present example) would remain unchanged.

% We avoid such issues by modifying the prior on the relevant predicted value to $\mu_{t\ell} \sim \mbox{N}(0, \sigma^{2}_{\mu} \approx 0)$, which in turn shrinks the posterior predicted value towards zero. The prior on the other terminal nodes ($\mu_{11}$ and $\mu_{12}$ in the present example) would remain unchanged.

Regarding the change and swap moves, we stress that they are kept intact as `single' moves in CSP-BART. Equivalent `double-change' and `double-swap' moves are not required to deal with non-identifiability issues that may arise between the linear and BART components. However, more stringent checks are placed on the validity of trees proposed by these moves. In particular, change and swap moves are iteratively proposed until a valid tree structure is found; i.e., one which ensures the parameters in the linear component are identifiable, with~a minimum number of observations in each terminal node. If a valid tree is not found in some small number of iterations, a stump is proposed instead. In the end, proposed trees are always accepted or rejected according to a Metropolis-Hastings step, as in the standard BART model. Further details on the stringent checks and how the novel double moves and other modifications described above lead to identifiable estimates of the coefficients in the linear component in Equation \eqref{semi_BART_model} are described in Appendix \ref{appendix_identifiability_csp}.%add \ref

Unlike the SSP-BART of \citet{zeldow2019semiparametric}, CSP-BART allows for~$\mathbf{X}_1 \cap\allowdisplaybreaks \mathbf{X}_2 \ne \emptyset$ %, which implies that $\mathbf{X}_1 \subset \mathbf{X}_2$, $\mathbf{X}_1 \supset \mathbf{X}_2$ or even $\mathbf{X}_1 = \mathbf{X}_2$. 
and thus accommodates a range of situations in which the sets of covariates are~\mbox{either} identical or nested ($\mathbf{X}_1 \subset \mathbf{X}_2$ or $\mathbf{X}_1 \supset \mathbf{X}_2$), while also allowing non-overlapping sets as per SSP-BART as a special case. The specification of the predictors of primary interpretational interest ($\mathbf{X}_{1}$) depends on the problem at hand, similar in spirit to the specification of linear predictors in GLMs and SSP-BART. For example, a user of SSP-BART needs to specify $\mathbf{X}_1$ correctly just as much as a user of CSP-BART (i.e., for both models it is assumed that the user knows beforehand which effects are of primary interest). In parallel, the set of variables of non-primary interpretational interest ($\mathbf{X}_{2}$) should contain all predictors that can form complex, non-linear interactions. If the practitioner has $\mathbf{X}_1$ set up but is not sure which predictors should be included in $\mathbf{X}_2$, they could consider adding all available predictors\footnote{We advise adding all available predictors to $\mathbf{X}_2$, but caution against adding covariates to $\mathbf{X}_2$ which are highly correlated among themselves, for reasons of parsimony, or with those in $\mathbf{X}_1$, to avoid biasing the $\boldsymbol{\beta}$ estimates.} to $\mathbf{X}_2$ (including those in $\mathbf{X}_1$) to ensure they allow for any possible interactions to be estimated. In contrast, if the user wants to prevent any interactions between predictors of primary and non-primary interest or among predictors of primary interest, they can set up $\mathbf{X}_2$ so that $\mathbf{X}_1 \cap \mathbf{X}_2 \ne \emptyset$.

Equations \eqref{update_beta}--\eqref{update_sigma_semi} below present the respective full conditional distributions for $\boldsymbol\beta$, $\boldsymbol\Omega_\beta$, and $\sigma^2$. These expressions are needed due to the inclusion of the linear predictor in the CSP-BART model; see Appendix \ref{appendix_semiBART} for full details on the CSP-BART implementation. An outline algorithm for the process is given~by:
\begin{enumerate}[label=\roman*)]
    \item \label{step1} Update the linear predictor, with $\mathbf{r} = \mathbf{y} - \sum_{t=1}^{T} g\left(\mathbf{X}_{2}, \bm{\mathcal{M}}_{t}, \mathcal{T}_{t}\right)$, via
\begin{align}
\label{update_beta}
    \boldsymbol\beta \given \mathbf{X}_{1}, \mathbf{r}, \sigma^{2}, \mathbf{b}, \boldsymbol\Omega_{\beta} & \sim \mbox{MVN}\left( \mu_{\beta} = \Sigma_{\beta} \left( \sigma^{-2} \mathbf{X}_{1}^{\top} \mathbf{r} + \boldsymbol\Omega_{\beta}^{-1} \mathbf{b} \right), \right. \\[-1ex]
    & \phantom{\sim \mbox{MVN}\left(\,\right.}\left.\Sigma_{\beta} =  \left( \sigma^{-2} \mathbf{X}_{1}^{\top} \mathbf{X}_{1} + \boldsymbol\Omega_{\beta}^{-1}\right)^{-1} \right), \nonumber \\
    \label{update_omega}
    \boldsymbol\Omega_{\beta} \given \boldsymbol\beta, \mathbf{b}, \mathbf{V}, v & \sim \mbox{IW}\left( \left(\boldsymbol\beta - \mathbf{b}\right) \left(\boldsymbol\beta - \mathbf{b}\right)^{\top} + \mathbf{V}, v + 1 \right).
\end{align}

\item \label{step2} Then, sequentially update all $T$ trees, one at a time, via 
\[\mathbf{R}_{t} = \mathbf{y} - \mathbf{X}_{1}\boldsymbol\beta - \sum_{j \ne t}^{T} g\left(\mathbf{X}_{2}, \bm{\mathcal{M}}_{j}, \mathcal{T}_{j}\right).\]

\item \label{step3} Finally, update
\begin{align}
\label{update_sigma_semi}
   \sigma^{2} \sim \mbox{IG}\left(\frac{n + \nu}{2}, \frac{S + \nu \lambda}{2}\right),
\end{align}
where $S = \left(\mathbf{y} - \hat{\mathbf{y}}\right)^{\top} \left(\mathbf{y} - \hat{\mathbf{y}}\right)$ and $\hat{\mathbf{y}} = \mathbf{X}_{1} \boldsymbol\beta + \sum_{t = 1}^{T} g\left(\mathbf{X}_{2}, \bm{\mathcal{M}}_{t}, \mathcal{T}_{t}\right)$.
\end{enumerate}
In Step \ref{step1}, the linear predictor's parameter estimates and covariance matrix are updated, taking into account the difference between the response and the predictions from all trees. In Step \ref{step2}, each tree $t$ is modified considering the updated parameter estimates $\boldsymbol\beta$. Finally, the error variance is updated in Step \ref{step3}.

The main benefits of our approach are i) ease of implementation, relative to GLMs and GAMs, as we can model interactions and non-linearities without requiring pre-specification, ii) improved predictive performance relative to other tree-based methods, and iii) reduced bias relative to other semi-parametric BART models. In CSP-BART, practitioners do not need to iv) have prior knowledge of whether $\mathbf{X}_1 \cap \mathbf{X}_2 \ne \emptyset$, v) rely on previous studies to specify interaction effects in $\mathbf{X}_2$, vi) examine interaction plots to determine possible interactions to then specify $\mathbf{X}_2$, or vii) fit several BART-based models in order to compare them with CSP-BART. Regarding computational cost, viii) CSP-BART adds negligible time overhead to the standard BART model, especially if the number of columns in $\mathbf{X}_{1}$ is moderate. The computational cost of CSP-BART is also comparable to that of SSP-BART, as our novel double moves are not computationally intensive; see Appendix \ref{appendix_semiBART} for details on the comparison of the computational time of BART, SSP-BART, and CSP-BART.

Finally, we note that CSP-BART may not be suitable for analyses where the number of variables in the BART component is large, given the uniform sampling of the splitting rules. In such cases, instead of drawing the splitting variable uniformly as in BART, we recommend the DART model of \citet{linero2018bayesian} which uses a Dirichlet prior on the splitting probabilities, so that more important predictors can be favoured over those which have little or no influence on the response. In our implementation\footnote{Available at \url{https://github.com/ebprado/CSP-BART}.}, the `CSP-DART' model is available by specifying \texttt{sparse=TRUE}, though we retain the \texttt{sparse=FALSE} default in all applications herein (apart from the additional results on the TIMSS 2019 data presented in Appendix \ref{appendix_csp_missing}).

\subsection{Incorporating random effects in CSP-BART}\label{random_effect_sec}

Although we have introduced CSP-BART considering only fixed effects, it is straightforward to extend it to a setting with additional random effects, whereby covariates of primary interest are conditioned on categorical predictors. This yields

\begin{equation}
y_{i} \given \mathbf{x}_{1i}, \mathbf{z}_{i}, \mathbf{x}_{2i}, \boldsymbol\beta, \boldsymbol\gamma, \bm{\mathcal{M}}, \bm{\mathcal{T}}, \sigma^{2} \sim \mbox{N}\left( \mathbf{x}_{1i} \boldsymbol\beta + \mathbf{z}_{i} \boldsymbol\gamma + \sum_{t = 1}^{T} g\left(\mathbf{x}_{2i}, \bm{\mathcal{M}}_{t}, \mathcal{T}_{t}\right), \sigma^{2} \right),\label{eq:csp_random}
\end{equation}
where $\boldsymbol{\gamma}$ is the $q$-dimensional random effects vector with associated design matrix $\mathbf{Z}$. Conceptually, all effects are random under the Bayesian paradigm, but we use the terms `fixed' and `random' to distinguish between $\boldsymbol{\beta}$ and $\boldsymbol{\gamma}$ nonetheless. To fit such a model, we define $\boldsymbol\beta^\star=(\boldsymbol\beta,\boldsymbol\gamma)^\top$ and $\mathbf{x}_{1i}^\star=(\mathbf{x}_{1i},\mathbf{z}_{i})$. With $\boldsymbol\beta \sim \mbox{MVN}(\mathbf{b}, \boldsymbol\Omega_{\beta})$ as above, and a $\mbox{MVN}(\mathbf{0}_q, \boldsymbol\Omega_{\gamma})$ prior assumed for $\boldsymbol{\gamma}$, a block-diagonal covariance matrix $\boldsymbol\Omega_{\beta^{\star}}$ is obtained in the induced prior for $\boldsymbol\beta^\star$, which implies that $\boldsymbol\beta$ and $\boldsymbol\gamma$ are correlated among themselves but not with each other. We relax this assumption by letting $\mathbf{b}^{\star} = (\mathbf{b}, \mathbf{0}_{q})^\top$ and assuming $\boldsymbol\beta^{\star} \sim \mbox{MVN}(\mathbf{b}^{\star}, \boldsymbol\Omega_{\beta^{\star}})$, where now $\boldsymbol\Omega_{\beta^{\star}} \sim \mbox{IW}(\mathbf{V}^{\star}, v^{\star})$. Subject to
$\boldsymbol\beta = \boldsymbol\beta^{\star}$, $\mathbf{X}_{1} = \mathbf{X}^{\star}_{1}$, and $\boldsymbol\Omega_{\beta} = \boldsymbol\Omega_{\beta^{\star}}$, both prior settings allow direct application of the model-fitting algorithm outlined in Section \ref{CSP_BART_sec}. Notably, only $\boldsymbol\Omega_{\beta^{\star}}$ under the latter approach accounts for potential correlations between the fixed and random effects, while the isotropic prior employed in SSP-BART by \citet{zeldow2019semiparametric} would not. As ever, SSP-BART would also be unable to capture interactions involving random effects in $\mathbf{X}_1$ and other covariates of non-primary interest in $\mathbf{X}_2$.

In our implementation, we adapt the mixed-model formula syntax from the \texttt{lme4} \citep{lme4} package, so that the linear fixed and random effects can be easily specified through a formula (e.g., \texttt{y $\mathtt{\sim}$ 0 + x1 + (x2 | x3)}, where $\texttt{y}$ denotes a univariate response, \texttt{0} ensures that no intercept is included, $\texttt{x1}$ and $\texttt{x2}$ represent continuous covariates, and $\texttt{x3}$ is a factor with multiple levels; see Table 2 in \citet{lme4} for more examples). When specifying the linear predictor, the user needs only to supply the main fixed and random effects, as any interactions among covariates of primary interest are also determined automatically by BART. We note that polynomial effects, if any are of primary interpretational interest, should also be specified in the linear predictor only, as splitting rules based on $x_1$ or $x_1^3$, for example, would yield equivalent trees, and it would be necessary to avoid trees whose only splits involve both $x_1$ and monotonic transformations thereof.

The model in Equation \eqref{eq:csp_random} shares some similarities with the model of \citet{dorie2022stan} implemented in the \texttt{stan4bart} \textsf{R} package. Both models include a nonparametric BART component, allow both fixed and random effects in a parametric component, and~omit the global intercept. Though our model updates the parametric component via Gibbs steps rather than the No-U-Turn sampler, the more substantial differences between our model and \texttt{stan4bart} are analogous to the aforementioned differences between CSP-BART with only fixed effects and the SSP-BART model of \citet{zeldow2019semiparametric}. Specifically, \cite{dorie2022stan} advise that the parametric component (both fixed and random effects) should have no predictors in common with the BART component if interpretability of the regression coefficients is of interest. Their model thus suffers from similar limitations as SSP-BART in this regard. Conversely, the benefits of our novel double-grow and double-prune moves (originally motivated in the context of a linear predictor with only fixed effects) extend to settings where CSP-BART also has random effects in its parametric component.

In the context of CSP-BART with a multilevel structure, we need only additionally ensure that all branches across all trees in the BART component are not exclusively defined by splitting rules related only to the predictors involved in the specification of the random effects. For example, in the aforementioned formula \texttt{y $\mathtt{\sim}$ 0 + x1 + (x2 | x3)}, \texttt{x2} and \texttt{x3} could not define splitting rules in a given tree whenever the resulting terminal nodes would have only these two variables as ancestors, while partitions defined only by \texttt{x1} and \texttt{x3}, \texttt{x1} and \texttt{x3}, and \texttt{x1}, \texttt{x2} and \texttt{x3} would be allowed. However, we stress that we do not explicitly compare the results of CSP-BART with those of \texttt{stan4bart}; as we do not make use of random effects in either our simulation experiments or our applications, it suffices to compare CSP-BART with SSP-BART.

\section{Simulation experiments}\label{simulation_sec}

In this Section, we compare our novel CSP-BART with GAMs, SSP-BART, and VCBART in terms of bias (i.e., the difference between the posterior mean parameter estimates and the true parameter values) using three sets of synthetic data. The results were obtained using \textsf{R} \citep{R} version 4.11 and the \textsf{R} packages \texttt{mgcv} \citep{wood2017generalized}, \texttt{semibart} \citep{zeldow2019semiparametric}, and \texttt{VCBART} \citep{deshpande2020vcbart}. In addition to the standard SSP-BART, we also consider a modified version of SSP-BART throughout the simulations, which we denote by SSP-BART\textsuperscript{$\star$}, whereby the variables in $\mathbf{X}_1$ are shared with $\mathbf{X}_2$ (i.e., $\mathbf{X}_1 \cap \mathbf{X}_2 \ne \emptyset$), as per CSP-BART, but \emph{without} the CSP-BART's additional innovations (i.e., the novel double moves and stricter checks on tree-structure validity). Such a model can also be thought of as a version of CSP-BART which only employs the standard `single' moves. This can be achieved using the \texttt{semibart} package as it places no checks on the mutual-exclusivity of the two sets of predictors, in spite of this model assumption being crucial. By considering SSP-BART\textsuperscript{$\star$}, our goal is to empirically demonstrate the benefits of the innovations proposed in CSP-BART and show that merely sharing covariates is insufficient by itself to lead to reduced bias and complete isolation of the parameter estimates in the linear predictor.

For CSP-BART, SSP-BART, and SSP-BART\textsuperscript{$\star$}, we use $T=50$ trees, $2{,}000$ MCMC iterations as burn-in, and $2{,}000$ as post-burn-in. We use the default arguments of the \texttt{mgcv} and \texttt{VCBART} packages, with the exception of \texttt{intercept=FALSE} being specified for \texttt{VCBART} for the sake of comparability with the other BART-based models. We note that the GAM is the only non-Bayesian method among the set of comparators. As GAMs require explicit specification of terms to be included in the linear predictor, we supply the true structure used to simulate the data in both experiments. This gives GAMs an unfair advantage over the other methods, but does provide a baseline that the BART-based methods can aim for. In practice, a misspecified GAM could be expected to perform much worse.

\subsection{Friedman dataset}\label{Friedman_sec}

In this first scenario, we consider the Friedman equation:
\[y_{i} = 10\,\sin\left(\pi x_{i1} x_{i2}\right) + 20\left(x_{i3} - 0.5\right)^{2} + 10 x_{i4} + 5 x_{i5} + \epsilon_{i},\:i = 1, \ldots, n,\]
where $x_{.j} \sim \mbox{Uniform}(0,1) \:\forall\: j = 1, \ldots, p$ and $\epsilon_{i} \sim \mbox{N}(0, \sigma^{2})$. This equation \citep{friedman1991multivariate} is used for benchmarking tree-based methods using synthetic data, and has been used in many other papers, e.g., \citet{chipman2010bart, linero2018bayesian, deshpande2020vcbart}. In this experiment, we set $n = 1000$, $p = (10, 50)$, and $\sigma^{2} = (1, 10)$, totalling four scenarios. To evaluate model performance, we use the bias of the parameter estimates as the accuracy measure, across $50$ replicates of the data-generation process. As the Friedman equation uses only $5$ covariates to generate the response, the additional $x_{.j}$ are noise, and have no impact on $y_{i}$. In this simulation, we aim to estimate the $p_1=2$ linear effects associated with $x_{4}$ and $x_{5}$ (denoted by $\beta_{4} = 10$ and $\beta_{5} = 5$, respectively) using the linear predictor, i.e., we set up $\mathbf{X}_{1}$ in SSP-BART, SSP-BART\textsuperscript{$\star$}, and CSP-BART so that it contains only $x_{4}$ and $x_{5}$. In contrast, we let BART take care of the non-linear and interaction effects by setting $\mathbf{X}_{2}$ to contain all $p$ covariates (including $x_{4}$ and $x_{5}$) for CSP-BART and SSP-BART\textsuperscript{$\star$}; SSP-BART's $\mathbf{X}_{2}$ is set up to contain all $p$ covariates, except $x_{4}$ and $x_{5}$.

\begin{figure}[H]
    \centering
  \includegraphics[width=\textwidth]{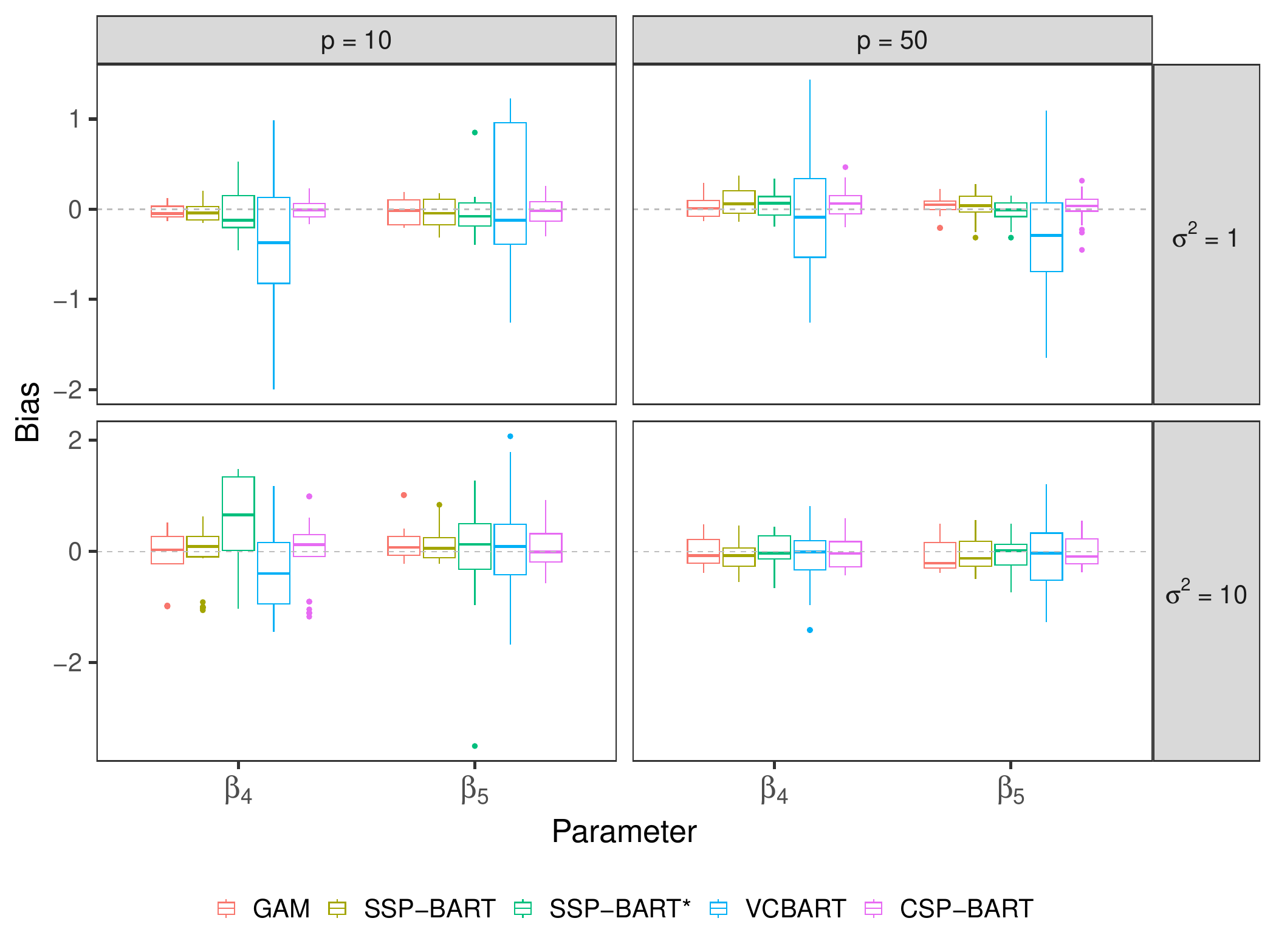}\vfill
    \caption{Boxplots of simulation results obtained across $50$ replicate datasets generated according to the Friedman equation, considering $n = 1000$, $p = (10, 50)$, and $\sigma^{2} = (1, 10)$. The y-axis exhibits the bias related to the parameter estimates $\hat{\beta}_{4}$ and $\hat{\beta}_{5}$ for the novel CSP-BART and various competitors. Recall that the GAM has been given the true model structure so its superior performance is expected.}
     \label{fig_simulation_friedman_results}
 \end{figure}

Figure \ref{fig_simulation_friedman_results} shows the results of bias exhibited by the novel CSP-BART and its competitors for each combination of $p$ and $\sigma^{2}$. As GAMs require all terms that are estimated by the model to be specified, we supply the true structure of the Friedman equation so that it can be used as a reference in the comparison. The CSP-BART, SSP-BART, and SSP-BART\textsuperscript{$\star$} estimates are notably similar. We can see that the bias of the parameter estimates is low and each model recovers the true effects in all four scenarios, with the exception of the $\beta_4$ parameter in the $p=10$ setting for SSP-BART\textsuperscript{$\star$}. This is expected and can be attributed to the fact that $x_{4}$ and $x_{5}$ do not interact with other covariates. Consequently, the trees in CSP-BART tend not to contain $x_{4}$ and $x_{5}$ as both effects are captured solely by the linear predictor. We note also that VCBART presents larger bias for both $\beta_{4}$ and $\beta_{5}$ in all but one scenario. As VCBART estimates $\beta_{4}$ and $\beta_{5}$ using BART models that employ a set of effect modifiers (i.e., all covariates of non-primary interest), the results shown in Figure \ref{fig_simulation_friedman_results} are unsurprising since, in this example, $\beta_{4}$ and $\beta_{5}$ depend exclusively on $x_4$ and $x_5$, respectively.
We have also run simulation experiments (not shown here, for brevity) in which the linear predictor of CSP-BART contains i) $x_1, \ldots, x_5$ and ii) $x_1, \ldots, x_{p}$ to check how much bias is added to the effects of primary interest, which are associated with $x_4$ and $x_5$ only, when the linear predictor is misspecified. For both $x_1, \ldots, x_5 \in \mathbf{X}_{1}$ and $x_1, \ldots, x_p \in \mathbf{X}_{1}$, the results were similar to those presented in Figure \ref{fig_simulation_friedman_results} as no systematic bias was observed on the parameter estimates of primary interest. Finally, we point out that as $x_4$ and $x_5$ are the only predictors of primary interpretational interest, it would make little sense to leave them both out of $\mathbf{X}_1$. In this case, we ran additional experiments (for CSP-BART only) in which $\mathbf{X}_{2}$ contains all $p$ predictors but $x_4$ is left out of $\mathbf{X}_{1}$ and $x_{5}$ is included (and vice-versa). Again, no systematic bias was observed. These results are expected because the predictors in the Friedman equation are uncorrelated and when $x_{4}$ (or $x_{5}$) is left out of $\mathbf{X}_{1}$, its marginal effect is reasonably well estimated by the BART component of CSP-BART.

\subsection{Estimating main effects in the presence of interactions}

In the scenario above, we have shown that the novel CSP-BART correctly estimates the main effects when they do not have any interactions with other effects and thus established that CSP-BART and SSP-BART yield comparable results in such settings. However, in practice, the covariates of primary interest may interact, either among themselves or with other effects, which should be taken into account. In this sense, the simulation setting which follows is likely to better reflect the nature of the TIMSS 2019 data (see Section \ref{section_timss2019}) and other real-world applications.

\begin{figure}[H]
    \centering
  \includegraphics[width=\textwidth]{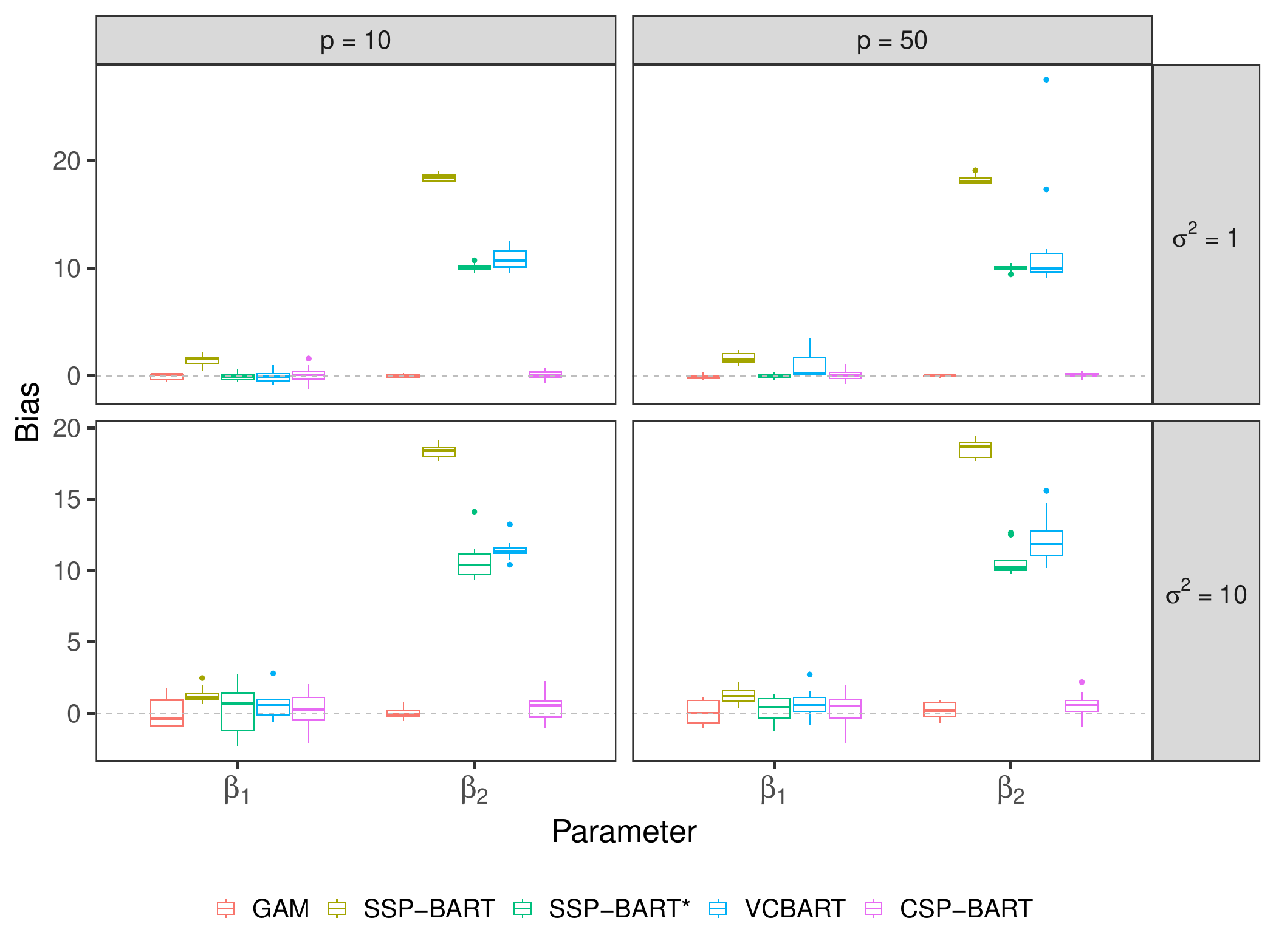}\vfill
    \caption{Boxplots of the simulation results obtained across $50$ replicate datasets generated according to Equation \eqref{tree_based_equation}, considering $n = 1000$, $p = (10, 50)$, and $\sigma^{2} = (1, 10)$. The y-axis exhibits the bias related to the parameter estimates $\hat{\beta}_{1}$ and $\hat{\beta}_{2}$ for the novel CSP-BART and various competitors. Recall that the GAM has been given the true model structure so its superior performance is expected.}
     \label{fig_simulation_linear_results}
 \end{figure}

In this scenario, we compare the methods using the following regression functions:
\begin{align}
\label{tree_based_equation}
y_{i} & = 10 x_{i1} - 5 x_{i2} + \left(\mathcal{T}_{1} \given \mathbf{x}_{1i}, \mathbf{x}_{2i}\right) + \epsilon_{i}, \\
\label{smooth_function_with_interaction}
y_i & = 10x_{i1} - 5 x_{i2} + 10 \cos(\pi x_{i2} x_{i3}), \:i = 1, \ldots, n,
\end{align}
where $x_{.j} \sim \mbox{Uniform}(0,1)\:\forall\:j = 1, \ldots, p$ and $\epsilon_{i} \sim \mbox{N}(0, \sigma^{2})$, as before, and $\mathcal{T}_{1}\given \mathbf{x}_{1i}, \mathbf{x}_{2i}$ represents the tree structure shown in Figure \ref{LM_BART_tree}. As per Section \ref{Friedman_sec}, we consider $n = 1000$, $p = (10, 50)$, and $\sigma^{2} = (1,10)$, where the additional covariates have no impact on the response. We are now interested in estimating the effects associated with $x_{1}$ and $x_{2}$ (denoted by $\beta_{1} = 10$ and $\beta_{2} = -5$, respectively). This is achievable under CSP-BART and SSP-BART\textsuperscript{$\star$} by specifying $\mathbf{X}_{1}$ to contain only $x_{1}$ and $x_{2}$ and $\mathbf{X}_{2}$ to contain all $p$ covariates, including $x_1$ and $x_2$ and the covariate $x_3$, which contributes only to the tree in Figure \ref{LM_BART_tree}. In SSP-BART, however, $x_{1}$ and $x_{2}$ are exclusive to $\mathbf{X}_{1}$ (i.e., neither $x_1$ nor $x_2$ are in $\mathbf{X}_{2}$).
\begin{figure}[H]
\centering
            \begin{forest}
for tree={
    grow=south, draw, minimum size=5ex, inner sep=5pt, s sep=8mm, l sep=7mm
        }
[$\:x_{1} \le 0.5\:$,
    [$\:x_{2} \le 0.5\:$, , edge label={node[midway, font=\footnotesize, left]{$\mbox{ TRUE }$}}
        [$\:\:4\:\:$, circle]
        [$-7$, circle]
    ]
    [$\:x_{3} \le 0.5\:$, , edge label={node[midway, font=\footnotesize, right]{$\mbox{ FALSE }$}}
        [$\:\:3\:\:$, circle]
        [$-8$, circle]
    ]
]
\end{forest}\vfill
\caption{An illustration of the tree structure used to generate the response via Equation \eqref{tree_based_equation}. In if-else format this can be written as $\mathcal{T}_{1} \given \mathbf{x}_i = f(x_{i1}, x_{i2}, x_{i3})= 4\mathds{1}(x_{i1} \le 0.5) \times \mathds{1}(x_{i2} \le 0.5) -7 \mathds{1}(x_{i1} \le 0.5)\times \mathds{1}(x_{i2} > 0.5) + 3 \mathds{1}(x_{i1} > 0.5)\times \mathds{1}(x_{i3} \le 0.5) - 8 \mathds{1}(x_{i1} > 0.5)\times \mathds{1}(x_{i3} > 0.5)$, where $\mathds{1}(\cdot)$ denotes the indicator function. Note that the tree splits on both primary ($x_{1}$ and $x_{2}$) and non-primary ($x_3$) covariates.}
\label{LM_BART_tree}
\end{figure}
Figure \ref{fig_simulation_linear_results} shows the bias in the estimates of $\beta_{1}$ and $\beta_{2}$ for Equation \eqref{tree_based_equation}. While CSP-BART estimates both parameters with low bias, regardless of $p$ and/or $\sigma^2$, SSP-BART gives large bias for $\beta_1$ and even more pronounced bias for $\beta_2$ in all scenarios. These biases occur as $x_{1}$ and $x_{2}$ are not available to the BART component of SSP-BART. However, as per CSP-BART, $x_1$ and $x_2$ are part of $\mathbf{X}_2$ in SSP-BART\textsuperscript{$\star$}, which exhibits less bias than SSP-BART. We conjecture that $\beta_2$ exhibits greater bias than $\beta_1$ --- for both versions of SSP-BART --- because $x_2$ appears at a lower depth than $x_1$ in Figure \ref{LM_BART_tree}; i.e., in closer proximity to terminal nodes. This notion is supported by further experiments, conducted but not shown here, using alternative tree structures with varying depth levels for $x_2$.

We have also created an additional scenario (not shown here, for brevity), where an interaction between $x_1$ and $x_2$ is included in the linear predictor of Equation \eqref{tree_based_equation} in the hope of reducing the bias of SSP-BART's parameter estimates; we omitted this interaction term from CSP-BART as it is not of primary interest and the bias is already low. However, we observed that the inclusion of the interaction between $x_1$ and $x_2$ in the linear predictor has a negligible impact on the bias of both $\beta_1$ and $\beta_2$ as the interaction effect statistically does not differ from zero. Nonetheless, we describe in Appendix \ref{sec_interactions_primary_interest} some strategies for specifying interaction terms in the linear predictor of CSP-BART, which may be of interest in other applications.\vspace{-1em}
\vspace{-1em}%
Furthermore, it can be seen that VCBART and CSP-BART provide similar bias for both parameters, and match well with the baseline GAM model to which the true structure is supplied, as it is unable to capture non-specified interactions. However, it is worth recalling that VCBART uses a BART model to estimate each parameter in the linear predictor. For these data, VCBART uses $50 \times 2 = 100$ trees in total to estimate $\beta_{1}$ and $\beta_{2}$, as the \texttt{VCBART} package uses $50$ trees for each parameter, by default. In this sense, the greater the number of parameters to be estimated in the linear predictor, the more computationally intensive VCBART becomes, since the total number of trees used to estimate all covariate effects is a function of the number of covariates in the linear predictor and the number of trees used to approximate each effect.

\begin{figure}[H]
\centering
\includegraphics[width=\textwidth]{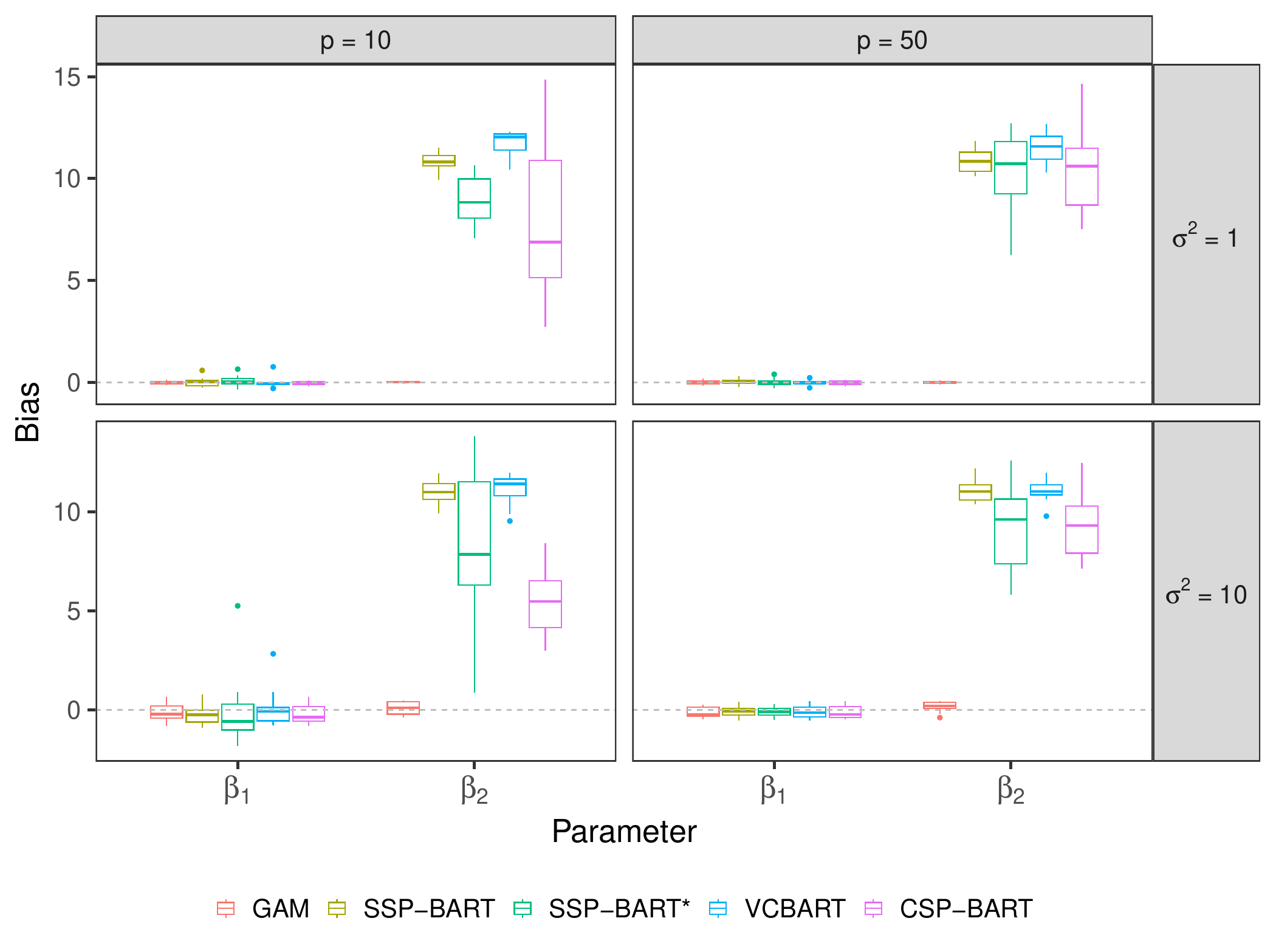}
\caption{Boxplots of the simulation results obtained across 50 replicate datasets generated according to Equation \eqref{smooth_function_with_interaction}, considering $n = 1000$, $p = (10, 50)$, and $\sigma^{2} = (1, 10)$. The y-axis exhibits the bias related to the parameter estimates $\hat{\beta_{1}}$ and $\hat{\beta_{2}}$ for the novel CSP-BART and various competitors. Recall that the GAM has been given the true model structure so its superior performance is expected.}
\label{fig_simulation_smooth_results}
\end{figure}

Unlike Equation \eqref{tree_based_equation} where the interaction term is given by a decision tree, in Equation \eqref{smooth_function_with_interaction} the interaction component is a smooth function which depends on the interaction between predictors in $\mathbf{X}_{1}$ and $\mathbf{X}_{2}$. Figure \ref{fig_simulation_smooth_results} shows the bias in the estimates of $\beta_{1}$ and $\beta_{2}$. While all models estimate $\beta_{1}$ with low bias since it does not form interactions of any kind, SSP-BART and VCBART exhibit larger bias for $\beta_2$ than CSP-BART. The fact that all models exhibit bias is partly due to BART's difficulty to estimate smooth functions properly, which in turn is inherently connected to the step functions used as predicted values. Though the biases of the parameter estimates for CSP-BART are not close to zero, they are better than those for SSP-BART, SSP-BART\textsuperscript{$\star$}, and VCBART, particularly for $p=10$. These results are illuminating because they show that the bias can be reduced by sharing covariates between the linear predictor and BART components. However, the SSP-BART\textsuperscript{$\star$} results indicate that this alone is not sufficient; clearly the other innovations of CSP-BART, particularly the novel double moves, are necessary to achieve a greater reduction in bias. This is consistent with our argument throughout Appendix \ref{appendix_identifiability_csp} that the novel double moves are necessary to ensure the identifiability of the coefficients of the parametric linear component of CSP-BART.\vspace{-1em}

\section{TIMSS 2019 Application\label{real_data_sec}}

We now turn to an analysis of Trends in International Mathematics and Science Study (TIMSS) data, which initially motivated~the development of CSP-BART and offers a large and challenging test of the model. The focus in Section \ref{section_timss2019} is on comparing CSP-BART's performance to other previously proposed tree-based methods, namely BCF, SSP-BART, and VCBART. We defer additional analyses of these data using only our model to Appendix \ref{appendix_csp_missing}. We also demonstrate the use of CSP-BART in a classification rather than regression setting via another, smaller application to a well-known benchmark dataset in Appendix \ref{appendix_pima}. We begin here by outlining the comparative analysis of TIMSS data in Section \ref{section_timss2019}.

TIMSS is an international series of assessments which takes place every four years. The TIMSS 2019 dataset records students' achievements in mathematics and science at the fourth and eighth grade levels in $64$ countries \citep{mullis2020timss,timss2019}, along with information sourced from surveys of students, teachers, and school principals. Here, we are specifically interested in quantifying the impact of some covariates on students' mathematics scores (variable `BSMMAT01'). In our analysis, we only consider data from Ireland (where mathematics is a compulsory subject) pertaining to students at the eighth grade level, comprising $4{,}118$ observations. 

We selected three covariates of primary interpretational interest as candidates for inclusion in the linear predictor via the $\mathbf{X}_1$ matrix. Notably, all three are categorical variables:, `parents' education level' ($6$ levels), `minutes spent on homework' ($6$ levels), and `school discipline problems' ($3$ levels). Our interest in these covariates follows work in the applied literature which shows that students' achievement is influenced by these factors. For instance, a previous investigation into the relation between students' performance in mathematics and various student-level and school-level factors using data from various countries gathered under the third cycle of TIMSS identified significant effects due to family background and time spent on academic activities \citep{martin2000effective}. These three factors as well as a number of others, such as the gender of the student, an index of the wealth of the school and its surrounding area, and a measure of the resources available for learning at home, have also been shown to be strongly related to the outcome in similar international education studies \citep{mohammadpour2015multilevel,grilli2016exploiting}.

As the TIMSS 2019 data were originally split by the sources of information, some data manipulation was required. In particular, we found high levels of missing information across many of the predictors. If we were to keep all $270$ predictors without missing data, we would have only $1{,}448$ complete observations instead of $4{,}118$. In order to avoid the use of imputation methods and avoid reducing the target population by keeping as many observations as possible, we adopt a few different strategies to account for this missingness in our analyses. Firstly, we identify predictors for inclusion in $\mathbf{X}_2$ by performing a variable-screening step using BART on the complete consolidated dataset with $1{,}448$ observations and $270$ predictors. This allows us to work with a smaller subset of predictors (i.e., the 20 most-used by BART) and greatly increase the number of complete observations used from $1{,}448$ to $3{,}224$. Reassuringly, we note that the $p_1=3$ covariates above are among the $20$ most-used covariates under such a model. The remaining $p_2=17$ covariates, which may help to improve prediction but are not of primary interpretational interest, are specified in the $\mathbf{X}_2$ matrix. In what follows, selected primary covariates are also shared with $\mathbf{X}_2$ when fitting the CSP-BART model but excluded from $\mathbf{X}_2$ under SSP-BART and VCBART. Full details of the identified covariates are provided in Table \ref{TIMSS_covs} in Appendix \ref{appendix_timss_dataset}.

It should be noted, however, that the BART-based variable-screening step is arguably a sub-optimal way to analyse the TIMSS 2019 data, given that a key selling point of a BART-based model like CSP-BART is its ability to handle a large number of covariates without pre-specification. We perform the variable screening step in order to facilitate a comparison with competing BART-based methods --- which cannot accommodate missing values in their trees --- with as much completely observed data as possible, similarly to other analyses of TIMSS data \citep[e.g.,][]{grilli2016exploiting}. Thus, we note that the conclusions of the analysis carried out in Section \ref{section_timss2019} may be limited, since only $20$ variables are used and not the full set of initially available predictors. We discuss the rationale for the BART-based variable-screening step in greater detail in Appendix \ref{appendix_timss_screening} and present two additional analyses, using only CSP-DART (given the larger number of predictors), in Appendix \ref{appendix_csp_missing}; we firstly apply CSP-DART on the much smaller number of complete cases and secondly adapt CSP-DART to accommodate missing values in its trees using the strategies proposed by \citet{loh2009improving}. It is worth noting that the conclusions drawn from this latter analysis are broadly consistent with those presented in the comparative analysis which now follows.

\subsection{TIMSS 2019 results for CSP-BART and competing methods \label{section_timss2019}}

Initially, we compare CSP-BART with the Bayesian causal forest model \citep[BCF;][]{hahn2020bayesian}. To do so, we only consider the covariate `school discipline problems' in the linear predictor of CSP-BART and as the treatment variable for BCF, with SSP-BART included as an additional comparator. CSP-BART and SSP-BART thus differ in that this covariate is also specified in $\mathbf{X}_2$ under CSP-BART, but is exclusive to $\mathbf{X}_1$ under SSP-BART. Though this is an ordinal variable with $3$ levels (`hardly any problems', `minor problems', and `moderate to severe problems'), we binarise it by collapsing the first two levels. These modelling decisions are to the advantage of BCF, as it can only deal with a single binary covariate as the treatment effect. The goal is to quantify the impact of discipline problems on students' mathematics scores along with the other $19$ covariates (i.e., the other two primary covariates are specified only in $\mathbf{X}_2$ for this preliminary analysis). 

In Table \ref{tab1_BCF_semiBART}, we summarise the posterior distributions of the parameter estimates for BCF, CSP-BART, and SSP-BART. The marginal effect of school discipline problems is negative in each case, which means that students who study in schools with moderate to severe discipline issues tend to have lower mathematics scores than those in schools with hardly any or minor discipline problems. However, this covariate also defines at least one split in $2.8\%$ of the sampled trees in the BART component of CSP-BART; i.e., it also interacts with non-primary covariates in $\mathbf{X}_2$. Notably, BCF yields a much wider credible interval (CI) than both CSP-BART and SSP-BART, though all CIs exclude zero.
\begin{table}[H]%
\centering
\caption{Descriptive measures of the posterior distribution of the `school discipline problems' covariate's effect on students' mathematics scores. The estimates relate to the level `moderate to severe problems', as the reference level merges those with `hardly any' or `minor' problems.\label{tab1_BCF_semiBART}}\vskip\smallskipamount%
\extrarowheight 2pt
\begin{tabular*}{406pt}{@{\extracolsep\fill}lccc@{\extracolsep\fill}}
\toprule
\textbf{Method} & \textbf{Mean}  & \textbf{$\mathbf{2.5}$-th percentile}  & \textbf{$\mathbf{97.5}$-th percentile} \\
\midrule
BCF & $-36.07$ & $-62.05$ & $-12.24$ \\
CSP-BART & $-37.43$ & $-54.45$ & $-24.78$ \\
SSP-BART & $-38.05$ & $-48.09$ & $-27.73$\\
\bottomrule
\end{tabular*}
\end{table}
As we now consider all three aforementioned categorical covariates of primary interest, we note that BCF is inadequate for this application as it admits only one binary covariate. As VCBART extends BCF to allow for more covariates (of any type) in the linear predictor, we replace BCF with VCBART in the comparison with CSP-BART and SSP-BART. We use $80\%$ of the data for training and use the remaining $20\%$ as a test set to evaluate out-of-sample prediction performance. As a base for comparison, we note that the RMSEs obtained using just the average of the training and test outcomes are $65.10$ and $64.48$, respectively. Firstly, the RMSEs on the training and test sets are comparable for CSP-BART ($57.2$ and $58.2$) and SSP-BART ($57.6$ and $58.6$)\footnotemark\footnotetext{Notably, we use code from our own implementation of CSP-BART in order to fit both CSP-BART and SSP-BART, as it is not possible to predict on out-of-sample data using the \textsf{R} implementation of SSP-BART provided by the authors of \citet{zeldow2019semiparametric}. This is achieved by adopting the diffuse isotropic prior $\boldsymbol\beta \sim \mbox{MVN}(\mathbf{0}_{p_1}, \sigma^{2}_{b}\mathbf{I}_{p_1})$ and appropriately specifying the design matrices $\mathbf{X}_1$ and $\mathbf{X}_2$ when fitting SSP-BART.}, but VCBART ($57.28$ and $60.35$) is slightly worse for the test set. Secondly, we present the parameter estimates based on the training set under each model for the three chosen primary covariates, along with associated $90\%$ CIs, in Table \ref{tab2_VCBART_semiBART}.

\begin{landscape}
\begin{table}%
\caption{Posterior mean estimates and corresponding $90\%$ credible intervals (in parentheses) for the effects of parents' education level, minutes spent on homework, and school discipline problems on students' mathematics scores. The number of observations in each categorical level is shown in parentheses throughout the `Category' column. The results are based on a training subset ($80\%$) of the TIMSS 2019 dataset. The boldface font is used to highlight the cases where the CI does not contain zero. \label{tab2_VCBART_semiBART}}\vskip\smallskipamount
\centering
\setlength{\tabcolsep}{4.125pt}
\extrarowheight 6.5pt
\begin{tabular*}{580pt}{llccc}
\toprule
& &\multicolumn{3}{c}{\textbf{Estimate ($\mathbf{95}$\% CI)}} \\ \cmidrule{3-5} \textbf{Covariate} & \textbf{Category} $\bm{(n)}$ & \textbf{CSP-BART}  & \textbf{SSP-BART}  & \textbf{VCBART} \\
\midrule
\multirow{6}{*}{Parents' education level}   
& University or higher $(870)$              & $\mathbf{20.94 (14.94; 26.58)}$   & $23.39  (-30.51;82.2)$            & $\bm{30.72 (24.42;41.80)}$  \\ 
& Post-secondary but not university $(546)$ & $\mathbf{18.77 (13.05; 24.40)}$   & $19.54  (-28.1;79.22)$            & $\bm{21.52 (13.76;24.25)}$ \\ 
& Upper secondary $(340)$                   & $-3.62 (-10.70; 3.80)$            & $-15.96 (-86.55;26.39)$           & $-7.75 (-30.99;32.15)$ \\ 
& Lower secondary $(96)$                    & $\mathbf{-11.84 (-22.86; -1.31)}$ & $-8.58  (-49.71;36.2)$            & $-18.65 (-36.02;.0.26)$ \\ 
& Primary, secondary, or no school $(42)$    & $\mathbf{-21.08 (-36.62; -5.38)}$ & $-26.77 (-75.14;13.94)$           & $-\bm{23.71 (-45.92;-2.12)}$ \\ 
& Not informed $(685)$                      & $-3.17 (-9.34; 2.61)$             & $8.37 (-27.53; 52.63)$            & $-2.14 (-28.12;16.38)$ \\[1.5ex]
\hline
\multirow{6}{*}{Minutes spent on homework} 
& No homework $(21)$            & $\mathbf{-24.92 (-45.88; -2.19)}$ & $-27.63 (-85.41;26.96)$   & $-185.60 (-283.72;84.96)$ \\ 
& $1$ to $15$ minutes $(862)$   & $1.62 (-6.57; 11.95)$             & $2.10 (-35.49;50.92)$     & $33.74 (-85.35;69.11)$ \\
& $16$ to $30$ minutes $(1158)$ & $6.04 (-1.35; 12.83)$             & $8.19  (-40.4;43.55) $    & $38.09 (-68.91;66.89)$ \\
& $31$ to $60$ minutes $(441)$  & $7.83 (-1.30; 15.88)$             & $10.12  (-23.49;42.7)$    & $37.11 (-60.23;67.10)$ \\
& $61$ to $90$ minutes $(62)$   & $9.00 (-3.36; 23.19)$             & $9.88  (-32.3;54.68)$     & $20.54 (-77.53;65.39)$ \\
& More than $90$ minutes $(35)$ & $0.42 (-20.32; 20.57)$            & $-2.64 (-57.06; 40.97)$   & $54.13 (-11.64;400.80)$ \\[1.5ex]
\hline
\multirow{3}{*}{School discipline problems} 
& Hardly any problems $(1621)$          & $\mathbf{14.52 (9.05; 19.98)}$        & $10.27 (-28;36.18)$       & $16.49 (-32.07;75.77)$ \\
& Minor problems $(891)$                & $\mathbf{10.06 (4.68; 15.56)}$        & $13.59 (-19.96;44.23)$    & $3.72 (-35.32;11.35)$ \\
& Moderate to severe problems $(67)$    & $\mathbf{-24.58 (-33.51; -15.60)}$    & $-23.85 (-61.83; 10.98)$  & $-20.22 (-86.80; 66.82)$ \\[1ex]
\bottomrule
\end{tabular*}
\end{table}
\end{landscape}

Students whose parents studied at `university or higher' or obtained `post-secondary' qualifications tend to have higher mathematics scores than those whose parents were educated up to secondary level at most. The effects become more pronounced at lower education levels. A similar pattern of higher scores is observed under VCBART for students who devote increasingly more time to homework. However, both CSP-BART and SSP-BART suggest that students who spend `more than $90$ minutes' on homework score less than those who spend less time (but still more than those who have/do `no homework'). VCBART's estimates are quite extreme for these two levels, possibly due to the small numbers of observations therein. Lastly, all models suggest that students in schools with `moderate to severe' discipline problems tend to have lower scores than those in schools with `hardly any' or `minor' problems.

Though their posterior mean estimates differ only in magnitude and not in sign (with~only two exceptions), another important aspect shown in Table \ref{tab2_VCBART_semiBART} is the difference between the CIs from CSP-BART and the other methods. Notably, all CIs are much wider for SSP-BART and VCBART. In particular, they all contain zero under SSP-BART, while the CIs for all effects associated with the covariate `school discipline problems` and some levels of the other primary covariates are bounded away from zero by CSP-BART. As CSP-BART and SSP-BART assume different priors for the linear regression parameters, we conducted additional experiments (not shown here, for brevity) by fitting hybrid models which swap their priors on $\boldsymbol{\beta}$. In doing so, we verified that the assumption of a diffuse isotropic prior under SSP-BART is driving the disparities in these intervals. Thus, CSP-BART allowing the effects of covariates of primary interpretational interest to be correlated and have different variances \emph{a priori} appears to have a strong impact on the posterior uncertainty of the estimates.

To show the benefits of CSP-BART sharing covariates across components, it is of interest to detect interaction effects between covariates in $\mathbf{X}_1$ and others of both primary and non-primary interest. Though the sum of the topology of the trees can be seen as a new~\mbox{single~tree}, which may be of interest in some applications, we focus on finding interactions from~the individual trees in the ensemble. According to \citet{damien2013bayesian},~an interaction exists between two variables if both variables (or one or more binary \mbox{indicators} associated with each variable) are in the same tree. This definition makes some sense for BART-based methods due to the way the trees are designed and learned. Here, $26.9\%$ of trees across all MCMC samples have interactions of this sort between at least one covariate~in $\mathbf{X}_1$ and another in either $\mathbf{X}_1$ or $\mathbf{X}_2$, while $1\%$ are stumps and $50.5\%$ split on one covariate only. Following a definition of non-spurious interactions, a stricter criterion for detecting interactions is provided by \citet{JSSv070i04}, whereby covariates must be in the same \emph{branch}. Among these $26.9\%$ of trees, the majority of interactions we detect involving at least one primary covariate are of this more specific nature; e.g., between `parents' education level' and `minutes spent on homework' (both in $\mathbf{X}_1$) and between `school discipline problems' (in $\mathbf{X}_1$) and `absenteeism' ($5$ levels, in $\mathbf{X}_2$). A major limitation of SSP-BART is that it would fail to detect key interactions such as these. Due to the assumption of mutual-exclusivity between $\mathbf{X}_1$ and $\mathbf{X}_2$, SSP-BART can only capture interactions between two or more non-primary covariates in $\mathbf{X}_2$. Our CSP-BART also detects frequent interactions of this sort in the remaining $21.6\%$ of trees; e.g., between `absenteeism' and `how often the student feels hungry' ($4$ levels). To detect important interactions in VCBART, one would need to examine all trees for all covariates in the linear predictor. This would amount to $150$ trees per iteration, as the effect associated with each of the $3$ primary covariates is approximated by $50$ trees (by default).

\begin{figure}[H]
\begin{center}
     \includegraphics[width=\textwidth]{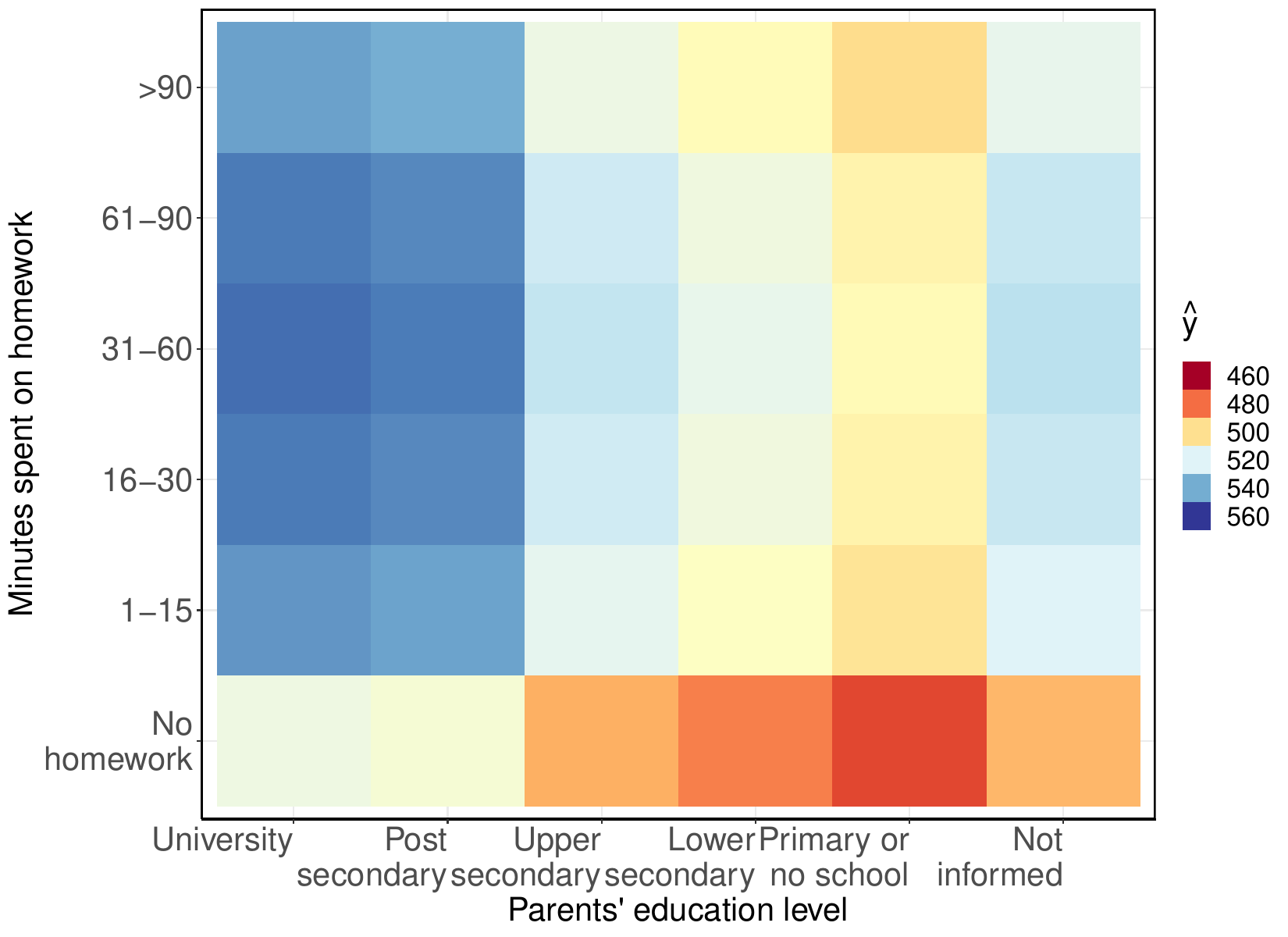}
     \captionsetup{justification=centering}
     \label{interaction_plot_b}
   %\captionsetup{justification=centering}
\end{center}
\caption{Bivariate partial dependence plot between the predictors `parents' education level' and `minutes spent on homework' for CSP-BART's estimate of the mathematics scores ($\hat{y}$) on the TIMSS 2019 data.}
\label{interaction_plot}
\end{figure}

The aforementioned interaction between parents' education level and minutes spent on homework is of particular interest, as both variables are specified in both $\mathbf{X}_1$ and $\mathbf{X}_2$. To further elucidate this interaction, we explore it visually in Figure \ref{interaction_plot}. This type of plot is due to \citet{inglis2022visualizing}, who introduce metric- and model-agnostic visualisations for exploring variable importance and variable interactions for supervised machine learning algorithms. They present the generalised pairs partial dependence plot in the context of numerical predictors, which displays one-way partial dependence plots and individual conditional expectation curves \citep[ICE;][]{goldstein2015peeking} with superimposed partial dependence curves along the diagonal, as well as off-diagonal bivariate partial dependence plots and scatterplots, all coloured by the predicted values. Though the authors demonstrate their visualisations by fitting random forests and $k$-nearest neighbours to well-known benchmark datasets, it is straightforward to use them to visualise variable interactions in the context of BART-based models and adapt them to categorical predictors\footnote{Recall that the covariates of primary interpretational interest in the TIMSS 2019 dataset --- parents’ education level, minutes spent on homework, and school discipline problems --- are all categorical predictors. Moreover, all variables of non-primary interest are also
categorical. See Table \ref{TIMSS_covs} in Appendix \ref{appendix_timss_dataset} for a list of all $20$ covariates pre-selected by the BART-based variable-screening step.}. Given the categorical nature of the predictors comprising the one interaction of interest, we choose to display one bivariate partial dependence plot only in Figure \ref{interaction_plot}, thus omitting the scatterplots and the ICE curves.
\vspace{-1ex}%
We know from Table \ref{tab2_VCBART_semiBART} that i) students with no homework tend to have lower mathematics scores and ii) students with parents with high level of education (`university or higher' and `post-secondary but not university') tend to perform better than students whose parents attended school up to secondary level at most. In Figure \ref{interaction_plot}, however, we can see that students with parents with high level of education with no homework tend to perform particularly worse than their counterparts who have some homework, with the interaction effect becoming more pronounced at lower levels of education. Notably, within each level of education background, students who spend more than $90$ minutes on homework tend to have slightly worse/similar mathematics scores compared to those who spend $1$ to $15$ minutes, which suggests that the homework load does not have a linear effect (e.g., the more homework, the better the score) on Irish students' mathematics performance at eighth grade level. This non-linearity demonstrates the importance of relaxing the mutual-exclusivity assumption of SSP-BART by sharing these predictors between the linear and BART components of CSP-BART. 

\section{Discussion}\label{conclusion_sec}

In this work, we have extended BART to a semi-parametric framework which circumvents many of the restrictions and non-identifiability issues found in other versions of semi-parametric BART. In semi-parametric BART models, the main effects are estimated via a linear predictor, while interactions and non-linearities are dealt with by a BART component. The main novelties of our CSP-BART are i) the sharing of covariates between the linear and BART components, in tandem with ii) additional double-grow and double-prune moves. These innovations combine to induce additional interactions between covariates of primary interest, both among themselves and with those available to the BART component, which ensures that marginal effects of primary interest are strictly isolated in the identifiable linear component while interactions and non-linearities are strictly confined to the BART component. Our modifications can be interpreted as adjustments to the prior over the set of possible tree structures; effectively, a prior probability of zero is placed on invalid trees. We have implemented CSP-BART as an \textsf{R} package, which is currently available at \url{https://github.com/ebprado/CSP-BART}.

Through extensive simulation studies, applications to novel data from an international education assessment, and an additional application to a well-known benchmark dataset with a binary response variable, the ability of CSP-BART to estimate marginal effects with low bias, while not requiring pre-specification of interaction effects, has been demonstrated in both regression and classification settings. Regarding the motivating TIMSS 2019 application, we note that CSP-BART offers particularly interesting insight into the effect of the `minutes spent on homework' covariate on students' mathematics scores. While competing methods suggest either that this variable is not statistically significant or that scores improve indefinitely as the time devoted to homework increases, CSP-BART suggests that the effect of this predictor on mathematics scores begins to yield diminishing returns for those who spend an excessive amount of time (more than $90$ minutes) on homework, which implies that students who do so might actually be weak students who struggle with their homework exercises or mathematics classes in general.

We also showed that CSP-BART captures many important interactions between covariates of primary interest and others of both primary and non-primary interest in the TIMSS 2019 application, by virtue of CSP-BART allowing the covariates of primary interest to be shared with both the linear and BART components. Such interactions cannot be captured by the SSP-BART or VCBART models, and would need to be explicitly specified if instead fitting a linear model such as a GLM or a GAM. However, we note that mixed-effects models are widely used in the analysis of similar education assessment datasets \citep{mohammadpour2015multilevel,grilli2016exploiting}. As the present analysis in Section \ref{section_timss2019} only considers fixed effects in the linear component, our proposals outlined in Section \ref{random_effect_sec} for also incorporating random effects in the parametric component of CSP-BART are thus of interest for future practical work. This is especially worthy of further investigation in light of the model of \citet{dorie2022stan} suffering from the same limitations as SSP-BART, in terms of identifiability of the parametric component, when the assumption of mutual-exclusivity between the two sets of predictors is not enforced.

Overall, we anticipate CSP-BART enjoying great utility in a wide range of application settings, beyond the TIMSS 2019 data considered herein. The model accommodates multiple covariates, yields improved out-of-sample prediction performance, and ensures accurate inference of important linear effects while accounting for additional non-specified interactions (beyond those already accounted for by other semi-parametric BART models). Furthermore, the model-fitting algorithm enables straightforward incorporation of random effects and has built-in strategies to address non-identifiability issues. Notably, its run-time is comparable to SSP-BART and BART for small to moderate numbers of predictors in the linear predictor.

In terms of future methodological extensions, other BART-based models, such as SBART \citep{linero2018bayesian}, log-linear BART \citep{murray2021log}, and BART for gamma and log-normal hurdle data \citep{linero2020semipar} could be embedded in semi-parametric frameworks following a similar approach. A semi-parametric version of SBART, in particular, could prove especially fruitful for the TIMSS 2019 application. Theoretical results underlying CSP-BART could also be developed in order to explore its posterior convergence properties. Finally, two specific extensions to CSP-BART which may be of particular interest, relating to the inclusion of non-linear and interaction effects of primary interpretational interest \emph{in the linear predictor}, are outlined in greater detail in Appendix \ref{appendix_extra_linear}.

%%%%%%%%%%%%%%%%%%%%%%%%%%%%%%%%%%%%%%%%%%%%%%
%% Support information, if any,             %%
%% should be provided in the                %%
%% Acknowledgements section.                %%
%%%%%%%%%%%%%%%%%%%%%%%%%%%%%%%%%%%%%%%%%%%%%%
\begin{acks}[Acknowledgments]
We thank the editors and the three anonymous referees for their thorough comments, questions, and suggestions that substantially improved the earlier version of the paper. We also thank Dr. Alan Inglis for helping with the generalised pairs partial dependence plot in Section \ref{section_timss2019}. Estev\~ao Prado and Andrew Parnell were supported by SFI~\mbox{Research} Centre award 12/RC/2289P2 and Science Foundation Ireland Career Development Award 17/CDA/4695. Prof. Parnell's work was also supported by: an investigator award (16/IA/4520); a Marine Research Programme funded by the Irish Government, co-financed by the European Regional Development Fund (Grant-Aid Agreement No. PBA/CC/18/01); European Union's Horizon 2020 research and innovation programme (grant agreement No. 818144); SFI Centre for Research Training 18CRT/6049; and SFI Research Centre award 16/RC/3872.
\end{acks}\clearpage

%% if your bibliography is in bibtex format, uncomment commands:
\bibliographystyle{imsart-nameyear} % Style BST file
\bibliography{mybibfile}       % Bibliography file (usually '*.bib')

\begin{thebibliography}{71}
% BibTex style file: imsart-nameyear.bst, 2017-11-03
% Default style options (sort=1,type=nameyear).
% Used options (sort=1,type=nameyear).

\bibitem[\protect\citeauthoryear{Albert and Chib}{1993}]{albert1993bayesian}
\begin{barticle}[author]
\bauthor{\bsnm{Albert},~\bfnm{James~H}\binits{J.~H.}} \AND
  \bauthor{\bsnm{Chib},~\bfnm{Siddhartha}\binits{S.}}
(\byear{1993}).
\btitle{Bayesian analysis of binary and polychotomous response data}.
\bjournal{Journal of the American Statistical Association}
\bvolume{88}
\bpages{669--679}.
\end{barticle}
\endbibitem

\bibitem[\protect\citeauthoryear{Bates et~al.}{2015}]{lme4}
\begin{barticle}[author]
\bauthor{\bsnm{Bates},~\bfnm{Douglas}\binits{D.}},
  \bauthor{\bsnm{M{\"a}chler},~\bfnm{Martin}\binits{M.}},
  \bauthor{\bsnm{Bolker},~\bfnm{Ben}\binits{B.}} \AND
  \bauthor{\bsnm{Walker},~\bfnm{Steve}\binits{S.}}
(\byear{2015}).
\btitle{Fitting Linear Mixed-Effects Models Using {lme4}}.
\bjournal{Journal of Statistical Software}
\bvolume{67}
\bpages{1--48}.
\end{barticle}
\endbibitem

\bibitem[\protect\citeauthoryear{Beaton et~al.}{1996}]{beaton1996mathematics}
\begin{bbook}[author]
\bauthor{\bsnm{Beaton},~\bfnm{Albert~E}\binits{A.~E.}} \betal{et~al.}
(\byear{1996}).
\btitle{Mathematics Achievement in the Middle School Years. IEA's Third
  International Mathematics and Science Study (TIMSS).}
\bpublisher{ERIC}.
\end{bbook}
\endbibitem

\bibitem[\protect\citeauthoryear{Bouhlila and
  Sellaouti}{2013}]{bouhlila2013multiple}
\begin{barticle}[author]
\bauthor{\bsnm{Bouhlila},~\bfnm{Donia~Smaali}\binits{D.~S.}} \AND
  \bauthor{\bsnm{Sellaouti},~\bfnm{Fethi}\binits{F.}}
(\byear{2013}).
\btitle{Multiple imputation using chained equations for missing data in
  {TIMSS}: a case study}.
\bjournal{Large-scale Assessments in Education}
\bvolume{1}
\bpages{1--33}.
\end{barticle}
\endbibitem

\bibitem[\protect\citeauthoryear{Breiman}{2001}]{breiman2001random}
\begin{barticle}[author]
\bauthor{\bsnm{Breiman},~\bfnm{Leo}\binits{L.}}
(\byear{2001}).
\btitle{Random forests}.
\bjournal{Machine Learning}
\bvolume{45}
\bpages{5--32}.
\end{barticle}
\endbibitem

\bibitem[\protect\citeauthoryear{Chen}{2022}]{chen2022effects}
\begin{barticle}[author]
\bauthor{\bsnm{Chen},~\bfnm{Xin}\binits{X.}}
(\byear{2022}).
\btitle{The effects of individual-and class-level achievement on attitudes
  towards mathematics: An analysis of Hong Kong students using {TIMSS} 2019}.
\bjournal{Studies in Educational Evaluation}
\bvolume{72}
\bpages{101113}.
\end{barticle}
\endbibitem

\bibitem[\protect\citeauthoryear{Chipman, George and
  McCulloch}{1998}]{chipman1998bayesian}
\begin{barticle}[author]
\bauthor{\bsnm{Chipman},~\bfnm{Hugh~A}\binits{H.~A.}},
  \bauthor{\bsnm{George},~\bfnm{Edward~I}\binits{E.~I.}} \AND
  \bauthor{\bsnm{McCulloch},~\bfnm{Robert~E}\binits{R.~E.}}
(\byear{1998}).
\btitle{Bayesian {CART} model search}.
\bjournal{Journal of the American Statistical Association}
\bvolume{93}
\bpages{935--948}.
\end{barticle}
\endbibitem

\bibitem[\protect\citeauthoryear{Chipman, George and
  McCulloch}{2010}]{chipman2010bart}
\begin{barticle}[author]
\bauthor{\bsnm{Chipman},~\bfnm{Hugh~A}\binits{H.~A.}},
  \bauthor{\bsnm{George},~\bfnm{Edward~I}\binits{E.~I.}} \AND
  \bauthor{\bsnm{McCulloch},~\bfnm{Robert~E}\binits{R.~E.}}
(\byear{2010}).
\btitle{{BART: {B}ayesian additive regression trees}}.
\bjournal{The Annals of Applied Statistics}
\bvolume{4}
\bpages{266--298}.
\end{barticle}
\endbibitem

\bibitem[\protect\citeauthoryear{Chipman, George and
  McCulloch}{2013}]{damien2013bayesian}
\begin{bincollection}[author]
\bauthor{\bsnm{Chipman},~\bfnm{Hugh~A}\binits{H.~A.}},
  \bauthor{\bsnm{George},~\bfnm{Edward~I}\binits{E.~I.}} \AND
  \bauthor{\bsnm{McCulloch},~\bfnm{Robert~E}\binits{R.~E.}}
(\byear{2013}).
\btitle{Bayesian regression structure discovery}.
In \bbooktitle{Bayesian Theory and Applications}
(\beditor{\bfnm{Paul}\binits{P.}~\bsnm{Damien}},
  \beditor{\bfnm{Petros}\binits{P.}~\bsnm{Dellaportas}},
  \beditor{\bfnm{Nicholas~G}\binits{N.~G.}~\bsnm{Polson}} \AND
  \beditor{\bfnm{David~A}\binits{D.~A.}~\bsnm{Stephens}}, eds.)
\bchapter{22},
\bpages{451--465}.
\bpublisher{Oxford University Press}.
\end{bincollection}
\endbibitem

\bibitem[\protect\citeauthoryear{Chipman et~al.}{2022}]{chipman2022mbart}
\begin{barticle}[author]
\bauthor{\bsnm{Chipman},~\bfnm{Hugh~A}\binits{H.~A.}},
  \bauthor{\bsnm{George},~\bfnm{Edward~I}\binits{E.~I.}},
  \bauthor{\bsnm{McCulloch},~\bfnm{Robert~E}\binits{R.~E.}} \AND
  \bauthor{\bsnm{Shively},~\bfnm{Thomas~S}\binits{T.~S.}}
(\byear{2022}).
\btitle{{mBART: multidimensional monotone BART}}.
\bjournal{Bayesian Analysis}
\bvolume{17}
\bpages{515--544}.
\end{barticle}
\endbibitem

\bibitem[\protect\citeauthoryear{Deshpande et~al.}{2020}]{deshpande2020vcbart}
\begin{barticle}[author]
\bauthor{\bsnm{Deshpande},~\bfnm{Sameer~K}\binits{S.~K.}},
  \bauthor{\bsnm{Bai},~\bfnm{Ray}\binits{R.}},
  \bauthor{\bsnm{Balocchi},~\bfnm{Cecilia}\binits{C.}},
  \bauthor{\bsnm{Starling},~\bfnm{Jennifer~E}\binits{J.~E.}} \AND
  \bauthor{\bsnm{Weiss},~\bfnm{Jordan}\binits{J.}}
(\byear{2020}).
\btitle{{VCBART: B}ayesian trees for varying coefficients}.
\bjournal{arXiv preprint arXiv:2003.06416}.
\end{barticle}
\endbibitem

\bibitem[\protect\citeauthoryear{Dorie}{2020}]{dbarts}
\begin{bmisc}[author]
\bauthor{\bsnm{Dorie},~\bfnm{Vincent}\binits{V.}}
(\byear{2020}).
\btitle{dbarts: discrete {B}ayesian additive regression trees sampler}.
\bnote{{R} package version 0.9-19}.
\end{bmisc}
\endbibitem

\bibitem[\protect\citeauthoryear{Dorie et~al.}{2022}]{dorie2022stan}
\begin{barticle}[author]
\bauthor{\bsnm{Dorie},~\bfnm{Vincent}\binits{V.}},
  \bauthor{\bsnm{Perrett},~\bfnm{George}\binits{G.}},
  \bauthor{\bsnm{Hill},~\bfnm{Jennifer~L}\binits{J.~L.}} \AND
  \bauthor{\bsnm{Goodrich},~\bfnm{Benjamin}\binits{B.}}
(\byear{2022}).
\btitle{{Stan and BART for causal inference: estimating heterogeneous treatment
  effects using the power of Stan and the flexibility of machine learning}}.
\bjournal{Entropy}
\bvolume{24}
\bpages{1782}.
\end{barticle}
\endbibitem

\bibitem[\protect\citeauthoryear{Duchon}{1977}]{duchon1977splines}
\begin{binproceedings}[author]
\bauthor{\bsnm{Duchon},~\bfnm{Jean}\binits{J.}}
(\byear{1977}).
\btitle{Splines minimizing rotation-invariant semi-norms in {S}obolev spaces}.
In \bbooktitle{Constructive Theory of Functions of Several Variables:
  Proceedings of a Conference Held at Oberwolfach April 25--May 1, 1976}
\bpages{85--100}.
\bpublisher{Springer}.
\end{binproceedings}
\endbibitem

\bibitem[\protect\citeauthoryear{Eilers and Marx}{1996}]{eilers1996flexible}
\begin{barticle}[author]
\bauthor{\bsnm{Eilers},~\bfnm{Paul~HC}\binits{P.~H.}} \AND
  \bauthor{\bsnm{Marx},~\bfnm{Brian~D}\binits{B.~D.}}
(\byear{1996}).
\btitle{Flexible smoothing with {B}-splines and penalties}.
\bjournal{Statistical Science}
\bvolume{11}
\bpages{89--121}.
\end{barticle}
\endbibitem

\bibitem[\protect\citeauthoryear{Fishbein, Foy and Yin}{2021}]{timss2019}
\begin{btechreport}[author]
\bauthor{\bsnm{Fishbein},~\bfnm{B}\binits{B.}},
  \bauthor{\bsnm{Foy},~\bfnm{P}\binits{P.}} \AND
  \bauthor{\bsnm{Yin},~\bfnm{L}\binits{L.}}
(\byear{2021}).
\btitle{{TIMSS 2019 User Guide for the International Database}}
\btype{Technical Report},
\bpublisher{TIMSS \& PIRLS International Study Center},
\baddress{Boston, USA}
\bnote{Retrieved from Boston College, TIMSS \& PIRLS International Study Center
  website:
  \url{https://timssandpirls.bc.edu/timss2019/international-database/}}.
\end{btechreport}
\endbibitem

\bibitem[\protect\citeauthoryear{Foy}{2017}]{foy2017}
\begin{btechreport}[author]
\bauthor{\bsnm{Foy},~\bfnm{P.}\binits{P.}}
(\byear{2017}).
\btitle{{TIMSS 2015 User Guide for the International Database}}
\btype{Technical Report},
\bpublisher{TIMSS \& PIRLS International Study Center, Lynch School of
  Education, Boston College, and International Association for the Evaluation
  of Educational Achievement (IEA)}.
\end{btechreport}
\endbibitem

\bibitem[\protect\citeauthoryear{Foy and O'Dwyer}{2013}]{foy2013technical}
\begin{bincollection}[author]
\bauthor{\bsnm{Foy},~\bfnm{Pierre}\binits{P.}} \AND
  \bauthor{\bsnm{O'Dwyer},~\bfnm{Laura~M.}\binits{L.~M.}}
(\byear{2013}).
\btitle{{Technical Appendix B}. {S}chool effectiveness models and analyses}.
In \bbooktitle{TIMSS and PIRLS 2011 Relationships Among Reading, Mathematics,
  and Science Achievement at the Fourth Grade-Implications for Early Learning}
(\beditor{\bfnm{MO}\binits{M.}~\bsnm{Martin}} \AND
  \beditor{\bfnm{VS}\binits{V.}~\bsnm{Mullis}}, eds.)
\bpublisher{TIMSS \& PIRLS International Study Center, Boston College, Chestnut
  Hill, MA, U.S.A.}
\end{bincollection}
\endbibitem

\bibitem[\protect\citeauthoryear{Friedman}{1991}]{friedman1991multivariate}
\begin{barticle}[author]
\bauthor{\bsnm{Friedman},~\bfnm{Jerome~H}\binits{J.~H.}}
(\byear{1991}).
\btitle{Multivariate adaptive regression splines}.
\bjournal{The Annals of Statistics}
\bvolume{19}
\bpages{1--67}.
\end{barticle}
\endbibitem

\bibitem[\protect\citeauthoryear{Friedman}{2001}]{friedman2001greedy}
\begin{barticle}[author]
\bauthor{\bsnm{Friedman},~\bfnm{Jerome~H}\binits{J.~H.}}
(\byear{2001}).
\btitle{Greedy function approximation: a gradient boosting machine}.
\bjournal{The Annals of Statistics}
\bvolume{29}
\bpages{1189--1232}.
\end{barticle}
\endbibitem

\bibitem[\protect\citeauthoryear{Goldstein et~al.}{2015}]{goldstein2015peeking}
\begin{barticle}[author]
\bauthor{\bsnm{Goldstein},~\bfnm{Alex}\binits{A.}},
  \bauthor{\bsnm{Kapelner},~\bfnm{Adam}\binits{A.}},
  \bauthor{\bsnm{Bleich},~\bfnm{Justin}\binits{J.}} \AND
  \bauthor{\bsnm{Pitkin},~\bfnm{Emil}\binits{E.}}
(\byear{2015}).
\btitle{Peeking inside the black box: Visualizing statistical learning with
  plots of individual conditional expectation}.
\bjournal{Journal of Computational and Graphical Statistics}
\bvolume{24}
\bpages{44--65}.
\end{barticle}
\endbibitem

\bibitem[\protect\citeauthoryear{Green and Kern}{2012}]{green2012modeling}
\begin{barticle}[author]
\bauthor{\bsnm{Green},~\bfnm{Donald~P}\binits{D.~P.}} \AND
  \bauthor{\bsnm{Kern},~\bfnm{Holger~L}\binits{H.~L.}}
(\byear{2012}).
\btitle{Modeling heterogeneous treatment effects in survey experiments with
  {B}ayesian additive regression trees}.
\bjournal{Public Opinion Quarterly}
\bvolume{76}
\bpages{491--511}.
\end{barticle}
\endbibitem

\bibitem[\protect\citeauthoryear{Grilli et~al.}{2016}]{grilli2016exploiting}
\begin{barticle}[author]
\bauthor{\bsnm{Grilli},~\bfnm{Leonardo}\binits{L.}},
  \bauthor{\bsnm{Pennoni},~\bfnm{Fulvia}\binits{F.}},
  \bauthor{\bsnm{Rampichini},~\bfnm{Carla}\binits{C.}} \AND
  \bauthor{\bsnm{Romeo},~\bfnm{Isabella}\binits{I.}}
(\byear{2016}).
\btitle{{Exploiting TIMSS and PIRLS combined data: multivariate multilevel
  modelling of student achievement}}.
\bjournal{The Annals of Applied Statistics}
\bvolume{10}
\bpages{2405--2426}.
\end{barticle}
\endbibitem

\bibitem[\protect\citeauthoryear{Hahn et~al.}{2020}]{hahn2020bayesian}
\begin{barticle}[author]
\bauthor{\bsnm{Hahn},~\bfnm{P~Richard}\binits{P.~R.}},
  \bauthor{\bsnm{Murray},~\bfnm{Jared~S}\binits{J.~S.}},
  \bauthor{\bsnm{Carvalho},~\bfnm{Carlos~M}\binits{C.~M.}} \betal{et~al.}
(\byear{2020}).
\btitle{Bayesian regression tree models for causal inference: regularization,
  confounding, and heterogeneous effects}.
\bjournal{Bayesian Analysis}
\bvolume{15}
\bpages{965--1056}.
\end{barticle}
\endbibitem

\bibitem[\protect\citeauthoryear{Harezlak, Ruppert and
  Wand}{2018}]{harezlak2018semiparametric}
\begin{bbook}[author]
\bauthor{\bsnm{Harezlak},~\bfnm{Jaroslaw}\binits{J.}},
  \bauthor{\bsnm{Ruppert},~\bfnm{David}\binits{D.}} \AND
  \bauthor{\bsnm{Wand},~\bfnm{Matt~P}\binits{M.~P.}}
(\byear{2018}).
\btitle{Semiparametric Regression with R}.
\bpublisher{Springer}.
\end{bbook}
\endbibitem

\bibitem[\protect\citeauthoryear{Hastie and
  Tibshirani}{1990}]{hastie1990generalized}
\begin{bbook}[author]
\bauthor{\bsnm{Hastie},~\bfnm{Trevor~J}\binits{T.~J.}} \AND
  \bauthor{\bsnm{Tibshirani},~\bfnm{Robert~J}\binits{R.~J.}}
(\byear{1990}).
\btitle{Generalized additive models}.
\bseries{Monographs on Statistics and Applied Probability}
\bvolume{43}.
\bpublisher{CRC press}.
\end{bbook}
\endbibitem

\bibitem[\protect\citeauthoryear{Hastie and
  Tibshirani}{1993}]{hastie1993varying}
\begin{barticle}[author]
\bauthor{\bsnm{Hastie},~\bfnm{Trevor~J}\binits{T.~J.}} \AND
  \bauthor{\bsnm{Tibshirani},~\bfnm{Robert~J}\binits{R.~J.}}
(\byear{1993}).
\btitle{Varying-coefficient models}.
\bjournal{Journal of the Royal Statistical Society: Series B (Methodological)}
\bvolume{55}
\bpages{757--796}.
\end{barticle}
\endbibitem

\bibitem[\protect\citeauthoryear{Hern{\'a}ndez, Pennington and
  Parnell}{2015}]{hernandez2015bayesian}
\begin{barticle}[author]
\bauthor{\bsnm{Hern{\'a}ndez},~\bfnm{Belinda}\binits{B.}},
  \bauthor{\bsnm{Pennington},~\bfnm{Stephen~R}\binits{S.~R.}} \AND
  \bauthor{\bsnm{Parnell},~\bfnm{Andrew~C}\binits{A.~C.}}
(\byear{2015}).
\btitle{Bayesian methods for proteomic biomarker development}.
\bjournal{EuPA Open Proteomics}
\bvolume{9}
\bpages{54--64}.
\end{barticle}
\endbibitem

\bibitem[\protect\citeauthoryear{Hern{\'a}ndez
  et~al.}{2018}]{hernandez2018bayesian}
\begin{barticle}[author]
\bauthor{\bsnm{Hern{\'a}ndez},~\bfnm{Belinda}\binits{B.}},
  \bauthor{\bsnm{Raftery},~\bfnm{Adrian~E}\binits{A.~E.}},
  \bauthor{\bsnm{Pennington},~\bfnm{Stephen~R}\binits{S.~R.}} \AND
  \bauthor{\bsnm{Parnell},~\bfnm{Andrew~C}\binits{A.~C.}}
(\byear{2018}).
\btitle{{Bayesian additive regression trees using Bayesian model averaging}}.
\bjournal{Statistics and computing}
\bvolume{28}
\bpages{869--890}.
\end{barticle}
\endbibitem

\bibitem[\protect\citeauthoryear{Hill}{2011}]{hill2011bayesian}
\begin{barticle}[author]
\bauthor{\bsnm{Hill},~\bfnm{Jennifer~L}\binits{J.~L.}}
(\byear{2011}).
\btitle{Bayesian nonparametric modeling for causal inference}.
\bjournal{Journal of Computational and Graphical Statistics}
\bvolume{20}
\bpages{217--240}.
\end{barticle}
\endbibitem

\bibitem[\protect\citeauthoryear{Inglis, Parnell and
  Hurley}{2022}]{inglis2022visualizing}
\begin{barticle}[author]
\bauthor{\bsnm{Inglis},~\bfnm{Alan}\binits{A.}},
  \bauthor{\bsnm{Parnell},~\bfnm{Andrew}\binits{A.}} \AND
  \bauthor{\bsnm{Hurley},~\bfnm{Catherine~B}\binits{C.~B.}}
(\byear{2022}).
\btitle{Visualizing variable importance and variable interaction effects in
  machine learning models}.
\bjournal{Journal of Computational and Graphical Statistics}
\bvolume{31}
\bpages{766--778}.
\end{barticle}
\endbibitem

\bibitem[\protect\citeauthoryear{Kapelner and Bleich}{2016}]{JSSv070i04}
\begin{barticle}[author]
\bauthor{\bsnm{Kapelner},~\bfnm{Adam}\binits{A.}} \AND
  \bauthor{\bsnm{Bleich},~\bfnm{Justin}\binits{J.}}
(\byear{2016}).
\btitle{bartMachine: machine learning with {B}ayesian additive regression
  trees}.
\bjournal{Journal of Statistical Software}
\bvolume{70}
\bpages{1--40}.
\end{barticle}
\endbibitem

\bibitem[\protect\citeauthoryear{Kindo, Wang and
  Pe{\~n}a}{2016}]{kindo2016multinomial}
\begin{barticle}[author]
\bauthor{\bsnm{Kindo},~\bfnm{Bereket~P}\binits{B.~P.}},
  \bauthor{\bsnm{Wang},~\bfnm{Hao}\binits{H.}} \AND
  \bauthor{\bsnm{Pe{\~n}a},~\bfnm{Edsel~A}\binits{E.~A.}}
(\byear{2016}).
\btitle{Multinomial probit {B}ayesian additive regression trees}.
\bjournal{Stat}
\bvolume{5}
\bpages{119--131}.
\end{barticle}
\endbibitem

\bibitem[\protect\citeauthoryear{Lang and Brezger}{2004}]{lang2004bayesian}
\begin{barticle}[author]
\bauthor{\bsnm{Lang},~\bfnm{Stefan}\binits{S.}} \AND
  \bauthor{\bsnm{Brezger},~\bfnm{Andreas}\binits{A.}}
(\byear{2004}).
\btitle{Bayesian {P}-splines}.
\bjournal{Journal of Computational and Graphical Statistics}
\bvolume{13}
\bpages{183--212}.
\end{barticle}
\endbibitem

\bibitem[\protect\citeauthoryear{Leisch and Dimitriadou}{2021}]{mlbench}
\begin{bmanual}[author]
\bauthor{\bsnm{Leisch},~\bfnm{Friedrich}\binits{F.}} \AND
  \bauthor{\bsnm{Dimitriadou},~\bfnm{Evgenia}\binits{E.}}
(\byear{2021}).
\btitle{{mlbench: Machine Learning Benchmark Problems}}
\bnote{{R} package version 2.1-3}.
\end{bmanual}
\endbibitem

\bibitem[\protect\citeauthoryear{Linero}{2018}]{linero2018bayesian}
\begin{barticle}[author]
\bauthor{\bsnm{Linero},~\bfnm{Antonio~R}\binits{A.~R.}}
(\byear{2018}).
\btitle{Bayesian regression trees for high-dimensional prediction and variable
  selection}.
\bjournal{Journal of the American Statistical Association}
\bvolume{113}
\bpages{626--636}.
\end{barticle}
\endbibitem

\bibitem[\protect\citeauthoryear{Linero, Sinha and
  Lipsitz}{2020}]{linero2020semipar}
\begin{barticle}[author]
\bauthor{\bsnm{Linero},~\bfnm{Antonio~R}\binits{A.~R.}},
  \bauthor{\bsnm{Sinha},~\bfnm{Debajyoti}\binits{D.}} \AND
  \bauthor{\bsnm{Lipsitz},~\bfnm{Stuart~R}\binits{S.~R.}}
(\byear{2020}).
\btitle{Semiparametric mixed-scale models using shared {B}ayesian forests}.
\bjournal{Biometrics}
\bvolume{76}
\bpages{131--144}.
\end{barticle}
\endbibitem

\bibitem[\protect\citeauthoryear{Linero and Yang}{2018}]{lineroAnDyang2018}
\begin{barticle}[author]
\bauthor{\bsnm{Linero},~\bfnm{Antonio~R}\binits{A.~R.}} \AND
  \bauthor{\bsnm{Yang},~\bfnm{Yun}\binits{Y.}}
(\byear{2018}).
\btitle{Bayesian regression tree ensembles that adapt to smoothness and
  sparsity}.
\bjournal{Journal of the Royal Statistical Society: Series B (Statistical
  Methodology)}
\bvolume{80}
\bpages{1087--1110}.
\end{barticle}
\endbibitem

\bibitem[\protect\citeauthoryear{Linero et~al.}{2021}]{linero2021bayesian}
\begin{barticle}[author]
\bauthor{\bsnm{Linero},~\bfnm{Antonio~R}\binits{A.~R.}},
  \bauthor{\bsnm{Basak},~\bfnm{Piyali}\binits{P.}},
  \bauthor{\bsnm{Li},~\bfnm{Yinpu}\binits{Y.}} \AND
  \bauthor{\bsnm{Sinha},~\bfnm{Debajyoti}\binits{D.}}
(\byear{2021}).
\btitle{Bayesian survival tree ensembles with submodel shrinkage}.
\bjournal{Bayesian Analysis}
\bpages{1--24}.
\bnote{Advance publication}.
\end{barticle}
\endbibitem

\bibitem[\protect\citeauthoryear{Loh}{2009}]{loh2009improving}
\begin{barticle}[author]
\bauthor{\bsnm{Loh},~\bfnm{Wei-Yin}\binits{W.-Y.}}
(\byear{2009}).
\btitle{{Improving the precision of classification trees}}.
\bjournal{The Annals of Applied Statistics}
\bvolume{3}
\bpages{1710--1737}.
\end{barticle}
\endbibitem

\bibitem[\protect\citeauthoryear{Martin et~al.}{2000}]{martin2000effective}
\begin{barticle}[author]
\bauthor{\bsnm{Martin},~\bfnm{Michael~O}\binits{M.~O.}},
  \bauthor{\bsnm{Mullis},~\bfnm{Ina~VS}\binits{I.~V.}},
  \bauthor{\bsnm{Gregory},~\bfnm{Kelvin~D}\binits{K.~D.}},
  \bauthor{\bsnm{Hoyle},~\bfnm{Craig}\binits{C.}} \AND
  \bauthor{\bsnm{Shen},~\bfnm{Ce}\binits{C.}}
(\byear{2000}).
\btitle{Effective schools in science and mathematics}.
\bjournal{IEA’s third international mathematics and science study, IEA:
  Chestnut Hill, MA}.
\end{barticle}
\endbibitem

\bibitem[\protect\citeauthoryear{McCullagh and
  Nelder}{1989}]{mccullagh1989generalized}
\begin{bbook}[author]
\bauthor{\bsnm{McCullagh},~\bfnm{P.}\binits{P.}} \AND
  \bauthor{\bsnm{Nelder},~\bfnm{J.~A.}\binits{J.~A.}}
(\byear{1989}).
\btitle{Generalized Linear Models},
\bedition{2} ed.
\bseries{Chapman and Hall/CRC Monographs on Statistics and Applied Probability
  Series}.
\bpublisher{Chapman \& Hall}.
\end{bbook}
\endbibitem

\bibitem[\protect\citeauthoryear{McJames et~al.}{2023}]{mcjames2023bayesian}
\begin{barticle}[author]
\bauthor{\bsnm{McJames},~\bfnm{Nathan}\binits{N.}},
  \bauthor{\bsnm{Parnell},~\bfnm{Andrew}\binits{A.}},
  \bauthor{\bsnm{Goh},~\bfnm{Yong~Chen}\binits{Y.~C.}} \AND
  \bauthor{\bsnm{O'Shea},~\bfnm{Ann}\binits{A.}}
(\byear{2023}).
\btitle{Bayesian Causal Forests for Multivariate Outcomes: Application to
  {I}rish Data From an International Large Scale Education Assessment}.
\bjournal{arXiv preprint arXiv:2303.04874}.
\end{barticle}
\endbibitem

\bibitem[\protect\citeauthoryear{Mohammadpour, Shekarchizadeh and
  Kalantarrashidi}{2015}]{mohammadpour2015multilevel}
\begin{barticle}[author]
\bauthor{\bsnm{Mohammadpour},~\bfnm{Ebrahim}\binits{E.}},
  \bauthor{\bsnm{Shekarchizadeh},~\bfnm{Ahmadreza}\binits{A.}} \AND
  \bauthor{\bsnm{Kalantarrashidi},~\bfnm{Shojae~Aldin}\binits{S.~A.}}
(\byear{2015}).
\btitle{{Multilevel modeling of science achievement in the TIMSS participating
  countries}}.
\bjournal{The Journal of Educational Research}
\bvolume{108}
\bpages{449--464}.
\end{barticle}
\endbibitem

\bibitem[\protect\citeauthoryear{Mullis et~al.}{2020}]{mullis2020timss}
\begin{btechreport}[author]
\bauthor{\bsnm{Mullis},~\bfnm{IVS}\binits{I.}},
  \bauthor{\bsnm{Martin},~\bfnm{MO}\binits{M.}},
  \bauthor{\bsnm{Foy},~\bfnm{P}\binits{P.}},
  \bauthor{\bsnm{Kelly},~\bfnm{D}\binits{D.}} \AND
  \bauthor{\bsnm{Fishbein},~\bfnm{B}\binits{B.}}
(\byear{2020}).
\btitle{{TIMSS 2019 International Results in Mathematics and Science}}
\btype{Technical Report},
\bpublisher{TIMSS \& PIRLS International Study Center},
\baddress{Boston, USA}
\bnote{Retrieved from Boston College, TIMSS \& PIRLS International Study Center
  website:
  \url{https://timssandpirls.bc.edu/timss2019/international-results/}}.
\end{btechreport}
\endbibitem

\bibitem[\protect\citeauthoryear{Murray}{2021}]{murray2021log}
\begin{barticle}[author]
\bauthor{\bsnm{Murray},~\bfnm{Jared~S.}\binits{J.~S.}}
(\byear{2021}).
\btitle{Log-linear {B}ayesian Additive Regression Trees for multinomial
  logistic and count regression models}.
\bjournal{Journal of the American Statistical Association}
\bvolume{116}
\bpages{756--769}.
\end{barticle}
\endbibitem

\bibitem[\protect\citeauthoryear{Nelder and
  Wedderburn}{1972}]{nelder1972generalized}
\begin{barticle}[author]
\bauthor{\bsnm{Nelder},~\bfnm{John~Ashworth}\binits{J.~A.}} \AND
  \bauthor{\bsnm{Wedderburn},~\bfnm{Robert~WM}\binits{R.~W.}}
(\byear{1972}).
\btitle{Generalized linear models}.
\bjournal{Journal of the Royal Statistical Society: Series A (General)}
\bvolume{135}
\bpages{370--384}.
\end{barticle}
\endbibitem

\bibitem[\protect\citeauthoryear{Newman et~al.}{1998}]{uci_repo}
\begin{bmisc}[author]
\bauthor{\bsnm{Newman},~\bfnm{D.~J.}\binits{D.~J.}},
  \bauthor{\bsnm{Hettich},~\bfnm{S.}\binits{S.}},
  \bauthor{\bsnm{Blake},~\bfnm{C.~L.}\binits{C.~L.}} \AND
  \bauthor{\bsnm{Merz},~\bfnm{C.~J.}\binits{C.~J.}}
(\byear{1998}).
\btitle{{UCI Repository of Machine Learning Databases}}.
\end{bmisc}
\endbibitem

\bibitem[\protect\citeauthoryear{Prado, Moral and
  Parnell}{2021}]{prado2021bayesian}
\begin{barticle}[author]
\bauthor{\bsnm{Prado},~\bfnm{Estev{\~a}o~B}\binits{E.~B.}},
  \bauthor{\bsnm{Moral},~\bfnm{Rafael~A}\binits{R.~A.}} \AND
  \bauthor{\bsnm{Parnell},~\bfnm{Andrew~C}\binits{A.~C.}}
(\byear{2021}).
\btitle{{B}ayesian Additive Regression Trees with model trees}.
\bjournal{Statistics and Computing}
\bvolume{31}
\bpages{1--13}.
\end{barticle}
\endbibitem

\bibitem[\protect\citeauthoryear{Pratola
  et~al.}{2020}]{pratola2017heteroscedastic}
\begin{barticle}[author]
\bauthor{\bsnm{Pratola},~\bfnm{M.~T.}\binits{M.~T.}},
  \bauthor{\bsnm{Chipman},~\bfnm{H.~A.}\binits{H.~A.}},
  \bauthor{\bsnm{George},~\bfnm{E.~I.}\binits{E.~I.}} \AND
  \bauthor{\bsnm{McCulloch},~\bfnm{R.~E.}\binits{R.~E.}}
(\byear{2020}).
\btitle{Heteroscedastic {BART} via Multiplicative Regression Trees}.
\bjournal{Journal of Computational and Graphical Statistics}
\bvolume{29}
\bpages{405--417}.
\end{barticle}
\endbibitem

\bibitem[\protect\citeauthoryear{Reinsch}{1967}]{reinsch1967smoothing}
\begin{barticle}[author]
\bauthor{\bsnm{Reinsch},~\bfnm{Christian~H}\binits{C.~H.}}
(\byear{1967}).
\btitle{Smoothing by spline functions}.
\bjournal{Numerische Mathematik}
\bvolume{10}
\bpages{177--183}.
\end{barticle}
\endbibitem

\bibitem[\protect\citeauthoryear{Ro{\v{c}}kov{\'a} and
  Saha}{2019}]{rockova2019theory}
\begin{binproceedings}[author]
\bauthor{\bsnm{Ro{\v{c}}kov{\'a}},~\bfnm{Veronika}\binits{V.}} \AND
  \bauthor{\bsnm{Saha},~\bfnm{Enakshi}\binits{E.}}
(\byear{2019}).
\btitle{{On theory for BART}}.
In \bbooktitle{The 22nd International Conference on Artificial Intelligence and
  Statistics}
\bvolume{89}
\bpages{2839--2848}.
\bpublisher{PMLR}.
\end{binproceedings}
\endbibitem

\bibitem[\protect\citeauthoryear{Ro{\v{c}}kov{\'a} and van~der
  Pas}{2020}]{rockova2020posterior}
\begin{barticle}[author]
\bauthor{\bsnm{Ro{\v{c}}kov{\'a}},~\bfnm{Veronika}\binits{V.}} \AND
  \bauthor{\bparticle{van~der} \bsnm{Pas},~\bfnm{St{\'e}phanie}\binits{S.}}
(\byear{2020}).
\btitle{Posterior concentration for {B}ayesian regression trees and forests}.
\bjournal{The Annals of Statistics}
\bvolume{48}
\bpages{2108--2131}.
\end{barticle}
\endbibitem

\bibitem[\protect\citeauthoryear{Rutkowski
  et~al.}{2010}]{rutkowski2010international}
\begin{barticle}[author]
\bauthor{\bsnm{Rutkowski},~\bfnm{Leslie}\binits{L.}},
  \bauthor{\bsnm{Gonzalez},~\bfnm{Eugenio}\binits{E.}},
  \bauthor{\bsnm{Joncas},~\bfnm{Marc}\binits{M.}} \AND \bauthor{\bsnm{{von
  Davier}},~\bfnm{Matthias}\binits{M.}}
(\byear{2010}).
\btitle{International large-scale assessment data: issues in secondary analysis
  and reporting}.
\bjournal{Educational Researcher}
\bvolume{39}
\bpages{142--151}.
\end{barticle}
\endbibitem

\bibitem[\protect\citeauthoryear{Sarti et~al.}{2023}]{sarti2023bayesian}
\begin{barticle}[author]
\bauthor{\bsnm{Sarti},~\bfnm{Danilo~A}\binits{D.~A.}},
  \bauthor{\bsnm{Prado},~\bfnm{Estev{\~a}o~B}\binits{E.~B.}},
  \bauthor{\bsnm{Inglis},~\bfnm{Alan~N}\binits{A.~N.}},
  \bauthor{\bsnm{Dos~Santos},~\bfnm{Ant{\^o}nia~AL}\binits{A.~A.}},
  \bauthor{\bsnm{Hurley},~\bfnm{Catherine~B}\binits{C.~B.}},
  \bauthor{\bsnm{Moral},~\bfnm{Rafael~A}\binits{R.~A.}} \AND
  \bauthor{\bsnm{Parnell},~\bfnm{Andrew~C}\binits{A.~C.}}
(\byear{2023}).
\btitle{Bayesian additive regression trees for genotype by environment
  interaction models}.
\bjournal{The Annals of Applied Statistics}
\bvolume{17}
\bpages{1936--1957}.
\end{barticle}
\endbibitem

\bibitem[\protect\citeauthoryear{Sparapani, Spanbauer and
  McCulloch}{2021}]{BART_Rpkg}
\begin{bmanual}[author]
\bauthor{\bsnm{Sparapani},~\bfnm{Rodney}\binits{R.}},
  \bauthor{\bsnm{Spanbauer},~\bfnm{Charles}\binits{C.}} \AND
  \bauthor{\bsnm{McCulloch},~\bfnm{Robert}\binits{R.}}
(\byear{2021}).
\btitle{Nonparametric Machine Learning and Efficient Computation with
  {B}ayesian additive regression trees: The {BART} {R} Package}.
\end{bmanual}
\endbibitem

\bibitem[\protect\citeauthoryear{Sparapani
  et~al.}{2016}]{sparapani2016nonparametric}
\begin{barticle}[author]
\bauthor{\bsnm{Sparapani},~\bfnm{Rodney~A}\binits{R.~A.}},
  \bauthor{\bsnm{Logan},~\bfnm{Brent~R}\binits{B.~R.}},
  \bauthor{\bsnm{McCulloch},~\bfnm{Robert~E}\binits{R.~E.}} \AND
  \bauthor{\bsnm{Laud},~\bfnm{Purushottam~W}\binits{P.~W.}}
(\byear{2016}).
\btitle{{Nonparametric survival analysis using Bayesian additive regression
  trees (BART)}}.
\bjournal{Statistics in Medicine}
\bvolume{35}
\bpages{2741--2753}.
\end{barticle}
\endbibitem

\bibitem[\protect\citeauthoryear{Sparapani
  et~al.}{2020}]{sparapani2020nonparametric}
\begin{barticle}[author]
\bauthor{\bsnm{Sparapani},~\bfnm{Rodney}\binits{R.}},
  \bauthor{\bsnm{Logan},~\bfnm{Brent~R}\binits{B.~R.}},
  \bauthor{\bsnm{McCulloch},~\bfnm{Robert~E}\binits{R.~E.}} \AND
  \bauthor{\bsnm{Laud},~\bfnm{Purushottam~W}\binits{P.~W.}}
(\byear{2020}).
\btitle{Nonparametric competing risks analysis using {B}ayesian Additive
  Regression Trees}.
\bjournal{Statistical Methods in Medical Research}
\bvolume{29}
\bpages{57--77}.
\end{barticle}
\endbibitem

\bibitem[\protect\citeauthoryear{Starling et~al.}{2020}]{starling2020bart}
\begin{barticle}[author]
\bauthor{\bsnm{Starling},~\bfnm{Jennifer~E}\binits{J.~E.}},
  \bauthor{\bsnm{Murray},~\bfnm{Jared~S}\binits{J.~S.}},
  \bauthor{\bsnm{Carvalho},~\bfnm{Carlos~M}\binits{C.~M.}},
  \bauthor{\bsnm{Bukowski},~\bfnm{Radek~K}\binits{R.~K.}} \AND
  \bauthor{\bsnm{Scott},~\bfnm{James~G}\binits{J.~G.}}
(\byear{2020}).
\btitle{{BART} with targeted smoothing: an analysis of patient-specific
  stillbirth risk}.
\bjournal{The Annals of Applied Statistics}
\bvolume{14}
\bpages{28--50}.
\end{barticle}
\endbibitem

\bibitem[\protect\citeauthoryear{Stigler, Gallimore and
  Hiebert}{2000}]{stigler2000using}
\begin{barticle}[author]
\bauthor{\bsnm{Stigler},~\bfnm{James~W}\binits{J.~W.}},
  \bauthor{\bsnm{Gallimore},~\bfnm{Ronald}\binits{R.}} \AND
  \bauthor{\bsnm{Hiebert},~\bfnm{James}\binits{J.}}
(\byear{2000}).
\btitle{Using video surveys to compare classrooms and teaching across cultures:
  Examples and lessons from the {TIMSS} video studies}.
\bjournal{Educational Psychologist}
\bvolume{35}
\bpages{87--100}.
\end{barticle}
\endbibitem

\bibitem[\protect\citeauthoryear{Stigler and
  Hiebert}{1997}]{stigler1997understanding}
\begin{barticle}[author]
\bauthor{\bsnm{Stigler},~\bfnm{James~W}\binits{J.~W.}} \AND
  \bauthor{\bsnm{Hiebert},~\bfnm{James}\binits{J.}}
(\byear{1997}).
\btitle{Understanding and improving classroom mathematics instruction: An
  overview of the {TIMSS} video study}.
\bjournal{Phi Delta Kappan}
\bvolume{79}
\bpages{14}.
\end{barticle}
\endbibitem

\bibitem[\protect\citeauthoryear{Suk, Kim and Kang}{2021}]{suk2021hybridizing}
\begin{barticle}[author]
\bauthor{\bsnm{Suk},~\bfnm{Youmi}\binits{Y.}},
  \bauthor{\bsnm{Kim},~\bfnm{Jee-Seon}\binits{J.-S.}} \AND
  \bauthor{\bsnm{Kang},~\bfnm{Hyunseung}\binits{H.}}
(\byear{2021}).
\btitle{Hybridizing machine learning methods and finite mixture models for
  estimating heterogeneous treatment effects in latent classes}.
\bjournal{Journal of Educational and Behavioral Statistics}
\bvolume{46}
\bpages{323--347}.
\end{barticle}
\endbibitem

\bibitem[\protect\citeauthoryear{Tan and Roy}{2019}]{tan2019bayesian}
\begin{barticle}[author]
\bauthor{\bsnm{Tan},~\bfnm{Yaoyuan~Vincent}\binits{Y.~V.}} \AND
  \bauthor{\bsnm{Roy},~\bfnm{Jason}\binits{J.}}
(\byear{2019}).
\btitle{{Bayesian additive regression trees and the General BART model}}.
\bjournal{Statistics in Medicine}
\bvolume{38}
\bpages{5048--5069}.
\end{barticle}
\endbibitem

\bibitem[\protect\citeauthoryear{Tang, Li and Liu}{2022}]{tang2022impact}
\begin{barticle}[author]
\bauthor{\bsnm{Tang},~\bfnm{AiBin}\binits{A.}},
  \bauthor{\bsnm{Li},~\bfnm{WenYe}\binits{W.}} \AND
  \bauthor{\bsnm{Liu},~\bfnm{Dawei}\binits{D.}}
(\byear{2022}).
\btitle{The Impact of Teachers' Professional Development in Science Pedagogy on
  Students' Achievement: Evidence from {TIMSS} 2019.}
\bjournal{Journal of Baltic Science Education}
\bvolume{21}
\bpages{258--274}.
\end{barticle}
\endbibitem

\bibitem[\protect\citeauthoryear{{R Core Team}}{2020}]{R}
\begin{bmanual}[author]
\bauthor{\bsnm{{R Core Team}}}
(\byear{2020}).
\btitle{R: A Language and Environment for Statistical Computing}
\bpublisher{R Foundation for Statistical Computing},
\baddress{Vienna, Austria}.
\end{bmanual}
\endbibitem

\bibitem[\protect\citeauthoryear{{Stan Development Team}}{2023}]{stan}
\begin{bmisc}[author]
\bauthor{\bsnm{{Stan Development Team}}}
(\byear{2023}).
\btitle{{RStan}: the {R} interface to {Stan}}.
\bnote{R package version 2.32.3}.
\end{bmisc}
\endbibitem

\bibitem[\protect\citeauthoryear{Weirich et~al.}{2014}]{weirich2014nested}
\begin{barticle}[author]
\bauthor{\bsnm{Weirich},~\bfnm{Sebastian}\binits{S.}},
  \bauthor{\bsnm{Haag},~\bfnm{Nicole}\binits{N.}},
  \bauthor{\bsnm{Hecht},~\bfnm{Martin}\binits{M.}},
  \bauthor{\bsnm{B{\"o}hme},~\bfnm{Katrin}\binits{K.}},
  \bauthor{\bsnm{Siegle},~\bfnm{Thilo}\binits{T.}} \AND
  \bauthor{\bsnm{L{\"u}dtke},~\bfnm{Oliver}\binits{O.}}
(\byear{2014}).
\btitle{Nested multiple imputation in large-scale assessments}.
\bjournal{Large-scale Assessments in Education}
\bvolume{2}
\bpages{1--18}.
\end{barticle}
\endbibitem

\bibitem[\protect\citeauthoryear{Wood}{2017}]{wood2017generalized}
\begin{bbook}[author]
\bauthor{\bsnm{Wood},~\bfnm{Simon~N}\binits{S.~N.}}
(\byear{2017}).
\btitle{Generalized Additive Models: An Introduction with R},
\bedition{2} ed.
\bpublisher{CRC press}.
\end{bbook}
\endbibitem

\bibitem[\protect\citeauthoryear{Wright and Ziegler}{2017}]{JSSv077i01}
\begin{barticle}[author]
\bauthor{\bsnm{Wright},~\bfnm{Marvin~N.}\binits{M.~N.}} \AND
  \bauthor{\bsnm{Ziegler},~\bfnm{Andreas}\binits{A.}}
(\byear{2017}).
\btitle{ranger: a fast implementation of random forests for high dimensional
  data in {C}\texttt{++} and \textsf{R}}.
\bjournal{Journal of Statistical Software}
\bvolume{77}
\bpages{1--17}.
\end{barticle}
\endbibitem

\bibitem[\protect\citeauthoryear{Zeldow, {Lo Re}~III and
  Roy}{2019}]{zeldow2019semiparametric}
\begin{barticle}[author]
\bauthor{\bsnm{Zeldow},~\bfnm{Bret}\binits{B.}}, \bauthor{\bsnm{{Lo
  Re}~III},~\bfnm{Vincent}\binits{V.}} \AND
  \bauthor{\bsnm{Roy},~\bfnm{Jason}\binits{J.}}
(\byear{2019}).
\btitle{A semiparametric modeling approach using {B}ayesian additive regression
  trees with an application to evaluate heterogeneous treatment effects}.
\bjournal{The Annals of Applied Statistics}
\bvolume{13}
\bpages{1989--2010}.
\end{barticle}
\endbibitem

\bibitem[\protect\citeauthoryear{Zhang and
  H{\"a}rdle}{2010}]{zhang2010bayesian}
\begin{barticle}[author]
\bauthor{\bsnm{Zhang},~\bfnm{Junni~L}\binits{J.~L.}} \AND
  \bauthor{\bsnm{H{\"a}rdle},~\bfnm{Wolfgang~K}\binits{W.~K.}}
(\byear{2010}).
\btitle{The {B}ayesian additive classification tree applied to credit risk
  modelling}.
\bjournal{Computational Statistics \& Data Analysis}
\bvolume{54}
\bpages{1197--1205}.
\end{barticle}
\endbibitem

\end{thebibliography}

%%%%%%%%%%%%%%%%%%%%%%%%%%%%%%%%%%%%%%%%%%%%%%
%% Example with single Appendix:            %%
%%%%%%%%%%%%%%%%%%%%%%%%%%%%%%%%%%%%%%%%%%%%%%
% \begin{appendix}
% \section*{Title}\label{appn} %% if no title is needed, leave empty \section*{}.
% Appendices should be provided in \verb|{appendix}| environment,
% before Acknowledgements.

% If there is only one appendix,
% then please refer to it in text as \ldots\ in the \hyperref[appn]{Appendix}.
% \end{appendix}
%%%%%%%%%%%%%%%%%%%%%%%%%%%%%%%%%%%%%%%%%%%%%%
%% Example with multiple Appendixes:        %%
%%%%%%%%%%%%%%%%%%%%%%%%%%%%%%%%%%%%%%%%%%%%%%
\newpage
\section*{Appendices}
\begin{appendix}
We present details of the implementation of the BART and CSP-BART algorithms in Appendix \ref{appendix_implementations}, details on the identifiability of the linear coefficients in CSP-BART in Appendix \ref{appendix_identifiability_csp}, details on the covariates identified by the BART-based variable-screening step for the TIMSS 2019 application in Section \ref{section_timss2019} and additional results which account for missing values in Appendix \ref{appendix_timss_dataset}, and details of another application to the well-known Pima Indians Diabetes dataset in Appendix \ref{appendix_pima}. Finally, details of further extensions to the CSP-BART model to accommodate higher-order effects in the linear predictor are outlined in Appendix~\ref{appendix_extra_linear}.

\section{Algorithmic implementations}
\label{appendix_implementations}
We first describe an implementation of the standard BART algorithm in Appendix \ref{appendix_BART}, then describe the implementation of our CSP-BART proposal in Appendix \ref{appendix_semiBART}, and finally compare the run-times of different implementations of BART, SSP-BART, and CSP-BART in Appendix \ref{appendix_runtimes}.

\subsection{BART implementation}
\label{appendix_BART}
\setcounter{table}{0}
\setcounter{figure}{0}
\setcounter{algorithm}{0}
\renewcommand{\thetable}{\Alph{section}.\arabic{table}}
\renewcommand{\thefigure}{\Alph{section}.\arabic{figure}}
\renewcommand{\thealgorithm}{\Alph{section}.\arabic{algorithm}}

In this Section, we provide the mathematical details of the BART model following \citet{prado2021bayesian} and \citet{tan2019bayesian}, which can be written as
\[y_{i}\given\mathbf{x}_{i}, \bm{\mathcal{M}}, \bm{\mathcal{T}}, \sigma^{2}   \sim \mbox{N}\left( \sum_{t = 1}^{T} g\left(\mathbf{x}_{i}, \bm{\mathcal{M}}_{t}, \mathcal{T}_{t}\right), \sigma^{2} \right),\]
where $g(\cdot) = \mu_{t\ell}$ is a function which returns predicted values, given the design matrix $\mathbf{X}$ and tree structure $\mathcal{T}_{t}$, and $\bm{\mathcal{M}}_{t} = \left( \mu_{t1}, \ldots, \mu_{t b_{t}} \right)$ is a vector comprising the predicted values from the $b_{t}$ terminal nodes of tree $t$. To obtain the full conditionals for $\mu_{t\ell}$ and $\sigma^{2}$, \citet{chipman2010bart} assume the following priors
\begin{align*}
\mu_{t\ell}\given\mathcal{T}_{t}  & \sim \mbox{N}\left(0, \sigma^{2}_{\mu}\right),\\
\sigma^{2} & \sim \mbox{IG}\left(\nu/2, \nu\lambda/2\right),
\end{align*}
where $\sigma_{\mu} = 0.5k/\sqrt{T}$, with $k \in \lbrack1, 3\rbrack$. To control the tree structure/depth, \citet{chipman2010bart} place the following prior on $\mathcal{T}_{t}$:
\[p\left(\mathcal{T}_{t}\right) = \prod_{\ell^\prime \in \mathcal{L}_{I}^{(t)}} \left\lbrack \eta \left(1 + d_{t {\ell^\prime}}\right)^{-\zeta} \right\rbrack \times \prod_{\ell \in \mathcal{L}_{T}^{(t)}} \left\lbrack 1 - \eta \left(1 + d_{t \ell}\right)^{-\zeta} \right\rbrack,\]
where $d_{t{\ell^\prime}}$ and $d_{t\ell}$ respectively denote the depth of an internal node $\ell^\prime$ and a terminal node $\ell$, $\eta \in (0,1)$ and $\zeta \ge 0$ are hyperparameters that control the shape of the tree, and $\mathcal{L}_{I}^{(t)}$ and $\mathcal{L}_{T}^{(t)}$ denote the sets of internal and terminal nodes of tree $t$, respectively. Hence, the joint posterior distribution can be written as
\begin{align*}
%\label{BART_joint_posterior}
    p\left(\bm{\mathcal{M}},\bm{\mathcal{T}} \given\mathbf{y}, \mathbf{X}, \sigma^{2}\right)  \propto \left\lbrack \prod_{t=1}^{T} \prod_{\ell = 1}^{b_{t}} \left\lbrack \prod_{i\colon \mathbf{x}_{i} \in \mathcal{P}_{t\ell}} p\left(y_{i}\given\mathbf{X}, \bm{\mathcal{M}}_{t}, \mathcal{T}_{t}, \sigma^{2}\right) \right\rbrack p\left(\bm{\mathcal{M}}_{t}\right) p\left(\mathcal{T}_{t}\right) \right\rbrack p\left(\sigma^{2}\right)\\
     \propto \left\lbrack \prod_{t=1}^{T} \prod_{\ell = 1}^{b_{t}} \left\lbrack \prod_{i\colon \mathbf{x}_{i} \in \mathcal{P}_{t\ell}} \mbox{N}\left(y_{i}\given[\Big] \sum_{t = 1}^{T} g\left(\mathbf{x}_{i}, \bm{\mathcal{M}}_{t}, \mathcal{T}_{t}\right), \sigma^{2}\right) \right\rbrack \mbox{N}\left(\mu_{t \ell}\given 0, \sigma^{2}_{\mu}\right) p\left(\mathcal{T}_{t}\right) \right\rbrack\nonumber \\
    \times\:\mbox{IG}\left(\sigma^{2}\given \nu/2, \nu\lambda/2\right),
\end{align*}
where $\mathcal{P}_{t \ell}$ is the set of splitting rules which define terminal node $\ell$ of tree $t$. It is possible to sample from this joint posterior in two steps, by substituting the response variable~$\mathbf{y}$ by the partial residuals $\mathbf{R}_{t} \equiv \mathbf{y} - \sum_{j \ne t}^{T} g(\mathbf{X}, \bm{\mathcal{M}}_{j}, \mathcal{T}_{j})$. We now outline the steps involved:
\begin{enumerate}[label=\roman*)]
    \item A new tree $\mathcal{T}_{t}^{\star}$ is proposed by a grow, prune, change, or swap move and then compared to its previous version $\mathcal{T}_{t}$ via
    \begin{align*}
        p\left(\mathcal{T}_{t}\given \mathbf{R}_{t}, \sigma^{2}\right) & \propto p\left(\mathbf{R}_{t}\given \sigma^{2}\right) p\left(\mathcal{T}_{t}\right)\\
        & \propto \prod_{\ell = 1}^{b_{t}} \left\lbrack \left(\frac{\sigma^{2}}{\sigma^{2}_{\mu} n_{t \ell} + \sigma^{2}}\right)^{1/2} \exp \left( \frac{\sigma^{2}_{\mu} \left( n_{t \ell} \bar{R}_{\ell} \right)^2}{2 \sigma^{2} \left(\sigma^{2}_{\mu} n_{t \ell} + \sigma^{2}\right)} \right) \right\rbrack p\left(\mathcal{T}_{t}\right),
    \end{align*}
     where $\bar{R}_{\ell} = n_{t \ell}^{-1}\sum_{i \in \mathcal{P}_{t \ell}} r_{i}$, $r_{i} \in \mathbf{R}_{t}$, and $n_{t \ell}$ is the number of observations belonging to terminal node $\ell$ of tree $t$. This sampling is conducted via a Metropolis-Hastings step.\smallskip
     
    \item Since $\mu_{t \ell}$ are i.i.d, their posterior is given by
    \begin{align}
    \label{full_conditional_mu}
    \mu_{t \ell}\given \mathcal{T}_{t}, \mathbf{R}_{t}, \sigma^{2} \sim  \mbox{N}\left(\frac{ \sigma^{-2}\sum_{i \in \mathcal{P}_{t \ell}} r_{i}}{n_{t \ell}/\sigma^{2} + \sigma^{-2}_{\mu}}, \frac{1}{n_{t \ell}/\sigma^{2} + \sigma^{-2}_{\mu}} \right).
    \end{align}
\end{enumerate}
Finally, the full conditional of $\sigma^{2}$ is given by
\begin{align}
\label{BART_full_conditional_sigma2}
\sigma^{2} \given  \mathbf{X}, \bm{\mathcal{M}}, \bm{\mathcal{T}}, \mathbf{y} \sim \mbox{IG}\left(\frac{n + \nu}{2}, \frac{\sum_{i=1}^{n}\left(y_{i} - \hat{y}_{i}\right)^{2} + \nu \lambda}{2}\right),
\end{align}
where $\hat{y}_{i} = \sum_{t = 1}^{T} g(\mathbf{x}_{i}, \bm{\mathcal{M}}_{t}, \mathcal{T}_{t})$ is the predicted response. In Algorithm \ref{BART_algorithm}, the full structure of the BART model is presented.
\begin{algorithm}[H]
% \setstretch{1.05}
\linespread{1.05}\selectfont
	\caption{BART model} 
	\begin{algorithmic}[1]
	\State \textbf{Input}: $\mathbf{y}$, $\mathbf{X}$, number of trees $T$, and number of MCMC iterations $M$.
	\State \textbf{Initialise}: $\{\mathcal{T}_{t}\}_{1}^{T}$ and set all hyperparameters of the prior distributions.
		\For {($m=1$ to $M$)}
			\For {($t=1$ to $T$)}
                \State Compute $\mathbf{R}_{t} = \mathbf{y} - \sum_{j \neq t}^{T} g\left(\mathbf{X}, \bm{\mathcal{M}}_{j}, \mathcal{T}_{j}\right)$.
        
                \State Propose a new tree $\mathcal{T}_{t}^{\star}$ by a grow, prune, change, or swap move;\newline\hspace*{9mm} iterate until a valid tree structure is obtained.
        
                \State Compare the current ($\mathcal{T}_{t}$) and proposed ($\mathcal{T}_{t}^{\star}$) trees via Metropolis-Hastings, with\smallskip
                
       \qquad\qquad$\alpha\left(\mathcal{T}_{t}, \mathcal{T}_{t}^{\star}\right) = 
        \mbox{min}\left \lbrace 1, 
        \frac{p\left(\mathcal{T}_{t}^{\star}\given \mathbf{R}_{t}, \sigma^{2}\right) q\left(\mathcal{T}_{t}^{\star} \rightarrow \mathcal{T}_{t}\right)}{p\left(\mathcal{T}_{t}\given \mathbf{R}_{t}, \sigma^{2}\right) q\left(\mathcal{T}_{t} \rightarrow \mathcal{T}_{t}^{\star}\right)} \right 
        \rbrace$.\smallskip
        
        \State Sample $u \sim \mbox{Uniform}\left(0,1\right)$: if $\alpha\left(\mathcal{T}_{t}, \mathcal{T}_{t}^{\star}\right) < u$, set $\mathcal{T}_{t} = \mathcal{T}_{t}$, otherwise set $\mathcal{T}_{t} = \mathcal{T}_{t}^{\star}$.
        
        \State Update all node-level parameters $\mu_{t \ell}$ via Equation \eqref{full_conditional_mu}, for $\ell = 1, \ldots, b_{t}$.
			\EndFor
			\State Update $\sigma^{2}$ via Equation \eqref{BART_full_conditional_sigma2}.
			\State Update the predicted response $\hat{\mathbf{y}}$.
		\EndFor
		\State \textbf{Output}: samples of the posterior distribution of $\bm{\mathcal{T}}$. 
	\end{algorithmic} 
\label{BART_algorithm}
\end{algorithm}

\subsection{Combined semi-parametric BART implementation}
\label{appendix_semiBART}

In this Section, we provide details for the implementation of the CSP-BART model, which can be written as
\[y_{i} \given \mathbf{x}_{1i}, \mathbf{x}_{2i}, \boldsymbol\beta, \bm{\mathcal{M}}, \bm{\mathcal{T}}, \sigma^{2} \sim \mbox{N}\left( \mathbf{x}_{1i} \boldsymbol\beta + \sum_{t = 1}^{T} g\left(\mathbf{x}_{2i}, \bm{\mathcal{M}}_{t}, \mathcal{T}_{t}\right), \sigma^{2} \right).\]
We recall that CSP-BART and SSP-BART \citep{zeldow2019semiparametric} differ in many aspects, with the latter assuming that i) $\mathbf{X}_{1}$ and $\mathbf{X}_{2}$ are disjoint matrices, such that only `single' grow/prune moves are considered, and ii) $\boldsymbol\beta \sim \mbox{MVN}(\mathbf{0}_{p_1}, \sigma^{2}_{\beta}\mathbf{I}_{p_1})$, where $\mathbf{0}_{p_1}$ and $\mathbf{I}_{p_1}$ respectively denote a vector of zeros and an identity matrix of appropriate dimension and $\sigma^2_b$ is a fixed, large scalar, such that the prior on $\boldsymbol\beta$ is uninformative. In contrast, CSP-BART i) allows $\mathbf{X}_{1}$ and $\mathbf{X}_{2}$ to share covariates, which is rendered valid by the novel double-grow and double-prune moves employed, and ii) assumes $\boldsymbol\beta \sim \mbox{MVN}(\mathbf{b}, \boldsymbol\Omega_{\beta})$, with the associated hyperprior $\boldsymbol\Omega_{\beta} \sim \mbox{IW}(\mathbf{V}, v)$. This hierarchical prior allows for more complex covariance structures for the linear predictor's parameters to be explicitly modelled. In terms of commonalities, both methods consider that $\sigma^{2} \sim \mbox{IG}(\nu/2, \nu\lambda/2)$ and define the partial residuals as $\mathbf{R}_{t} = \mathbf{y} - \mathbf{X}_{1}\boldsymbol\beta - \sum_{j \neq t}^{T} g(\mathbf{X}_{2}, \bm{\mathcal{M}}_{j}, \mathcal{T}_{j})$.

In Algorithm \ref{semiBART_algorithm}, the structure of CSP-BART is presented. Firstly, the response~and the design matrices $\mathbf{X}_{1}$ and $\mathbf{X}_{2}$ are specified, along with the number of trees (e.g., $T = 50$), number of MCMC iterations $M$, and all hyperparameters associated with~the priors for $\boldsymbol\beta$, $\boldsymbol\Omega_{\beta}$, $\mu_{t\ell}$, $\mathcal{T}_{t}$, and $\sigma^{2}$. Initially, all trees are set as stumps. Secondly, the parameter vector $\boldsymbol\beta$ and covariance matrix $\boldsymbol\Omega_{\beta}$ are updated. Thereafter, candidate trees ($\mathcal{T}_{t}^{\star}$) are sequentially proposed, one at a time --- via one of the four standard moves employed by SSP-BART, or one of the novel `double-grow' and `double-prune' moves --- and compared with their previous versions ($\mathcal{T}_{t}$) via a Metropolis-Hastings step. Later, the node-level parameters $\mu_{t\ell}$ are generated. Finally, the variance $\sigma^{2}$ is updated. For sufficiently large $M$, samples from the posterior distribution of the trees are obtained upon convergence.

Algorithm \ref{semiBART_algorithm} describes CSP-BART considering only fixed effects. However, we recall that the model can be extended to also incorporate random effects, such that
\[y_{i} \given \mathbf{x}_{1i}, \mathbf{z}_{i}, \mathbf{x}_{2i}, \boldsymbol\beta, \boldsymbol\gamma, \bm{\mathcal{M}}, \bm{\mathcal{T}}, \sigma^{2} \sim \mbox{N}\left( \mathbf{x}_{1i} \boldsymbol\beta + \mathbf{z}_{i} \boldsymbol\gamma + \sum_{t = 1}^{T} g\left(\mathbf{x}_{2i}, \bm{\mathcal{M}}_{t}, \mathcal{T}_{t}\right), \sigma^{2} \right),\]
where $\boldsymbol\gamma$ is the random effects vector of dimension $q$ and $\mathbf{z}_i$ denotes the $i$-th row of the associated design matrix $\mathbf{Z}$. For completeness, we reiterate that the same algorithm can be directly used to fit such a model, following the scheme outlined in Section \ref{random_effect_sec}.
\begin{algorithm}[H]
% \setstretch{1.05}
\linespread{1.05}\selectfont 
	\caption{CSP-BART model} 
	\begin{algorithmic}[1]
	\State \textbf{Input}: $\mathbf{y}$, $\mathbf{X}_{1}$, $\mathbf{X}_{2}$, number of trees $T$, and number of MCMC iterations $M$.
	\State \textbf{Initialise}: $\{\mathcal{T}_{t}\}_{1}^{T}$ and set all hyperparameters of the prior distributions.
		\For {($m=1$ to $M$)}
		    \State Update the parameter vector $\boldsymbol\beta$ via Equation \eqref{update_beta}.
		    \State Update the covariance matrix $\boldsymbol\Omega_{\beta}$ via Equation \eqref{update_omega}.
			\For {($t=1 \mbox{ to } T)$}
                \State Compute $\mathbf{R}_{t} = \mathbf{y} - \mathbf{X}_{1}\boldsymbol\beta - \sum_{j \neq t}^{T} g(\mathbf{X}_{2}, \bm{\mathcal{M}}_{j}, \mathcal{T}_{j})$.
        
                \State Propose a new tree $\mathcal{T}_{t}^{\star}$ by a grow, double-grow, prune, double-prune, change, or swap move;\newline\hspace*{9mm} iterate until a valid tree structure is obtained.
        
                \State Compare the current ($\mathcal{T}_{t}$) and proposed ($\mathcal{T}_{t}^{\star}$) trees via Metropolis-Hastings, with\smallskip
                
       \qquad\qquad$\alpha\left(\mathcal{T}_{t}, \mathcal{T}_{t}^{\star}\right) = 
        \mbox{min}\left \lbrace 1, 
        \frac{p\left(\mathcal{T}_{t}^{\star}\given \mathbf{R}_{t}, \sigma^{2}\right) q\left(\mathcal{T}_{t}^{\star} \rightarrow \mathcal{T}_{t}\right)}{p\left(\mathcal{T}_{t}\given \mathbf{R}_{t}, \sigma^{2}\right) q\left(\mathcal{T}_{t} \rightarrow \mathcal{T}_{t}^{\star}\right)} \right 
        \rbrace$.\smallskip
        
        \State Sample $u \sim \mbox{Uniform}\left(0,1\right)$: if $\alpha\left(\mathcal{T}_{t}, \mathcal{T}_{t}^{\star}\right) < u$, set $\mathcal{T}_{t} = \mathcal{T}_{t}$, otherwise set $\mathcal{T}_{t} = \mathcal{T}_{t}^{\star}$.
        
        \State Update all node-level parameters $\mu_{t \ell}$ via Equation \eqref{full_conditional_mu}, for $\ell = 1, \ldots, b_{t}$.
			\EndFor
			\State Update $\sigma^{2}$ via Equation \eqref{update_sigma_semi}.
			\State Update the predicted response $\hat{\mathbf{y}}$.
		\EndFor
		\State \textbf{Output}: samples of the posterior distribution of $\bm{\mathcal{T}}$. 
	\end{algorithmic} 
\label{semiBART_algorithm}
\end{algorithm}

The order in which our sampler updates the linear and tree components does not impact the fit, regardless of whether $\mathbf{X}_1 \subset \mathbf{X}_2$. The rationale for this is similar to that of the backfitting algorithm used to fit generalised additive models \citep{hastie1990generalized, wood2017generalized}. When fitting a GAM using the backfitting\footnote{Note that the \textsf{R} package \texttt{mgcv} \citep{wood2017generalized}, which provides functions to fit GAMs, uses other algorithms as alternatives/in addition to the vanilla backfitting algorithm introduced by \citet{hastie1990generalized}.} algorithm, the partial residuals related to the $j$-th component, which act like the response in the estimation of $\beta_j$, result from the difference between the response variable and the sum of all current parameter estimates times their corresponding basis functions, except for the $j$-th component (i.e., $R_j = y - \sum_{t \ne j} B_{t}(x) \hat{\beta}_{t}$, where $B_{t}(x)$ denotes the $t$-th basis function associated with predictor $x$). As long as all $\hat{\beta}_{t}$ converge/stabilise, the order in which they are updated is not important. 

In the context of CSP-BART, $f(x)$ is the sum of a linear predictor and a BART model, where the latter is prevented from estimating the effects specified in the former component. The linear predictor uses backfitting by virtue of being updated with respect to partial residuals $\mathbf{r} = \mathbf{y} - \sum_{t=1}^{T} g\left(\mathbf{X}_{2}, \bm{\mathcal{M}}_{t}, \mathcal{T}_{t}\right)$ when updating the linear component and $\mathbf{R}_{t} = \mathbf{y} - \mathbf{X}_{1}\boldsymbol\beta - \sum_{j \ne t}^{T} g\left(\mathbf{X}_{2}, \bm{\mathcal{M}}_{j}, \mathcal{T}_{j}\right)$ when updating the BART component. Thus, the order in which the two components (linear predictor and trees) or even the order in which the trees in the BART model are updated does not matter. For instance, if the trees were updated first instead, we recall that they would all initially be set as stumps. Furthermore, at the beginning of the iterative Bayesian backfitting MCMC algorithm (i.e., at the first iteration), the trees can (very poorly) estimate interaction effects of predictors in $\mathbf{X}_{1}$ and marginal effects involving predictors which are exclusive to $\mathbf{X}_{2}$, which should leave substantial variability for the linear term to explain. Finally, we note that the indifference to the order of the updates was confirmed by simulation experiments with the Friedman data and the TIMSS 2019 data, where both orderings yielded indistinguishable results.

In our implementation of the CSP-BART model, we convert categorical predictors into binary \emph{indicators} and also check whether there is more than one value available to the splitting variable. To elaborate on our point from Section \ref{CSP_BART_sec} about rejecting trees containing branches defined only by repeated splits on the same variable in $\mathbf{X}_1 \cap \mathbf{X}_2$, we give the following example which focuses specifically on categorical predictors and binary indicators thereof. Though~it is not possible to repeatedly split on the same binary indicator, this example emphasises that subsequent splits along the same branch on different binary indicators nonetheless associated with the same variable are not allowed when that variable is common to $\mathbf{X}_1$ and $\mathbf{X}_2$. Suppose $\mathbf{x}_{1i} = (x_{i1})$ contains a categorical variable of primary interpretational interest and $\mathbf{x}_{2i} = (x_{i1}, x_{i2})$ contains two variables of non-primary interpretational interest, where $x_{1} \in \{a, b, c\}$ and $x_{2} \in \mathbb{R}$. Here, we have
$$y_{i} = \mathbf{x}_{1i}\boldsymbol{\beta} + f(\mathbf{x}_{2i}) + \epsilon_{i},$$
where $\boldsymbol{\beta} = (\beta_{a}, \beta_{b})$ is the parameter vector which contains the coefficients associated with two levels ($a$ and $b$) of the variable of primary interest. Here, $c$ is the reference level and its effect could be calculated by a sum-to-zero contrast (i.e., $\beta_c = -(\beta_a + \beta_b)$. In our implementation, we convert the categorical predictor into 2 binary indicators, which we denote by $x.a$ and $x.b$, for its inclusion in $\mathbf{X}_1$, and also retain the column $x.c$ for its inclusion in $\mathbf{X}_2$. Figure \ref{tree_categorical} shows how a tree can also estimate the marginal effect of the levels `$a$', `$b$', and `$c$', respectively, which we seek to avoid. For instance, the path to the left-most nodes can be rewritten as $\mu_1 \mathds{1}(x.c = 0) \mathds{1}(x.b = 0)$ and $\mu_2 \mathds{1}(x.c = 0) \mathds{1}(x.b = 1)$, which implies that $\mu_1$ and $\mu_2$ are analogous to $\beta_a$ and $\beta_b$ because both $\mu_j$ and $\beta_j$ estimate the marginal effect of $x_1$. We point out, however, that our stricter checks on the tree-structure do not allow for trees like the one shown in Figure \ref{tree_categorical}; see Appendix \ref{appendix_identifiability_csp} for further details on the stricter checks.

% In Figure \ref{tree_categorical}, the node parameters $\mu_1$, $\mu_2$ and $\mu_c$ estimate the marginal effect of the levels `a', `b' and `c', respectively.

\begin{figure}[H]
\centering
\begin{forest}
for tree={
    grow=south, draw, minimum size=3ex, 
    inner sep=3pt, % control the overall tree size
    s sep=7mm,
    l sep=6mm
    }
[$\:x.c \leq 0\:$,
    [$\:x.b \leq 0\:$, edge label={node[midway,left, font=\footnotesize]{$\mbox{ TRUE }$}},
        [$\mu_{1}$, circle]
        [$\mu_{2}$, circle]
    ]
        [$\mu_{3}$, circle,  edge label={node[midway,right, font=\footnotesize]{$\mbox{ FALSE }$}},]
]
\end{forest}
\caption{An example of an invalid tree with binary indicators derived from the same categorical variable. \label{tree_categorical}}
\end{figure}

\subsection{Comparison of run-times}
\label{appendix_runtimes}

Our CSP-BART implementation is written entirely in \textsf{R}, so fair comparisons to the C\texttt{++} implementations of \texttt{BART}, \texttt{BCF}, and \texttt{VCBART}~are not straightforward. Nonetheless, to give an idea of the computational time of BART, SSP-BART, and CSP-BART, we compare them using our own \textsf{R} code to show the additional overheads from BART to SSP-BART and from SSP-BART to CSP-BART, by omitting the $\mathbf{X}_1$ matrix in the case of BART and omitting the covariates in $\mathbf{X}_1$ from $\mathbf{X}_2$ in the case of SSP-BART. In all experiments, the number of trees and posterior samples are $200$ and $1{,}000$, respectively.

Table \ref{tab_r_comparison} displays the computational time in minutes for BART, SSP-BART, and CSP-BART based on a pure \textsf{R} implementation. The times were obtained by running the algorithms on Friedman data with $n = 1{,}000$ observations and $d \in \{5, 10, 20, 30, 40, 50\}$ predictors. The relative additional run-times of SSP-BART over BART range from $5\%$ ($d = 5$) to $12\%$ ($d = 50$), while they range from $3\%$ ($d = 5$) to $7\%$ ($d = 50$) between CSP-BART and SSP-BART.\vspace{-1ex}
\begin{table}[H]
\caption{Run-times in minutes of BART, SSP-BART, and CSP-BART across 10 runs. The results are based on a pure \textsf{R} implementation of BART, which is the base for our CSP-BART implementation.\label{tab_r_comparison}} 
\centering
\begin{tabular}{cccccccc}
  \hline
 & $n$ & $d$ & BART & SSP & CSP & SSP/BART & CSP/SSP \\ 
  \hline
$1$ & $1{,}000$ & $5$ & $25.42$ & $26.78$ & $27.54$ & $1.05$ & $1.03$ \\ 
  $2$ & $1{,}000$ & $10$ & $25.71$ & $28.61$ & $29.68$ & $1.11$ & $1.04$ \\ 
  $3$ & $1{,}000$ & $20$ & $25.63$ & $27.28$ & $28.84$ & $1.06$ & $1.06$ \\ 
  $4$ & $1{,}000$ & $30$ & $25.26$ & $27.76$ & $28.54$ & $1.09$ & $1.03$ \\ 
  $5$ & $1{,}000$ & $40$ & $25.60$ & $28.66$ & $30.17$ & $1.12$ & $1.05$ \\ 
  $6$ & $1{,}000$ & $50$ & $25.97$ & $29.12$ & $31.08$ & $1.12$ & $1.07$ \\ 
   \hline
\end{tabular}
\end{table}\vspace{-1ex}%
We also compared the run-times of BART and SSP-BART using the \textsf{R} packages \texttt{BayesTree} and \texttt{semibart} \citep{zeldow2019semiparametric}. We stress that the latter's implementation is based on the C\texttt{++} code of the former, so a comparison of these package's run-times is reasonable. Table \ref{tab_cpp_comparison} shows relative additional overheads which vary from $16\%$ to $90\%$ depending on the number of covariates in the linear predictor of SSP-BART. As far as we checked the implementation of \texttt{semibart}, we believe the additional overheads are mostly attributable to a non-optimal sampling of the parameters in the linear predictor.
\begin{table}[H]
\caption{Run-times in minutes of BART and SSP-BART based on a C\texttt{++} implementation across 10 runs. The results for BART were obtained from the \textsf{R} package \texttt{BayesTree}, on which the implementation of SSP-BART is based. \label{tab_cpp_comparison}}
\centering
\begin{tabular}{cccccc}
  \hline
 & $n$ & $d$ & BART & SSP & SSP/BART \\ 
  \hline
$1$ & $1{,}000$ & $5$ & $14.76$ & $17.08$ & $1.16$ \\ 
  $2$ & $1{,}000$ & $10$ & $14.80$ & $17.87$ & $1.21$ \\ 
  $3$ & $1{,}000$ & $20$ & $14.73$ & $20.98$ & $1.42$ \\ 
  $4$ & $1{,}000$ & $30$ & $14.37$ & $22.95$ & $1.60$ \\ 
  $5$ & $1{,}000$ & $40$ & $14.54$ & $25.21$ & $1.73$ \\ 
  $6$ & $1{,}000$ & $50$ & $14.52$ & $27.76$ & $1.91$ \\ 
   \hline
\end{tabular}
\end{table}
\noindent We point out that we plan to implement CSP-BART in C\texttt{++}, which should speed up its run-time and make it more viable for real-world applications.

\section{Details on the identifiability of CSP-BART}
\label{appendix_identifiability_csp}
\setcounter{table}{0}
\setcounter{figure}{0}
\renewcommand{\thetable}{\Alph{section}.\arabic{table}}
\renewcommand{\thefigure}{\Alph{section}.\arabic{figure}}

\citet{dorie2022stan} claim that the sum of the parametric and non-parametric components of semi-parametric BART models which share covariates across components is identifiable but that the individual components are not. In light of this, we stress that non-identifiability of the BART component is of little concern but the identifiability of the main effects of primary interpretational interest in the linear predictor --- even when there are covariates common to both components --- is a key advantage of CSP-BART. To demonstrate the identifiability of CSP-BART in this regard, we begin with a simple example with four observations, two binary covariates, and one tree. We firstly show that the parameter estimates in the linear predictor are identifiable if we include the appropriate restrictions on the trees in the BART component (i.e., double-grow and double-prune moves \emph{along with} shrinking appropriate terminal node parameters to zero). Secondly, we show that this will generally be the case for any number of observations, any number of predictors, and any tree structures by expressing the CSP-BART model as a linear model, conditional on the tree topology. We subsequently elaborate on the identifiability even when the conditional posterior distributions of the terminal node parameters may allow some terminal node parameters to be similar. Finally, we discuss the situation when the number of trees increases.

The motivating example we provide to illustrate the identifiability of CSP-BART is as follows. Recall that the CSP-BART model can be written as 
\begin{equation}
\label{csp_bart_orig}
y_i \sim \mbox{N}\left(\mathbf{x}_{1i} \boldsymbol\beta + \sum_{t = 1}^{T} g\left(\mathbf{x}_{2i}, \bm{\mathcal{M}}_{t}, \mathcal{T}_{t}\right), \sigma^2\right).
\end{equation}
For simplicity, suppose that we have $T=1$ tree, $4$ observations, and two binary covariates, $x_1$ and $x_2$, where\vspace{-1em} 
\[\boldsymbol\beta = \lbrack\beta_1\rbrack, \quad
  \mathbf{X}_1 =\begin{blockarray}{cc}
    \matindex{$x_1$} & \\
    \begin{block}{(c)c}
 1\\
 1 \\
 0 \\
 0 \\
    \end{block}
  \end{blockarray}\negthickspace\mbox{\negthickspace,~ and}\quad
\mathbf{X}_2 =\begin{blockarray}{ccc}
    \matindex{$x_1$} & \matindex{$x_2$} & \\
    \begin{block}{(cc)c}
 1 & 0 \\
 1 & 1 \\
 0 & 0 \\
 0 & 1 \\
    \end{block}
  \end{blockarray}\negthickspace\negthickspace.\vspace{-1em}\]
Abusing notation slightly, we use $x_1$ and $x_2$ to denote the covariates inside the matrices $\mathbf{X}_1$ and $\mathbf{X}_2$. Here, we have $x_1 \in \{\mathbf{X}_1 \cap \mathbf{X}_2\}$ being a covariate of primary interpretational interest, a linear predictor without an intercept to model the main effect of $x_1$, and a single tree to model $x_2$ and the interaction between $x_1$ and $x_2$.

\subsection{Restricting the trees in the BART component}

In light of this simple example with a single tree, we show potential trees in Figure \ref{andrew_example_tree} in order to show the progression from a stump, through an invalid tree, to a valid tree obtained as the result of a double-grow move. The panel (a) of Figure \ref{andrew_example_tree} shows a stump. Such stumps are valid trees under CSP-BART, given that we omit the leading column of ones corresponding to an intercept from the linear component's design matrix $\mathbf{X}_1$. The tree in panel (b) illustrates a `single' grow move and exemplifies an invalid tree. As $x_1$ is common to both $\mathbf{X}_1$ and $\mathbf{X}_2$, $\mathcal{T}_{B}$ clearly would bring non-identifiability issues into the model in Equation \eqref{csp_bart_orig} as $\mu_{12}$ would be equivalent to $\beta_1$ (i.e., they would concomitantly model the main effect of $x_1$). If $\mathcal{T}_{B}$ instead split on $x_2$ rather than $x_1$, the single grow move would lead to a valid tree because $x_2$ is not in $\mathbf{X}_1$ (i.e., $x_2$ is exclusive to $\mathbf{X}_2$). On the other hand, the tree $\mathcal{T}_{C}$ in panel (c) exemplifies one of two potential outcomes of a `double-grow' move applied to the stump $\mathcal{T}_A$. When double-growing a tree, which only occurs when $x \in \{\mathbf{X}_{1} \cap \mathbf{X}_{2}\}$ is chosen to define a splitting rule for a stump, the branch on which to propose the second split is randomly sampled, with each branch being equally likely. Thus, a double-grow move could yield an alternative $\mathcal{T}_C$ where the second split is along the $x_1 > 0$ branch. In any case, here, $\mathcal{T}_C$ splits on the common $x_1$ at the top level \emph{and} $x_2$ at the second level along the $x_1 \le 0$ branch, such that it has terminal node parameters $\mu_{13}$ and $\mu_{14}$ at a depth of $2$ and $\mu_{15}$ at a depth of $1$. Notice that $\mu_{13}$ and $\mu_{14}$ estimate interaction effects between $x_1$ and $x_2$, while $\mu_{15}$ estimates the main effect of $x_1$ as it results from a splitting rule which splits on $x_1$ only. However, the main effect of $x_1$ is estimated in the linear predictor by $\beta_1$. To avoid estimating this effect which is already specified in the linear predictor, even after the double-grow move, the prior on $\mu_{15}$ is modified so that its posterior sample is shrunk to zero. As pointed out in Section \ref{CSP_BART_sec}, this is achieved by assuming that, \emph{a priori}, $\mu_{15} \sim \mbox{N}(0, \sigma^{2}_\mu \approx 0)$, which results in the posterior sample $\mu_{15}\approx 0$ illustrated in the corresponding terminal node in panel (c) of Figure \ref{andrew_example_tree}.\vspace{-1ex} 
\begin{figure}[H]
  \captionsetup[subfigure]{font=normalsize}
        \centering
        \hspace{-2cm}
                \begin{subfigure}[t]{100pt}
            \centering
            \caption[]%
            {$\mathcal{T}_A$}
            \begin{forest}
for tree={
    grow=south, draw, minimum size=3ex, 
    inner sep=3pt, % control the overall tree size
    s sep=7mm,
    l sep=6mm
    }
    [$\mu_{11}$, circle,  edge label={node[midway,right, font=\footnotesize]},]
\end{forest}
            %\label{fig:mean and std of net14}
        \end{subfigure}
        \hspace{1cm}
                \begin{subfigure}[t]{100pt}
            \centering
            \caption[]%
            {$\mathcal{T}_B$}
            \begin{forest}
for tree={
    grow=south, draw, minimum size=3ex, 
    inner sep=3pt, % control the overall tree size
    s sep=7mm,
    l sep=6mm
    }
[$\:x_{1} \le 0\:$,
    [$\mu_{11}$, circle,  edge label={node[midway,left, font=\footnotesize]{$\mbox{ TRUE }$}},]
    [$\mu_{12}$, circle,  edge label={node[midway,right, font=\footnotesize]{$\mbox{ FALSE }$}},]
]
\end{forest}
            %\label{fig:mean and std of net14}
        \end{subfigure}
        \hspace{1.5cm}
        \begin{subfigure}[t]{100pt}
            \centering
            \caption[]%
            {$\mathcal{T}_C$}
            \begin{forest}
for tree={
    grow=south, draw, minimum size=3ex, 
    inner sep=3pt, % control the overall tree size
    s sep=7mm,
    l sep=6mm
    }
[$\:x_{1} \le 0\:$,
    [$\:x_{2} \le 0\:$,
        [$\mu_{13}$, circle]
        [$\mu_{14}$, circle]
    ]
        [$\mu_{15} \approx 0$, circle]
]
\end{forest}
            %label{fig:mean and std of net14}
        \end{subfigure}
\caption{An example of the double-grow move in CSP-BART. Recall that the double-grow move applies only when a covariate $x \in \{\mathbf{X}_1 \cap \mathbf{X}_2\}$ is proposed for splitting a stump. The tree $\mathcal{T}_A$ is a stump, $\mathcal{T}_B$ shows the result of a single grow move (invalid in this case) using $x_1 \in \{\mathbf{X}_1 \cap \mathbf{X}_2\}$ applied to $\mathcal{T}_A$, and $\mathcal{T}_C$ illustrates an example of a tree obtained after CSP-BART's double-grow move. Note that tree index $t=1$ for all terminal node parameters $\mu_{t\ell}$, as we are assuming only $T=1$ tree in Equation \eqref{csp_bart_orig}.}
\label{andrew_example_tree}
\end{figure}\vspace{-1em}%
We therefore stress that the double-grow move --- which only applies when a splitting rule involving a covariate common to $\mathbf{X}_1$ and $\mathbf{X}_2$ is proposed for a stump --- consists of two operations which must be performed simultaneously: i) proposing a second splitting rule using any other variable, except the one used at the root node, and ii) shrinking the terminal node parameter of the terminal node on the opposing branch of the initial split by modifying its prior. Thus, by further splitting on $x_2$ and shrinking $\mu_{15}\approx 0$, we avoid having the main effect of $x_1$ be estimated twice (via $\beta_1$ in the linear term $x_1\beta$ and via $\mu_{15}$ through $\mu_{15}\mathds{1}(x_1 > 0)$). We also point out that there is no need for changes in the priors of $\mu_{13}$ and $\mu_{14}$ because they both model an (interaction) effect which is not specified in the linear predictor. Finally, we note that for all nodes except $\mu_{15}$, we follow \citet{chipman2010bart} by assuming \emph{a priori} that $\mu_{t\ell} \sim \mbox{N}(0, \sigma^{2}_\mu)$, where $\sigma_{\mu} = 0.5k/\sqrt{T}$, with $T$ denoting the number of trees and $k \in \lbrack1, 3\rbrack$.

To motivate the counterpart double-prune move, we recall $\mathcal{T}_{B}$ in panel (b) of Figure \ref{andrew_example_tree}, but now from a different perspective. Without loss of generality, instead of obtaining it via a single grow move as stated above, we can think of $\mathcal{T}_{B}$ as a tree which is obtained via a `single' prune move from panel (c) to (b). However, by pruning $\mathcal{T}_{C}$ once, the resulting tree ($\mathcal{T}_B$) and the linear predictor would both model the marginal effect of $x_1$, which would cause --- as per the single grow move --- a non-identifiability issue. To avoid this, we propose the `double-prune' move so that trees such as $\mathcal{T}_C$ are pruned twice. Consequently, $\mathcal{T}_{C}$ would become $\mathcal{T}_{A}$ after a double-prune, thus bypassing invalid trees like $\mathcal{T}_{B}$. Thus, the double-prune move is the counterpart of the double-grow move in the sense that it allows double-grow moves to be completely reversed. To clarify why $\mathcal{T}_C$ is a candidate for a double-prune move, we note that the double-prune move applies only to trees with exactly two splits (i.e., three terminal nodes) where the first split nearest the root of the tree splits on a variable common to $\mathbf{X}_1$ and $\mathbf{X}_2$ and the two resulting branches contain precisely one further split and no further splits, respectively. We also highlight that the double-prune move does not apply when the splitting rule at the root node of trees of this nature splits on a variable exclusive to $\mathbf{X}_2$, as subsequent terminal nodes will by definition either estimate main effects associated with covariates in $\mathbf{X}_2$ only or capture interactions between covariates in $\mathbf{X}_2$ and either those also in the linear predictor or others only in $\mathbf{X}_2$. Thus, were the splitting rule at the root node of $\mathcal{T}_{C}$ based on $x_2$ as opposed to $x_1$, no double-prune move would be needed since the resulting tree would be valid from an identifiability point of view, in the sense that it would avoid estimating main effects associated with covariates also in $\mathbf{X}_1$.

In Figure \ref{andrew_example_tree}, the tree $\mathcal{T}_B$ found after a single grow move would be considered invalid by CSP-BART. However, the double-grow move and the simultaneous shrinking of $\mu_{15}$ to zero ensures that $\mathcal{T}_C$ would be considered valid by CSP-BART. In other words, if the stump $\mathcal{T}_A$ is grown using a covariate $x \in \{\mathbf{X}_{1} \cap \mathbf{X}_{2}\}$, a double-grow as per $\mathcal{T}_C$ must be applied simultaneously with the shrinking of $\mu_{15}$ to zero. Consequently, we henceforth focus on $\mathcal{T}_C$ and provide further justification for these simultaneous operations by expressing the CSP-BART model as a linear model conditional on the tree topology.

\subsection{Expressing CSP-BART as a linear model}
\label{sec:CSPasLM}

With the fixed tree structure of $\mathcal{T}_C$, we can rewrite the model in Equation \eqref{csp_bart_orig} using the following equation:
\begin{equation}
\label{csp_bart_model}
\mathbf{y} = \underbrace{\mathbf{X}_1  \boldsymbol\beta}_\text{Linear predictor} + \underbrace{\mathbf{Z}_2 \bm{\mathcal{M}}_1}_\text{\quad BART\quad} + \quad\boldsymbol{\epsilon}
\end{equation}
where 
\[\mathbf{Z}_2 = \begin{bmatrix} 
 0 & 0 & 1 \\
 0 & 0 & 1\\
 1 & 0 & 0\\
 0 & 1 & 0
\end{bmatrix}, ~
\bm{\mathcal{M}}_1 = \begin{bmatrix} 
 \mu_{13}\\
 \mu_{14} \\
 \mu_{15}
\end{bmatrix},~\mbox{and}~
\mathbf{W} = \mathbf{Z}_2 ~ \mbox{diag}(\bm{\mathcal{M}}_1) = \begin{bmatrix} 
 0 & 0 & \mu_{15} \\
 0 & 0 & \mu_{15} \\
 \mu_{13} & 0 & 0\\
 0 & \mu_{14} & 0
\end{bmatrix},\]
subject to $\mu_{15}\approx 0$. The binary indicator matrix $\mathbf{Z}_2$ represents the tree structure of $\mathcal{T}_C$ in panel (c) of Figure \ref{andrew_example_tree} in terms of the allocations of observations in $\mathbf{X}_2$ to the terminal nodes, while the vector $\bm{\mathcal{M}}_1$ contains the terminal node parameters of the same tree. 

The model in Equation \eqref{csp_bart_model} will be identifiable whenever \emph{both} of the following conditions are satisfied:
\begin{enumerate}
\renewcommand{\labelenumi}{(\alph{enumi})}
    \item none of the columns in $\mathbf{Z}_2$ are created from a dichotomisation of a single covariate $x \in \{\mathbf{X}_1 \cap \mathbf{X}_2\}$.
    \item none of the $\mu_{1\ell}$ values in a tree are equal such that it would create an identical column of $\mathbf{X}_1$ through linear combinations of these values.
\end{enumerate} 
%For now, we turn to showing, in the context of the present purely categorical example, 
We now show that condition (a) can be addressed via certain modifications, while condition (b) will theoretically never occur provided our modifications have been made.

\subsubsection*{(a) Identifiability of the linear model representation}

Under condition (a), we look at the model in Equation \eqref{csp_bart_model} as $\mathbf{y} = \mathbf{Z} \mathbf{M} + \boldsymbol{\epsilon}$, where $\mathbf{Z} = \lbrack\mathbf{X}_1 ~ \mathbf{W}\rbrack$ and $\mathbf{M} = \lbrack\boldsymbol\beta ~ \bm{1}\rbrack$, with $\bm{1}$ denoting a vector of ones. In this example, for the sake of examining possible non-identifiability issues between the two components (linear predictor and BART) in the CSP-BART model, we point out that $\mathbf{X}_1$ is fixed by design and we otherwise assume that only the tree topology $\mathbf{Z}_2$ is known (i.e., both $\boldsymbol\beta$ and $\bm{\mathcal{M}}_1$ are unknown). Notably, we know from linear model theory that the least-squares solutions for the parameter estimates $\mathbf{B}=\lbrack\boldsymbol{\beta}~\bm{\mathcal{M}}_1\rbrack$ would exist and be unique if, and only if, $\mathbf{Z}$ is full rank (i.e., $(\mathbf{Z}^\top \mathbf{Z})^{-1}$ exists). Furthermore, we notice that $\mathbf{X}_1$ and $\mathbf{Z}_2$ have a column in common (the only column of $\mathbf{X}_1$ is identical to the final column of $\mathbf{Z}_2$). Consequently, the only column of $\mathbf{X}_1$ and the final column of $\mathbf{W}$ are linearly dependent, which means that $\mathbf{Z}$ is a rank-deficient matrix. However, we stress that columns of $\mathbf{Z}_2$ and $\mathbf{X}_1$ need not be strictly identical in order for this to occur; any dichotomisation of a covariate $x \in \{\mathbf{X}_1 \cap \mathbf{X}_2\}$ which yields a one-to-one mapping of a column in $\mathbf{X}_1$ will inhibit the inversion of $\mathbf{Z}^\top\mathbf{Z}$. To ensure the uniqueness of the solution, we recall that we address the linearly dependent column from $\mathbf{W}$ by changing the prior on terminal node parameters (such as $\mu_{15}$) whose indices in $\bm{\mathcal{M}}_1$ correspond to the indices of the associated columns in $\mathbf{Z}_2$, such that their posterior mean is shrunk to zero. In this way, the final column of $\mathbf{W}$ becomes redundant and can be removed, which in turn makes $\mathbf{Z}$ full rank. 

\subsubsection*{(b) Accounting for similar terminal node parameters}

The condition (b) is also from linear model theory, and again requires the unknown terminal node parameters in $\bm{\mathcal{M}}_1$ to be taken into account. Since we already deal with the columns of $\mathbf{W}$ and $\mathbf{X}_1$ being linearly dependent by changing the prior on $\mu_{15}$, we can now look for a linear combination between the two columns left in $\mathbf{W}$ (columns one and two) which, depending on $\bm{\mathcal{M}}_1$, could be linearly dependent with the column of $\mathbf{X}_1$. By calculating $\mathbf{Z}_2 \bm{\mathcal{M}}_1 = \lbrack\mu_{15} ~ \mu_{15} ~ \mu_{13} ~ \mu_{14}\rbrack^\top\approx \lbrack0 ~ 0 ~ \mu_{13} ~ \mu_{14}\rbrack^\top$ and $\mathbf{X}_1 \boldsymbol\beta = \lbrack\beta_1 ~ \beta_1 ~ 0 ~ 0\rbrack^\top$ as in Equation \eqref{csp_bart_model}, we notice that if $\mu_{13} = \mu_{14}=\mu^\star$, it would be equivalent to have
\[\mathbf{Z}^{\star}_2 = \begin{bmatrix} 
 0 & 1 \\
 0 & 1 \\
 1 & 0 \\
 1 & 0
\end{bmatrix}, ~
\bm{\mathcal{M}}^{\star}_1 = \begin{bmatrix} 
\mu_{14} \\
 \mu_{15}
\end{bmatrix}\approx\begin{bmatrix}
    \mu_{14} \\
    0
\end{bmatrix},~\mbox{and}~
\mathbf{W}^{\star} = \mathbf{Z}^{\star}_2 ~ \mbox{diag}(\bm{\mathcal{M}}^{\star}_1) \approx \begin{bmatrix} 
 0 & 0 \\
 0 & 0\\
 \mu^\star & 0\\
 \mu^\star & 0
\end{bmatrix},\]
and this would be an issue because now the first column in $\mathbf{W}^{\star}$ and the only column in $\mathbf{X}_1$ are linearly dependent --- recall that the second column of $\mathbf{W}^{\star}$ is not an issue because we already dealt with it by shrinking $\mu_{15}$ to zero. In particular, note that the first column of $\mathbf{Z}_2^\star$ arises from interchanging the $\{0, 1\}$ labels in the only column of $\mathbf{X}_1$. Consequently, there is a one-to-one mapping of the values $\{0,1\}$ in the only column of $\mathbf{X}_1$ with the values $\{\mu^\star, 0\}$ in the corresponding column of $\mathbf{W}^\star$. This, in turn, would lead to a model of the form $\mathbf{y}=\mathbf{Z}^\star\mathbf{M}^\star + \boldsymbol{\epsilon}$, with $\mathbf{M}^\star=\lbrack\boldsymbol{\beta} ~ \bm{1}^\star\rbrack$ and $\mathbf{Z}^\star=\lbrack\mathbf{X}_1 ~ \mathbf{W}^\star\rbrack$, where $\mathbf{Z}^\star$ is a now a rank-deficient matrix. This would be a clear example of non-identifiability between the two components (linear predictor and BART) of the CSP-BART model. However, we recall that the $\mu_{t \ell}$ parameters are assumed i.i.d \emph{a priori}, and that their conditional posterior distribution is given by
    \begin{align}
    \label{full_conditional_mu_identifiability}
    \mu_{t \ell}\given \mathcal{T}_{t}, \mathbf{R}_{t}, \sigma^{2} \sim  \mbox{N}\left(\frac{ \sigma^{-2}\sum_{i \in \mathcal{P}_{t \ell}} r_{i}}{n_{t \ell}/\sigma^{2} + \sigma^{-2}_{\mu}}, \frac{1}{n_{t \ell}/\sigma^{2} + \sigma^{-2}_{\mu}} \right),
    \end{align}
where $r_{i} \in \mathbf{R}_{t}$, $\mathbf{R}_{t} \equiv \mathbf{y} - \mathbf{X}_{1}\boldsymbol\beta - \sum_{j \neq t}^{T} g(\mathbf{X}_{2}, \bm{\mathcal{M}}_{j}, \mathcal{T}_{j})$, and $n_{t \ell}$ is the number of observations belonging to terminal node $\ell$ of tree $t$; see Appendix \ref{appendix_BART} for further details on the full conditionals of the BART model. Thus, in theory, $P(\mu_{th} = \mu_{tj}) = 0$, for all $h \ne j$, because the $\mu_{t\ell}$ parameters are updated via a \emph{continuous} (Gaussian) full conditional distribution (i.e., with probability $1$, $\mu_{13}$ and $\mu_{14}$ are different).

In practice, however, one could wonder about having \emph{similar} but not necessarily identical $\mu_{t\ell}$ parameters in the same tree, due to numeric precision. This could happen when the means of the partial residuals in each of the terminal nodes are similar \emph{and} $n_{t\ell}/\sigma^{2}$ is large, such that the variance term in Equation \eqref{full_conditional_mu_identifiability} is extremely small. Notably, it is extremely unlikely that the Metropolis-Hastings (MH) step in Algorithm \ref{semiBART_algorithm} --- which is used to sample from the full conditional of the trees --- will accept splits which lead to terminal nodes with similar means of the partial residuals, since the MH step filters out splitting rules which do not reduce the residual variance, as they are not supported by the likelihood. For example, if we assume that the variance term in Equation \eqref{full_conditional_mu_identifiability} goes to zero, it would imply that the $\mu_{t\ell}$ would approximately be the average of the partial residuals in each terminal node, which the MH step in turn says are substantially different. Moreover, in the context of the double-grow move, accepting spurious splits becomes even less likely, as \emph{both} splits must be accepted simultaneously by the MH step. Furthermore, given that the first split in a double-grow move applied to a stump is restricted to variables in $\mathbf{X}_1 \cap \mathbf{X}_2$, which are assumed to be of primary interpretational interest and linearly related to the response by definition, the interactions enforced by the double-grow move are guaranteed to be meaningful.

\subsection{Additional comments on the example}
\label{sec:extra_identifiability}

Overall, we point out that the example above illustrates how CSP-BART prevents non-identifiability issues through the double moves and the change in the prior of appropriate terminal node parameters. In addition, we note that the addition of the double moves to the transition kernel $q(\mathcal{T}_t\rightarrow \mathcal{T}_t^\star)$ in the MH step satisfies detailed balance (i.e., it yields a reversible Markov chain). That is to say, our modifications yield an identifiable model without compromising the validity of the MCMC sampler. We recall that the transition from $\mathcal{T}_t$ to $\mathcal{T}_t^\star$ via the grow move is reversed by the prune move in the vanilla BART, which also applies to the double-grow and double-prune moves in CSP-BART since the double-prune is the counterpart of the double-grow and vice-versa and the same rationale applies to the change and swap moves, which are themselves their own counterparts \citep{chipman1998bayesian}. 

Though we have focused more on the double moves and the modified prior on relevant $\mu_{t\ell}$ parameters in this example, we recall that we also place stricter checks on tree-structure validity after performing the (single) change and (single) swap moves to make sure the effects estimated by the resulting tree do not conflict with those in the linear predictor. After a new tree is proposed via a change or swap move, we check the ancestors of the affected nodes to ensure that no identifiability issues in line with those outlined and dealt with above arise. If the ancestors of the affected terminal nodes all split only on the same single variable which already has its marginal effect estimated in the linear predictor (i.e., if the same $x \in \mathbf{X}_1$ defines all splitting rules along the given branch), the tree is deemed invalid and thus automatically rejected, without being subjected to a MH step. Valid trees are subsequently either accepted or rejected via the MH step as usual. If no valid tree is found after some number of attempts to propose a new tree via a change or swap move, a stump is proposed instead.

Finally, we point out that the example above could be easily generalised in a number of directions, with some minor adjustments to the representations of $\mathbf{Z}_2$ and $\bm{\mathcal{M}}_{1}$. 
\begin{itemize}
    \item To include more trees, we note that an additive ensemble of trees can still be represented as a linear model conditional on the topology of the trees, as above. When $T>1$, the $\mathbf{Z}_2$ matrix is obtained by concatenating the binary allocations of observations to the terminal nodes of all trees across its columns, such that it retains the same number of rows, $n$, while all terminal node parameters are now gathered in the vector $\bm{\mathcal{M}}=\lbrack\bm{\mathcal{M}}_1,\ldots,\bm{\mathcal{M}}_T\rbrack$. Thereafter, the same conditions and modifications apply.
    \item It is also trivial to show that including additional predictors, whether as extra columns~\mbox{in~$\mathbf{X}_1$} only, $\mathbf{X}_2$ only, or both, would lead to the same representations and required modifications.
    \item All predictors in the example above are nominal and thus all splitting rules on their dummy representations are of the form $x \le 0$. Accounting for continuous or ordinal predictors confined exclusively to $\mathbf{X}_2$ is trivial; we simply allow for generic splitting rules of the form $x \le c$ and note that no confounding is possible, given that such predictors do not contribute to the linear component by construction. 

    To account for continuous predictors in $\mathbf{X}_1$, we note that the double-grow and double-prune moves are still required in order to bypass trees like $\mathcal{T}_B$ in Figure \ref{andrew_example_tree}. However, with a tree topology such as that of $\mathcal{T}_C$ thereafter, none of the continuous columns in $\mathbf{X}_1$ will be identical to any of those in the corresponding binary $\mathbf{Z}_2$ matrix, by definition. Moreover, no one-to-one mapping can exist between the continuous values of the corresponding column of $\mathbf{X}_1$ and any binary column of $\mathbf{Z}_2$. Nonetheless, condition (a) must be satisfied; i.e., to ensure the complete isolation of the effects of primary interest, $\mathbf{Z}_2$ should not contain columns obtained by dichotomising the values of a single variable $x \in \{ \mathbf{X}_1 \cap \mathbf{X}_2 \}$ via binary splitting rules of the form $x \le c$. We point out that, by construction, for both continuous and categorical predictors, $\mathbf{Z}_2$ will never contain columns of this nature due to the change on the prior for relevant terminal node parameters which shrinks them to zero as part of the double-grow move. Then, as per the purely categorical example above, it follows that condition (b) is automatically satisfied for continuous predictors also, provided those same modifications have been made.
\end{itemize}

Broadly speaking, the non-identifiability issues between the parametric (linear) and non-parametric (BART) components arise when the trees in the BART component estimate the effects specified in the linear predictor. Overall, our modifications aim to prevent linearly dependent columns between $\mathbf{X}_1$ and $\mathbf{Z}_2$ when $x \in \{\mathbf{X}_1 \cap \mathbf{X}_2\}$ is categorical, such that the linear model representation of CSP-BART ensures identifiability and guarantees, when such an $x$ is continuous, the complete isolation of the main effects in the linear predictor by effectively discarding columns in $\mathbf{Z}_2$ which arise from dichotomising columns of $\mathbf{X}_1$. In cases where mixed-type predictors are of primary interpretational interest, only the columns of $\mathbf{X}_1$ which relate to the categorical predictors are at risk of being linearly dependent with columns of $\mathbf{Z}_2$ and the same set of modifications address both identifiability conditions for both types of variable. Roughly speaking, our proposals can be interpreted as an adjustment to the prior over the set of possible tree structures such that a prior probability of zero is assigned to invalid trees which could bring non-identifiability issues into the CSP-BART model.

\section{Accounting for missing values in the TIMSS 2019 data}
\label{appendix_timss_dataset}
In Section \ref{section_timss2019}, we analysed a subset of the TIMSS 2019 data on Irish students at eighth grade level. To illustrate the novel CSP-BART, only one plausible value of the students’ mathematics scores was used, and sampling weights were not accounted for. Nonetheless, it would be necessary to consider all five mathematics scores along with sampling weights for a more complete analysis; see \citet{rutkowski2010international} and \citet{foy2017} for details. A more complete analysis would also account for the missing values in the additional covariates we discarded as part of this analysis.

We analysed a subset of the TIMSS 2019 data in Section \ref{section_timss2019} by identifying a set of $20$ predictors using a BART-based variable-screening step. By using the complete cases across these $20$ pre-selected predictors --- which include the three covariates of primary interest --- we were able to conduct a comparison between CSP-BART and a number of competing models which cannot accommodate missing values (namely, SSP-BART, VCBART, BCF) using an increased number of complete observations. However, other strategies for dealing with the missing values are available under CSP-BART, which can provide a better analysis of the TIMSS 2019 data when applying CSP-BART alone, outside of the context of the comparison of methods conducted in Section \ref{section_timss2019}. We discuss the rationale and results of the BART-based pre-screening step in greater detail in Appendix \ref{appendix_timss_screening} and present alternative approaches for accounting for missing values in the TIMSS 2019 data in CSP-BART in Appendix \ref{appendix_csp_missing}.

\subsection{BART-based pre-screening of the TIMSS 2019 dataset}
\label{appendix_timss_screening}
\setcounter{table}{0}
\setcounter{figure}{0}
\renewcommand{\thetable}{\Alph{section}.\arabic{table}}
\renewcommand{\thefigure}{\Alph{section}.\arabic{figure}}In Table \ref{TIMSS_covs}, we present the $20$ covariates that were pre-selected to demonstrate CSP-BART's performance relative to other BART-based competitors in Section \ref{section_timss2019}. These covariates were selected by identifying the $20$ variables used most often by a standard BART model fit to the complete cases of the TIMSS 2019 dataset. In the comparisons between CSP-BART, SSP-BART, and VCBART, $\mathbf{X}_{1}$ contained the covariates `BSDGEDUP', `BSBM42BA', and `BCDGDAS'. All $20$ covariates were included in $\mathbf{X}_{2}$ under CSP-BART, which is the matrix used by the BART component, but these three covariates of primary interpretational interest were excluded from the $\mathbf{X}_2$ matrix for the other methods. In the first comparison with BCF, $\mathbf{X}_{1}$ contained only the binarised version of the covariate `BCDGDAS'.

We performed the variable-screening step as we were required to work with complete cases only (i.e., no missing values) for the competing methods in particular. However, the number of complete cases ($1{,}448$) was quite low relative to the initial sample size ($4{,}118$). In an attempt to work with a larger number of observations of fewer variables, we first ran BART on the complete cases, to simply select predictors to subsequently carry out statistical analyses on a much bigger population of $3{,}224$. We now endeavour to explain the scarcity of complete observations in the original TIMSS 2019 data by describing its diverse sources.

As pointed out in Section \ref{sec_TIMSS} and Section \ref{section_timss2019}, the TIMSS 2019 data pertaining to eighth grade level students from Ireland comprised $4{,}118$ observations in total. This number corresponds to the number of students who were surveyed. In addition, $565$ of their teachers across $149$ associated schools were also surveyed.
We mention the number of students, teachers, and schools because the TIMSS 2019 data are split in a similar fashion. For instance, the school dataset has $149$ rows and $87\%$ of its $98$ columns have more than $5\%$ of observations missing. Similarly, $93\%$ of columns in the $565$-row teacher dataset have more than $5\%$ of observations missing. In the student dataset, a full $20\%$ of columns are completely missing, while only $5\%$ are fully observed. Consequently, if a teacher has some information missing, which is the case for most teachers here, all students of that teacher will have missing values when merging the data of the students with those of their teachers, given that individual teachers teach up to $56$ different students. Hence, what seems to be a minor missing value imputation in the $565$-row teacher dataset becomes a bigger issue when merging datasets. In addition, merging the data of students with the data of their school presents similar challenges for imputation.

In the context of large-scale assessment data, there are works which propose and use different imputation methods as alternative approaches for dealing with missing information in educational research settings \citep{foy2013technical, bouhlila2013multiple, weirich2014nested}. However, many of the predictors in the merged TIMSS 2019 data we analyse that could, in theory, benefit most from imputation methods tend to have levels of missingness in excess of $90\%$, which renders imputation infeasible. In our case, we found in the student questionnaire that general questions related to the use of computers/tablets have the highest percentage of missing observations (i.e., close to $100\%$).
\citet{grilli2016exploiting}, who analysed TIMSS 2011 data\footnote{Unlike our analysis of the Irish TIMSS 2019 data which focuses on a single mathematics score and allows for interactions, \citet{grilli2016exploiting} analysed Italian TIMSS 2011 data using a multivariate mixed model that modelled marginal effects only.} using multivariate mixed models, also observed that some sources of information have high levels of missingness, particularly for student-related data and usually associated with questions related to family background. In their analysis with only complete cases pertaining to fourth grade level Italian students, $3{,}741$ observations out of $4{,}200$ were used, which is roughly in line with the proportion used in our analysis ($3{,}224$ out of $4{,}118$ observations)). Furthermore, both works consider approximately the same number of predictors (specifically, $p=20$, after the BART-based variable-screening step here, $p=29$ in \citet{grilli2016exploiting}). 

\begin{landscape}
\begin{table}[H]
\caption{Covariates pre-selected by the BART-based variable-screening step for the analysis of the TIMSS 2019 dataset. All covariates are either binary, nominal or ordinal. Those marked with an asterisk ($\star$) were identified as being of primary interpretational interest \emph{prior to} this screening, though they are also used by the standard BART model. The `Source' column indicates whether the covariate arises from a student questionnaire or from a questionnaire completed by the school principal. Notably, no covariates sourced from surveys of teachers are included.\label{TIMSS_covs}}\vskip\smallskipamount
\centering
\extrarowheight 4.5pt
\setlength{\tabcolsep}{4.5pt}
\begin{tabular}{lll} 
\toprule
\textbf{Covariate} & \textbf{Label} & \textbf{Source}\\
\midrule 
BSBG10      & How often student is absent from school & Student \\
BSBM26AA    & How often teacher gives homework & Student\\
BSBM17A     & Does the student know what the teacher expects them to do? & Student \\
BSBG11B     & How often student feels hungry & Student \\
BSBM20I     & Does the student think it is important to do well in mathematics? & Student\\
BSBG05A     & Does the student have a tablet or a computer at home? & Student  \\
BCBG13BC    & Does the school have library resources? & Principal \\
BSBG13D     & Does the student think that teachers are fair in their school? & Student \\
BCBG13AD    & Does the school have heating systems? & Principal \\
BCBG06C     & How many instructional days in one calendar week does the school have? & Principal \\
BSBG05I     & Country-specific indicator of wealth & Student  \\
BSBM19A     & Does the student do usually well in mathematics? & Student  \\
BSBM42BA\textsuperscript{($\star$)} & Minutes spent on homework & Student \\
BSBM43BA    & For how many of the last $12$ months has the student attended extra lessons or tutoring in mathematics? & Student\\
BCDGDAS\textsuperscript{($\star$)} & Does the school have discipline problems? & Principal \\
BSBG11A     & How often does the student feel tired? & Student \\
BCBG13AB    & Does the school have shortage of supplies? & Principal \\
BCDGMRS     & Are the instructions affected by the material resources shortage? & Principal \\
BSDGEDUP\textsuperscript{($\star$)} & Parents' highest education level & Student \\
ITSEX       & Sex of student & Student \\ 
\bottomrule
\end{tabular}
\end{table}
\end{landscape}

We stress that if there were no missing values in the TIMSS 2019 data, there would have been no need to pre-select the predictors in Table \ref{TIMSS_covs}. We recall that a key selling point of~a BART-based model like CSP-BART is its ability to handle a large number of covariates without pre-specification. Furthermore, its BART component can be adapted to efficiently handle a larger number of predictors through specifying a Dirichlet prior \citep{linero2018bayesian} on the splitting probabilities, so that more important predictors can be favoured over those with little or no influence on the response. We termed this extension the `CSP-DART' model in Section \ref{CSP_BART_sec} and noted that it can be specified in our CSP-BART software implementation via the argument \texttt{sparse = TRUE}. Furthermore, tree-based methods can also be adapted to deal with missing observations. Different approaches have been proposed by \citep{loh2009improving}, depending on the nature of the predictor; missing observations are assigned to the left child node for numerical predictors and treated as a new category on which the trees can split for categorical predictors. Examples of the adoption of this strategy for categorical predictors are provided by the \texttt{bartMachine} \citep{JSSv070i04} and \texttt{ranger} packages \citep{JSSv077i01}, which offer implementations of BART and random forests, respectively. 
 
Such strategies, however, are limited from the point of view of CSP-BART in that they can only accommodate missing values in the trees. The linear predictor of CSP-BART cannot accommodate predictors with missing values, as it would not be possible to obtain the parameter estimates associated with the variables of primary interpretational interest without substantial modifications to the model and its MCMC scheme. Furthermore, there are $316$ observations with missing values across the three variables in the linear predictor (i.e., parents' education level, minutes spent on homework, and school discipline problems), which results in $3{,}802 = 4{,}118 - 316$ ``partially complete cases'' of $270$ variables after removing the missing values for the variables in $\mathbf{X}_1$ only. Notably, we use $3{,}224$ observations of $20$ variables in our analysis in Section \ref{section_timss2019}, which represents approximately $85\%$ of the aforementioned $3{,}802$ ``partially complete cases''. While considering a situation in which the covariates of primary interest are completely observed but missing values in $\mathbf{X}_2$ are preserved and accounted for is feasible for CSP-BART, it is not doable for BCF, SSP-BART, and VCBART. 

For these reasons, we opted to work in Section \ref{section_timss2019} with complete cases, as per \citet{grilli2016exploiting}, across the full set of pre-screened covariates only, to facilitate a fair comparison of the competing methodologies' performance on the TIMSS 2019 data. Consequently, we defer additional analyses of these data which account for missingness in different ways using only CSP-DART to Appendix \ref{appendix_csp_missing}.

\subsection{Additional analyses of the TIMSS 2019 data}\label{appendix_csp_missing} 

By way of addressing potential concerns about the data pre-processing performed in Section \ref{section_timss2019}, we now consider two alternative analysis approaches for the TIMSS 2019 data which make use of more predictors in $\mathbf{X}_2$, both by analysing only the $1{,}448$ complete cases and by adapting our model to accommodate missing values in its trees using the strategies of \citet{loh2009improving}, without the BART-based pre-selection step in either case. Given the larger number of predictors in each case, we employ the CSP-DART rather than CSP-BART model for the sake of efficiency for both analyses. We recall that the BART-based variable-screening step is arguably a sub-optimal way to analyse the TIMSS 2019 data and was only performed in order to facilitate a comparison between CSP-BART and its competitors (i.e., BCF, SSP-BART, and VCBART), whose available software implementations cannot handle missing values in their tree structures. With this in mind, we recap the results from Section \ref{section_timss2019} of CSP-BART obtained in conjunction with the variable-screening step when presenting the results of the additional analyses using CSP-DART in Table \ref{TIMSS_additional_results}. It is worth noting that the results obtained by adapting CSP-DART to handle missing values are nonetheless based on cases where the covariates of primary interest in $\mathbf{X}_1$ are completely observed, as CSP-BART cannot accommodate missing values in its parametric linear component.

\begin{landscape}
\begin{table}[H]%
\centering
\caption{Comparison of i) the results shown in Table \ref{tab2_VCBART_semiBART} of CSP-BART applied to the complete cases of the pre-selected subset of 20 variables most used by BART (PS), ii) CSP-DART applied to the complete cases with more predictors (CC), and iii) CSP-DART modified to handle missing values in the trees, using cases where the primary interest variables are completely observed and a greater number of predictors (MT). The numbers in parentheses denote the number of observations, $n$, and predictors in $\mathbf{X}_2$, $p_2$, used to fit the model in each case. As per Table \ref{tab2_VCBART_semiBART}, boldface font is used to highlight the cases where the 90\% CI does not contain zero.\label{TIMSS_additional_results}}\vskip\smallskipamount
\setlength{\tabcolsep}{4.125pt}
\extrarowheight 6.5pt
\begin{tabular*}{580pt}{llccc}
\toprule
\multirow{2}{*}{Covariate} & \multirow{2}{*}{Category} & CSP-BART: PS  & CSP-DART: CC & CSP-DART: MT\\ 
&&($n=3{,}224$, $p_2=20$)&($n=1{,}448$, $p_2=250$)&($n=3{,}802$, $p_2=250$)\\
\midrule
\multirow{6}{*}{Parents' education level} 
 & University or higher                 & $\mathbf{20.94 (14.94; 26.58)}$   & $5.67 (-1.23; 11.85)$             & $\mathbf{4.04 (1.15; 6.86)}$ \\ 
 & Post-secondary but not university    & $\mathbf{18.77 (13.05; 24.40)}$   & $\mathbf{11.83 (4.52; 18.78)}$    & $\mathbf{11.22 (8.34; 13.93)}$\\ 
 & Upper secondary                      & $-3.62 (-10.70; 3.80)$            & $4.55 (-3.18; 12.69)$             & $1.64 (-1.31; 4.85)$ \\ 
 & Lower secondary                      & $\mathbf{-11.84 (-22.86; -1.31)}$ & $-6.38 (-17.85; 3.91)$            & $-3.15 (-8.41; 1.88)$\\ 
 & Primary, secondary, or no school      & $\mathbf{-21.08 (-36.62; -5.38)}$ & $-16.18 (-39.61; 14.07)$          & $\mathbf{-12.01 (-18.61; -5.62)}$\\ 
 & Not informed                         & $-3.17 (-9.34; 2.61)$             & $-0.17 (-6.77; 7.01)$             & $-1.64 (-4.04; 0.81)$ \\
\hline
\multirow{6}{*}{Minutes spent on homework} 
 & No homework              & $\mathbf{-24.92 (-45.88; -2.19)}$ & $-3.68 (-26.76; 31.46)$           & $\mathbf{-33.29 (-43.86; -22.51)}$ \\ 
 & $1$ to $15$ minutes      & $1.62 (-6.57; 11.95)$             & $3.27 (-8.36; 11.01)$             & $\mathbf{4.87 (1.42; 8.49)}$ \\ 
 & $16$ to $30$ minutes     & $6.04 (-1.35; 12.83)$             & $5.85 (-6.04; 14.01)$             & $\mathbf{6.39 (2.97; 9.66)}$ \\
 & $31$ to $60$ minutes     & $7.83 (-1.30; 15.88)$             & $10.68 (-1.63; 19.68)$            & $\mathbf{6.11 (2.47; 9.84)}$ \\
 & $61$ to $90$ minutes     & $9.00 (-3.36; 23.19)$             & $6.51 (-7.69; 19.58)$             & $\mathbf{13.55 (7.56; 20.38)}$ \\
 & More than $90$ minutes   & $0.42 (-20.32; 20.57)$            & $-23.84 (-63.84; 21.53)$          & $2.11 (-6.42; 10.47)$ \\
\hline
\multirow{3}{*}{School discipline problems} 
 & Hardly any problems          & $\mathbf{14.52 (9.05; 19.98)}$         & $2.63 (-10.47; 13.84)$          & $\mathbf{5.37 (2.57; 8.21)}$\\
 & Minor problems               & $\mathbf{10.06 (4.68; 15.56)}$         & $-4.17 (-15.98; 7.58)$          & $\mathbf{4.40 (1.27; 7.39)}$\\
 & Moderate to severe problems  & $\mathbf{-24.58 (-33.51; -15.60)}$     & $1.87 (-21.16; 26.44)$          & $\mathbf{-9.80 (-15.21; -4.41)}$ \\
\bottomrule
\end{tabular*}
\end{table}
\end{landscape}

In Table \ref{TIMSS_additional_results}, we compare the results of CSP-DART applied to the complete cases (CC, $n=1{,}448$, $p_2 = 250$) and CSP-DART adapted to handle missing values in the trees while using the complete cases of the three covariates of primary interest (MT, $n=3{,}802$, $p_2 = 250$). As a reference for comparison, the first column of Table \ref{TIMSS_additional_results} exhibits the results already presented in Section \ref{section_timss2019} where CSP-BART is applied to the data pre-screened by BART (PS, $n=3{,}224$, $p_2=20$). In each case, $p_2$ only denotes the number of predictors in the $\mathbf{X}_2$ matrix available to the BART component; the same three covariates are of primary interpretational interest in each case. Notably, we do not use all $270$ predictors under the CC or MT results. In line with the discussion around identifiability issues presented in Appendix \ref{appendix_identifiability_csp}, we discarded some predictors which are perfectly collinear with the variables of primary interpretational interest, as well as variables for which one of the three categorical variables of primary interest is a one-to-one discretisation. We also offered advice in this regard in a footnote in Section \ref{CSP_BART_sec}. As one example of this phenomenon, we note that there are two variables which arise from two closely-related questions\footnote{In the 2019 School Context Data Almanac by Mathematics Achievement for the eighth grade level (\url{https://timss2019.org/international-database/}), these questions are ``School Discipline - Principals Reports' (Scale)'', which is numeric and can be found through the code BCBGDAS. The other variable is ``School Discipline - Principals Reports' (Index)'' (BCDGDAS), which is categorical and of primary interest in the results presented in the paper.} pertaining to discipline problems in the questionnaire responded by the school principals. As this ordinal variable --- which we specify in the linear predictor of CSP-BART --- can be completely defined by the other numeric variable of non-primary interest without any mismatch, we omit the numeric variable from the $\mathbf{X}_2$ matrix and likewise omit other variables similarly related to the other variables specified in $\mathbf{X}_1$ (`parents' education level' and `minutes spent on homework').

It is notable that the parameter estimates presented in the first column of Table \ref{TIMSS_additional_results}, using BART-based pre-screening where $p_2=20$ and CSP-BART, are broadly in line in terms of sign and significance (as determined by whether the reported 90\% credible intervals contain zero) with the alternative set of results for CSP-DART when $p_2 = 250$ and $n=3{,}802$ for all variables of primary interpretational interest. However, one noticeable aspect of the results of adapting CSP-DART to handle missing values in its trees --- in the third column of Table \ref{TIMSS_additional_results} --- is that a number of levels of the `minutes spent on homework' variable are now identified as significant effects, though their posterior means are similar in magnitude to the original CSP-BART results. This may be attributable to reduced posterior uncertainty given the larger sample size. However, we can also reiterate that the results for the additional scenarios with $p_2 = 250$ avail of a Dirichlet prior on the splitting probabilities \citep{linero2018bayesian}, so that important predictors are favoured over those which have little to no importance, and are thus based on fundamentally different models. Of course, this choice is particularly pertinent as in the new scenarios (i.e., CC and MT) the number of predictors available to the BART component via $\mathbf{X}_2$ is significantly greater than the number (i.e., $p_2 = 20$) used in the analysis of the TIMSS 2019 data in Section \ref{section_timss2019}.

Conversely, the CSP-DART results based on complete cases where $n = 1{,}448$ and $p_2 = 250$ show more pronounced differences. In particular, these result suggest that two out of the three variables of primary interest have no significant impact on students' mathematics scores, in sharp contrast to the two sets of results which consider more observations. It is not surprising that most of the estimates have greater variability when CSP-DART is applied on the complete cases, since in this case the number of observations $n$ is much less than in the other scenarios. Overall, despite some of these outlined differences, the conclusions that can be drawn from all three sets of results are mostly coherent; for the most part, we can still assert that students who experience school discipline problems tend to obtain worse mathematics scores, while students who spend more time on homework or whose parents have a higher level of education tend to exhibit better performance.

\section{Pima Indians Diabetes \label{appendix_pima}}
\setcounter{table}{0}
\setcounter{figure}{0}
\renewcommand{\thetable}{\Alph{section}.\arabic{table}}
\renewcommand{\thefigure}{\Alph{section}.\arabic{figure}}

We now analyse the well-known Pima Indians Diabetes dataset from the UCI Machine Learning Repository \citep{uci_repo}, which is available in \textsf{R} through the \texttt{mlbench} package \citep{mlbench}, to demonstrate the use of CSP-BART in a classification setting. Unlike the TIMSS 2019 data, here the response is binary rather than continuous and all covariates are continuous rather than categorical. The goal is to predict whether or not a patient has diabetes based on age, blood pressure, body mass index, glucose concentration, and $4$ other covariates. We analyse a corrected version of the data which treats physically impossible values of zero for a number of covariates as missing values, which we in turn omit. We are primarily interested in measuring the effects of age and glucose through the linear predictor along with possible non-specified interactions involving age, glucose, and/or the other six covariates accounted for by BART. As the response variable is binary, we use a probit link function following the data augmentation scheme of \citet{albert1993bayesian}. 

We only compare CSP-BART and SSP-BART, as the \texttt{VCBART} package cannot deal with binary responses. Henceforth, all parameter estimates are based on~a training set comprising a randomly chosen $80\%$ of the data and misclassification rates based on the remaining $20\%$ are used to quantify prediction accuracy. For CSP-BART, we specify age and glucose in $\mathbf{X}_1$ and supply all $8$ available covariates, including age and glucose, to the BART component. For SSP-BART, we specify age and glucose in $\mathbf{X}_1$ and the $6$ remaining covariates in $\mathbf{X}_2$, as SSP-BART does not allow for covariates to be shared across the linear and BART components. In both cases, the intercept is omitted from the $\mathbf{X}_1$ matrix, as described in Section \ref{CSP_BART_sec}.

We present the parameter estimates for age and glucose, with corresponding $90\%$ CIs, in Table \ref{tab3_pima}. Under both models, the estimates for both covariates indicate that, as they increase, the probability of observing~positive diabetes diagnoses also increases, and \emph{vice versa}. All CIs also have positive lower and upper limits. It is especially notable, however, that the CI for the age effect is bounded further away from zero under the CSP-BART model; i.e., we detect a more significant marginal age effect.

To highlight the efficacy of the hierarchical prior on $\boldsymbol\beta$, the double-grow and double-prune moves, and our other proposals for addressing non-identifiability, we also fit a hybrid model, equipped with the isotropic prior from SSP-BART, with age and glucose in both components, but without the double moves and stringent checks on tree-structure validity used in CSP-BART. Such a model achieves a misclassification rate of $19.23\%$ on the test set; slightly better than SSP-BART itself ($20.51\%$), though still inferior to the proper CSP-BART ($17.94\%$).

Under the hybrid model, we observe that the additional inclusion of age and glucose in the $\mathbf{X}_2$ matrix used by the BART component generates trees that occasionally use only age or only glucose. In this case, the linear predictor and BART component both try to estimate the effects of these covariates, which is not sensible as it generates non-identifiability issues between the two components and bias in the estimates of the parameters in the linear predictor. Overall, the benefits arising from i) sharing covariates among the components, ii) the employment of double-grow and double-prune moves, along with other checks on tree-structure validity, and iii) the adoption of the hierarchical prior on $\boldsymbol\beta$ are evident from the superior out-of-sample classification accuracy of CSP-BART.
\begin{table}[H]
\centering
\caption{Posterior mean estimates of the age (years) and glucose (mg/dL) effects on the diagnosis of diabetes, with corresponding $90\%$ CIs, according to CSP-BART and SSP-BART models fit to the training set $(80\%)$. \label{tab3_pima}}\vskip\smallskipamount
\setlength{\tabcolsep}{17.5pt}
\begin{tabular*}{406pt}{lcccc}
\toprule
&\multicolumn{2}{c}{\textbf{CSP-BART}} & \multicolumn{2}{c}{\textbf{SSP-BART}} \\\cmidrule{2-5}
\textbf{Covariate} & \textbf{Estimate}  & \textbf{$\mathbf{95}$\% CI}  & \textbf{Estimate}  & \textbf{$\mathbf{95}$\% CI}   \\
\midrule
Age       & $0.0634$ & $(0.0285; 0.1006)$ & $0.0287$  & $(0.0016; 0.0572)$   \\
Glucose   & $0.0359$ & $(0.0271; 0.0445)$ & $0.0296$  & $(0.0221; 0.0377)$   \\
\bottomrule
\end{tabular*}
\end{table}

\section{Extending the linear component in CSP-BART\label{appendix_extra_linear}}
\setcounter{table}{0}
\setcounter{figure}{0}
\renewcommand{\thetable}{\Alph{section}.\arabic{table}}
\renewcommand{\thefigure}{\Alph{section}.\arabic{figure}}

CSP-BART is largely presented throughout Section \ref{CSP_BART_sec} from the perspective of lending interpretability to main effects of primary interest in a linear component,~while allowing non-linearities and interactions to be estimated by a BART component, in order to ensure identifiability of the linear component even when both components share covariates in common. In Section \ref{random_effect_sec}, a version of CSP-BART is presented which also allows random effects along with fixed effects in the parametric component. In principle, the framework could be extended further to accommodate situations in which specific interactions and non-linear effects are of primary interpretational interest. For simplicity, we describe the additional extensions which follow in terms of a model which excludes random effects from the parametric component, though they could also be incorporated within a more flexible CSP-BART model with a multilevel structure.

\subsection{Non-linear effects of primary interest} \label{sec_nonlinear_primary_interest}

CSP-BART can be extended to model, via the linear predictor, \emph{non-linear} relationships between the response and (multiple) continuous covariates of primary interpretational interest, by replacing the linear component by an additive partial linear component. On the other hand, we recall that linear and non-linear effects of non-primary interpretational interest would continue to be estimated through the BART component of CSP-BART without pre-specification. To illustrate how CSP-BART could be extended, suppose we have the following model: 
$$y_{i} = \underbrace{\mathbf{x}_{1i}\boldsymbol\beta}_\text{Linear effects} + \underbrace{f_1(x_{i1}) + \ldots + f_p(x_{ip})}_\text{Non-linear effects} + \underbrace{g(\mathbf{x}_{2i})}_\text{BART} + \epsilon_{i}.$$
In this model, the predictors of primary interest are split into two \emph{disjoint} groups: linear and non-linear. The linear effects are modelled via $\mathbf{x}_{1i}\boldsymbol\beta$ and the non-linear effects are estimated via smooth functions of the form $f_j(x) = \sum_{k = 1}^{K_j} B_k(x) \alpha_k$, where $B_k(\cdot)$ denotes the $k$-th basis function and $\boldsymbol\alpha_j = (\alpha_1, \ldots, \alpha_{K_j})$ is a parameter vector of unknown coefficients which we aim to estimate. The flexibility of adding non-linear effects to the linear predictor of the model requires i) the choice of the basis functions, ii) a prior on $\boldsymbol\alpha_j$, and iii) a careful specification of the predictors used by the BART component (i.e., $\mathbf{X}_2$). 

Regarding the choice of the basis functions, there are various options which offer different theoretical and computational advantages. For instance, knot-based approaches, such as cubic smoothing splines \citep{reinsch1967smoothing} and P-splines \citep{eilers1996flexible}, offer both simplicity and flexibility, and are commonly used in practice. Knot-free approaches like thin plate splines \citep{duchon1977splines} are also an option. The prior on $\boldsymbol{\alpha}_j$ can be similar to the one placed on the parameter vector $\boldsymbol{\beta}$ in Section \ref{CSP_BART_sec}, where it is assumed that $\boldsymbol\beta \sim \mbox{MVN}(\mathbf{0}, \Omega_\beta)$. The only difference here is that we can assume $\boldsymbol{\alpha}_j\given\lambda_j \propto \exp ( - \frac{1}{2\lambda_j} \boldsymbol{\alpha}_j^\top \boldsymbol{\Omega}_\alpha^{-1} \boldsymbol{\alpha}_j)$, where $\boldsymbol{\Omega}_\alpha$ is an appropriate penalty matrix and $\lambda_j$ is a smoothing parameter, equipped with its own prior distribution, which balances the trade-off between model fit and smoothness of the fit \citep{lang2004bayesian, wood2017generalized}.

To allow for the BART component to estimate interactions among variables of primary interest and between variables of primary and non-primary interest in cases where such variables of primary interest are specified as non-linear effects, the key point is to add to $\mathbf{X}_2$ the \emph{original} predictors associated with non-linear effects instead of their basis functions, $B_k(x)$. In doing so, we avoid adding unnecessary predictors which represent pre-defined partitions of the domain of $x$ and allow the BART component to create its own splits based on the original predictor. In this way, we can use the double-grow and double-prune moves to prevent BART from estimating now both linear and non-linear marginal effects of primary interest. 

\subsection{Interactions of primary interest}
\label{sec_interactions_primary_interest}

To extend CSP-BART to model interaction effects of primary interpretational interest via the linear predictor, additional tree-generation moves\footnote{Along with related ideas already described in Section \ref{CSP_BART_sec}, such as stringent checks on the tree topology and changes on the priors of relevant terminal node parameters.} would be required. We now explain the rationale for introducing `triple-grow' and `triple-prune' moves in this context. If there is a primary interpretational interest in a two-way interaction effect, the `triple' moves apply in the specific situation where \emph{both} predictors comprising the interaction are i) also specified as main effects in the linear predictor and ii) also included in $\mathbf{X}_2$. To illustrate our point, we consider the following example:
\begin{equation}
\label{marginal_n_interaction_effects}
    y_i = \beta_1 x_i + \beta_2 z_i + \beta_3 x_i z_i + \sum_{t = 1}^{T} g\left(\mathbf{x}_{2i}, \bm{\mathcal{M}}_{t}, \mathcal{T}_{t}\right) + \epsilon_i. 
\end{equation} 
Here, $x$ and $z$ are both specified in $\mathbf{X}_1$ as main effects, along with the two-way interaction between them. Both $x$ and $z$ are also included in $\mathbf{X}_{2} = (x, z, a_{1}, \ldots, a_{p})$, where $a_{j}$ denotes an additional predictor of non-primary interest. Figure \ref{fig:invalid_trees} shows what happens when the single grow and double-grow moves are applied using predictors which belong to both $\mathbf{X}_1$ and $\mathbf{X}_2$ under the model in Equation \eqref{marginal_n_interaction_effects}. In panel (a), the tree $\mathcal{T}_1$ is obtained from a single grow move on $x \in \{\mathbf{X}_1 \cap \mathbf{X}_2\}$. However, $\mathcal{T}_1$ is \emph{invalid} in the sense that it would bring a non-identifiability issue into the model since it estimates the marginal effect of $x$, which in turn is already being estimated in the linear predictor via $\beta_1$. Analogously, in panel (b), the tree $\mathcal{T}_2$ splits on $x$ and $z$ through a double-grow move, thus modelling an interaction effect between them which is already accounted for by $\beta_3$. Nonetheless, the double-grow move by itself is not enough to ensure the validity of the resulting tree, since the interaction between $x$ and $z$ is of primary interest and is also specified in the linear predictor. We note that the trees in Figure \ref{fig:invalid_trees} would remain equally \emph{invalid} if $\mathcal{T}_1$ split on $z$ rather than $x$ and if $x$ and $z$ were interchanged in $\mathcal{T}_2$.
% In panels (a) and (b), both trees are obtained from a single grow move on $x \in \{\mathbf{X}_1 \cap \mathbf{X}_2\}$ and $z \in \{\mathbf{X}_1 \cap \mathbf{X}_2\}$, respectively. However, these trees are \emph{invalid} in the sense that they would bring a non-identifiability issue into the model since they estimate the marginal effects of $x$ and $z$, which in turn are already being estimated in the linear predictor via $\beta_1$ and $\beta_2$. Analogously, in panels (c) and (d), the trees split on $x$ and $z$ through a double-grow move, thus modelling an interaction effect between them. Nonetheless, the double-grow by itself is not enough to ensure the validity of these trees, since the interaction between $x$ and $z$ is of primary interest and is also specified in the linear predictor. 
\begin{figure}[H]
  \captionsetup[subfigure]{font=normalsize}
        \centering
        % \vspace{-15cm}
                \begin{subfigure}[t]{195pt}
                    \centering
                    \caption[]%
                    {$\mathcal{T}_{1}$}
                        \begin{forest}
                        for tree={
                            grow=south, draw, minimum size=3ex, 
                            inner sep=3pt, % control the overall tree size
                            s sep=7mm,
                            l sep=6mm
                            }
                        [$\:x \le 0.5\:$,
                            [$\mu_{1}$, circle,  edge label={node[midway,left, font=\footnotesize]{$\mbox{ TRUE }$}},]
                            [$\mu_{2}$, circle,  edge label={node[midway,right, font=\footnotesize]{$\mbox{ FALSE }$}},]
                        ]
                        \end{forest}
            %\label{fig:mean and std of net14}
                \end{subfigure}
        \hspace{-0.1cm}
        \begin{subfigure}[t]{195pt}
            \centering
            \caption[]%
            {$\mathcal{T}_{2}$}
            \begin{forest}
for tree={
    grow=south, draw, minimum size=3ex, 
    inner sep=3pt, % control the overall tree size
    s sep=7mm,
    l sep=6mm
    }
[$\:x \le 0.5\:$,
    [$\:z \le - 0.5\:$,
        [$\mu_{1}$, circle]
        [$\mu_{2}$, circle]
    ]
        [$\mu_{3}$, circle]
]
\end{forest}
            %label{fig:mean and std of net14}
        \end{subfigure}
\caption{Examples of invalid trees based on the model in Equation \eqref{marginal_n_interaction_effects}. Both $x$ and $z$ belong to $\mathbf{X}_1$ and $\mathbf{X}_2$. These trees redundantly estimate a marginal of $x$ and an interaction effect between $x$ and $z$, which would cause non-identifiability issues, as these effects are also specified in the linear predictor in Equation \eqref{marginal_n_interaction_effects} via $\beta_1$ and $\beta_3$, respectively. \label{fig:invalid_trees}}
\end{figure}
Figure \ref{fig:valid_trees} shows tree structures which are \emph{valid} in the context of the model in Equation \eqref{marginal_n_interaction_effects}. Through a double-grow move on $x$ and $a_j$, the tree $\mathcal{T}_3$ in panel (a) estimates an interaction effect between $x$ and a predictor ($a_j$) exclusive to $\mathbf{X}_2$. This tree is valid from an identifiability point of view, as the interaction it captures is not specified in the linear predictor. Notably, the tree $\mathcal{T}_4$ in panel (b) of Figure \ref{fig:valid_trees} exemplifies the `triple-grow' move involving an interaction of primary interpretational interest. Recalling $\mathcal{T}_2$ in panel (b) of Figure \ref{fig:invalid_trees}, a tree containing splits on $x$ and $z$ only is invalid because it estimates an effect which is already estimated by the model's linear component. However, a tree such as $\mathcal{T}_4$ in panel (b) of Figure \ref{fig:valid_trees}, containing $x$, $z$, \emph{and} any other predictor ($a_j$) in the same branch, is valid as it estimates a 3-way interaction which is not in the linear predictor. As before, the trees in Figure \ref{fig:valid_trees} would remain equally \emph{valid} if $\mathcal{T}_3$ initially split on $z$ rather than $x$ and if $x$ and $z$ were interchanged in $\mathcal{T}_4$.
\begin{figure}[H]
  \captionsetup[subfigure]{font=normalsize}
        \centering
        \begin{subfigure}[t]{195pt}% approx 0.475 * \textwidth
            \centering
            \caption[]%
            {$\mathcal{T}_{3}$}
            \begin{forest}
for tree={
    grow=south, draw, minimum size=3ex, 
    inner sep=3pt, % control the overall tree size
    s sep=7mm,
    l sep=6mm
    }
[$\:x \le 0.5\:$,
    [$\:a_{j} \le 0.5\:$, edge label={node[midway, font=\footnotesize, left]{$\mbox{ TRUE }$}}
        [$\mu_{11}$, circle,  edge label={node[midway,left, font=\footnotesize]{$\mbox{ TRUE }$}},]
        [$\mu_{12}$, circle,  edge label={node[midway,right, font=\footnotesize]{$\mbox{ FALSE }$}},]
    ]
    [$\mu_{13} \approx 0$, circle,  edge label={node[midway,right, font=\footnotesize]{$\mbox{ FALSE }$}},]
]
\end{forest}
            % \label{fig:mean and std of net14}
        \end{subfigure}
        \hfill
        \begin{subfigure}[t]{195pt}  
            \centering 
            \caption[]%
            {$\mathcal{T}_{4}$}
            \begin{forest}
for tree={
    grow=south, draw, minimum size=3ex, 
    inner sep=3pt, % control the overall tree size
    s sep=7mm,
    l sep=6mm
    }
[$\:x \le 0.5\:$,
    [$\:z \le -0.5\:$, edge label={node[midway, font=\footnotesize, left]{}}
        [$\:a_j \le 0.5\:$, edge label={node[midway, font=\footnotesize, left]{}}
        [$\mu_{11}$, circle,  edge label={node[midway,left, font=\footnotesize]{}},]
        [$\mu_{12}$, circle,  edge label={node[midway,right, font=\footnotesize]{}},]
    ]   
        [$\mu_{13} \approx 0$, circle,  edge label={node[midway,right, font=\footnotesize]{}},]
    ]
    [$\mu_{14} \approx 0$, circle,  edge label={node[midway,right, font=\footnotesize]{}},]
]
\end{forest}
            %\label{fig:mean and std of net24}
        \end{subfigure}
\caption{Examples of valid trees based on the model in Equation \eqref{marginal_n_interaction_effects}. While $a_j$ is exclusive to $\mathbf{X}_2$, $x$ and $z$ belong to both $\mathbf{X}_1$ and $\mathbf{X}_2$. $\mathcal{T}_3$ estimates an interaction effect between $x$ and $a_j$ via a double-grow move; were $a_j$ at the root node, then a single move could be applied and the resulting tree would also be valid. $\mathcal{T}_4$ illustrates a `triple-grow' move involving $x$, $z$, and $a_j$. Further non-identifiability issues are avoided by modifying the prior on the relevant predicted values to $\mu_{t\ell} \sim \mbox{N}(0, \sigma^2 \approx 0)$, which in turn shrinks the posterior predicted values towards zero, as per the double-grow move. \label{fig:valid_trees}}
\end{figure}
It is possible to further generalise this example to cases where there is a primary interpretational interest in a three-way interaction effect. If all three predictors comprising the interaction belong to $\mathbf{X}_1 \cap \mathbf{X}_2$, `quadruple' moves will be necessary. However, such moves would rarely be accepted in practice as the branching process prior placed on the structure of the trees in BART tends to penalise deep/asymmetric trees. Though we have focused in our example on the triple-grow move only, the counterpart `triple-prune' move would need to be added to the transition kernel of the BART component in CSP-BART in order to generate a reversible Markov chain. Consider, for example, the tree $\mathcal{T}_4$ in panel (b) of Figure \ref{fig:valid_trees}, which would need to be triple-pruned in order to bypass an invalid tree such as $\mathcal{T}_2$ in panel (b) of Figure \ref{fig:invalid_trees}. Consequently, quadruple-grow and quadruple-prune moves would be required whenever three-way interactions are specified in the linear predictor. Finally, we reiterate that `single-change' and `single-swap' moves are sufficient for dealing with non-identifiability issues that may arise between the linear and BART components, even in the presence of interaction effects (of any order) in the linear predictor. Nonetheless, as pointed out in Section \ref{CSP_BART_sec}, stringent checks should be placed on the validity of the trees proposed by such moves.

\end{appendix}

%%%%%%%%%%%%%%%%%%%%%%%%%%%%%%%%%%%%%%%%%%%%%%
%% Supplementary Material, including data   %%
%% sets and code, should be provided in     %%
%% {supplement} environment with title      %%
%% and short description. It cannot be      %%
%% available exclusively as external link.  %%
%% All Supplementary Material must be       %%
%% available to the reader on Project       %%
%% Euclid with the published article.       %%
%%%%%%%%%%%%%%%%%%%%%%%%%%%%%%%%%%%%%%%%%%%%%%
% \begin{supplement}
% \stitle{Title of Supplement A}
% \sdescription{Short description of Supplement A.}
% \end{supplement}
% \begin{supplement}
% \stitle{Title of Supplement B}
% \sdescription{Short description of Supplement B.}
% \end{supplement}

\end{document}